\pdfoutput=1
\documentclass[11pt]{article}
\usepackage{ACL2023}
\usepackage[utf8]{inputenc}
\usepackage[T2A,LAE,T1]{fontenc}
\usepackage{times}
\usepackage{latexsym}
\usepackage{tabularx}
\usepackage[main=english,farsi]{babel}
\babelprovide[import=fa]{farsi}
\usepackage{microtype}

\usepackage{makecell}
\usepackage{graphicx}
\usepackage{lipsum}
\usepackage{inconsolata}
\usepackage{booktabs}
\usepackage{color, colortbl}
\usepackage{enumitem}
\usepackage{caption}
\usepackage{stfloats}
\usepackage[export]{adjustbox} 
\usepackage{adjustbox}
\usepackage{amssymb} 
\usepackage{utfsym}
\usepackage{mathtools}
\usepackage{amsmath}
\usepackage{subcaption}
\usepackage{multirow}
\usepackage{afterpage}
\usepackage{xcolor}
\definecolor{Gray}{gray}{0.9}
\usepackage{hyperref}

\usepackage{alphalph}
\title{Crafting Tomorrow's Headlines: Neural News Generation and Detection \\ in English, Turkish, Hungarian, and Persian}

\author{Cem Üyük, Danica Rovó, Shaghayegh Kolli, Rabia Varol, \\ {\bf Georg Groh \and Daryna Dementieva} \\
Technical University of Munich, Department of Informatics, Germany \\ 
\texttt{\{danica.rovo, shaghayegh.kolli, rabia.varol, daryna.dementieva\}@tum.de} \\
\texttt{\{uyc, grohg\}@in.tum.de}
}

\begin{document}
\maketitle

\begin{abstract}
In an era dominated by information overload and its facilitation with Large Language Models (LLMs), the prevalence of misinformation poses a significant threat to public discourse and societal well-being.  A critical concern at present involves the identification of machine-generated news. In this work, we take a significant step by introducing a benchmark dataset designed for neural news detection in four languages: English, Turkish, Hungarian, and Persian. The dataset incorporates outputs from multiple multilingual generators (in both zero-shot and fine-tuned setups) such as BloomZ, LLaMa-2, Mistral, Mixtral, and GPT-4. Next, we experiment with a variety of classifiers, ranging from those based on linguistic features to advanced Transformer-based models and LLMs prompting. We present the detection results aiming to delve into the interpretability and robustness of machine-generated text detectors across all target languages.
\end{abstract}

\section{Introduction}
The remarkable power of current advances in Natural Language Processing (NLP) has enabled the creation of text that closely resembles human-authored content~\cite{achiam2023gpt}. While these technological achievements offer both positive and negative prospects, a significant concern arises from the proliferation of neural-generated news, which may not always present accurate information. Thus, in the pipeline of fake news detection, a pivotal stage can be authorship identification, either by a human or a machine. 

\begin{figure}[h!tbp]
    \centering
    \includegraphics[width=0.48\textwidth]{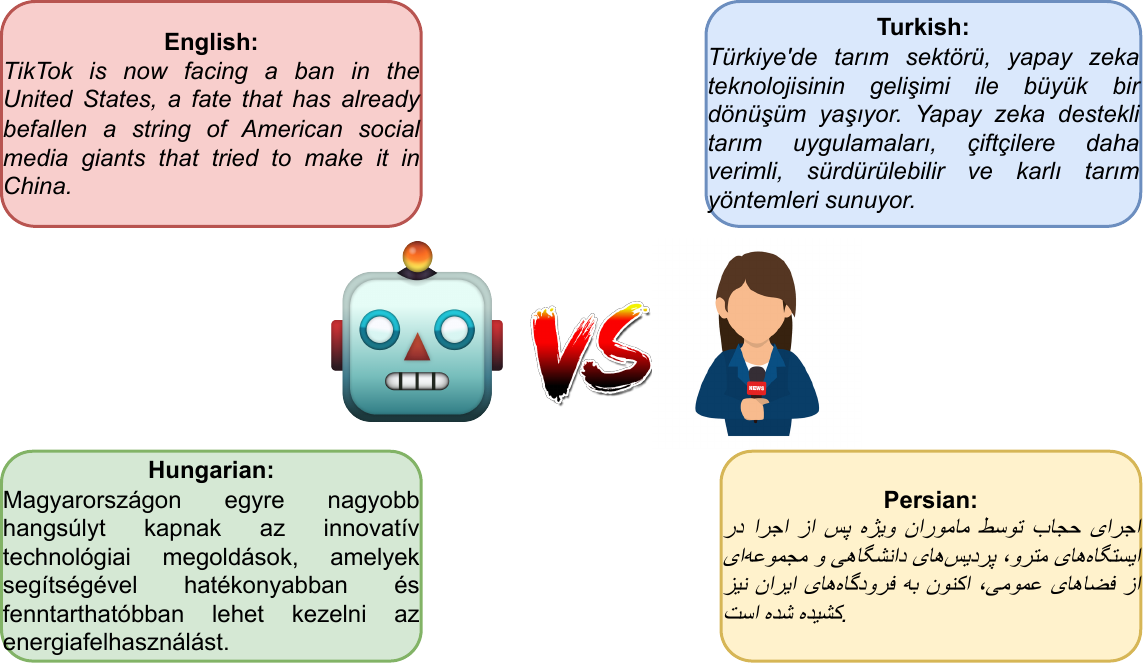}
    \caption{The aim of this work is to create a benchmark dataset for human-written vs. machine-generated news distinguishing in four languages---English, Turkish, Hungarian, and Persian.}
    \label{fig:logo}
\end{figure}

The task of news or story generation appeared to be a fascinating challenge already since the popularization of RNNs \cite{zhou-etal-2018-neural,DBLP:journals/access/ZhengWCS18}. Indeed, it can be an excellent technology for editing support \cite{murao-etal-2019-case}. Then, the appearance of GPT-2~\cite{radford2019language} opened new horizons in neural news generation~\cite{DBLP:conf/aaai/ShuLDL21}. However, generators still suffered from a lack of human fluency and were not able to handle proper long text generation, forgetting about the context. As a result, in the news domain, such generative tasks as news headline generation~\cite{DBLP:conf/www/GuM0L00FYZZ20} and news summarization~\cite{sethi2017automatic,DBLP:conf/emnlp/EgonmwanC19} received more attention.

In the past, several machine-generated text detection benchmarks were created in the domains of scientific papers~\cite{DBLP:journals/information/AbdallaMDMG23}, story-telling texts~\cite{DBLP:journals/corr/abs-2305-13242}, question-answering datasets, and Wikipedia articles~\cite{DBLP:journals/corr/abs-2305-14902}. For the news domain, such datasets were generated with the Grover model~\cite{DBLP:conf/nips/ZellersHRBFRC19} based on GPT-2. However, especially without evidence, e.g. historical facts, the nature of the task remains challenging, given the advances in the domain of NLP. Moreover, all these datasets were mostly designed only for one language, English, and did not cover the news domain and recent models sufficiently. To this end, we chose Turkish, Hungarian, and Persian (native languages of the authors) to create a proof-of-concept to determine whether underrepresented languages can also thrive with the recent generative AI advances, specifically for news creation and classification of AI-generated content. \\ 
Thus, our contributions are the following: 
\begin{itemize}
    \item We construct the first-of-its-kind benchmark dataset of neural news detection in English, Turkish, Hungarian, and Persian.
    \item In this dataset, we cover generation outputs from current state-of-the-art LLMs---BloomZ, LLaMa, Mistral, Mixtral, and GPT-4.
    \item We conduct a comprehensive study exploring the power of diverse baselines in this neural news detection task.
    \item We report the classification scores per discriminator, per language, and per generator, diving into the interpretations of the results.
\end{itemize}
The constructed benchmark dataset and the best-performing classifiers are available for public usage online.\footnote{\href{https://huggingface.co/datasets/tum-nlp/neural-news-benchmark}{tum-nlp/neural-news-benchmark} \newline Also, refer to Appendix~\ref{sec:app_opensource_models} for all models links.}

\section{Related Work}
The first dataset of generated texts that specifically focused on the news domain was created using Grover
~\cite{DBLP:conf/nips/ZellersHRBFRC19}. The~Grover model for neural news generation was based on GPT-2~\citep{radford2019language} and already created a strong benchmark for neural news detection in English. After that, a dataset of GPT-2-generated news headlines was also introduced in~\cite{maronikolakis2020identifying}.

Recently, in~\citet{DBLP:journals/information/AbdallaMDMG23}, a benchmark dataset for generated scientific texts detection was presented. The authors utilized powerful modern language models (LMs) including those specifically fine-tuned for scientific text domain, like SCIgen \citep{stribling2005scigen} and Galactica \citep{taylor2022galactica}. \

In the context of multilingual datasets, the M4 dataset~\cite{DBLP:journals/corr/abs-2305-14902} covers seven languages from various domains like question-answering, Wikipedia, news, and legal texts. The authors also benchmarked various baselines, ranging from linguistically informed approaches to Transformer-based classifiers, and concluded with zero-shot prompting of language models. Additionally, the multilingual benchmark collection can be expanded with the HC3 corpus~\cite{DBLP:journals/corr/abs-2301-07597}, which includes question-answering data for English and Chinese. For Russian, the RuATD~\cite{DBLP:journals/corr/abs-2206-01583} corpus provides text summarization, paraphrasing, and translation data.

However, previous studies have not addressed the examination of neural texts for underrepresented and complex languages such as Turkish, Hungarian, and Persian. We are closing this gap with our work while also including a popular language, English. 

\section{Models for News Generation}
The first step before training models for news generation was evaluating the pre-trained LLMs in producing news content akin to proficient news editors. Therefore, the aim of the initial experiments was to assess the capability of existing models in generating news content across all four considered languages in a zero-shot setting and then fine-tune promising models for the downstream task of news generation.

\subsection{LLM Assessment Downstream Fine-Tuning Selection}
We assessed various models' capability in a news generation task in four languages. For each model, 100 samples were generated to test the model's capacity for the task and each language (examples are in Appendix \ref{sec:initial_round_of_0_zhot_gens}). Specifically, zero-shot generations were done with the following models: 

\paragraph{English Monolingual Models} Alpaca~\cite{alpaca},  Dolly-v2~\cite{DatabricksBlog2023DollyV2} and GPT-2 XL\footnote{\href{https://huggingface.co/openai-community/gpt2-xl}{https://huggingface.co/openai-community/gpt2-xl}} were only prompted in English.
    
\paragraph{Monolingual Models with Emerging Abilities} LLaMa-2-7B and LLaMa-2-Chat-7B \cite{llama2_paper}, Persian-LLaMa-7B-Merged\footnote{\href{https://huggingface.co/mostafaamiri/persian_LLaMa_7B_merged} {https://huggingface.co/mostafaamiri/persian{\textunderscore}LLaMa{\textunderscore}7b}} and Mistral~\cite{mistral} which mostly were pre-trained on English data with possibility to transfer knowledge in unseen tasks to other languages.

\begin{figure*}[h!tbp]
  \centering
  \begin{subfigure}[b]{0.19\textwidth}
    \centering
    \includegraphics[width=\linewidth, height=3cm]{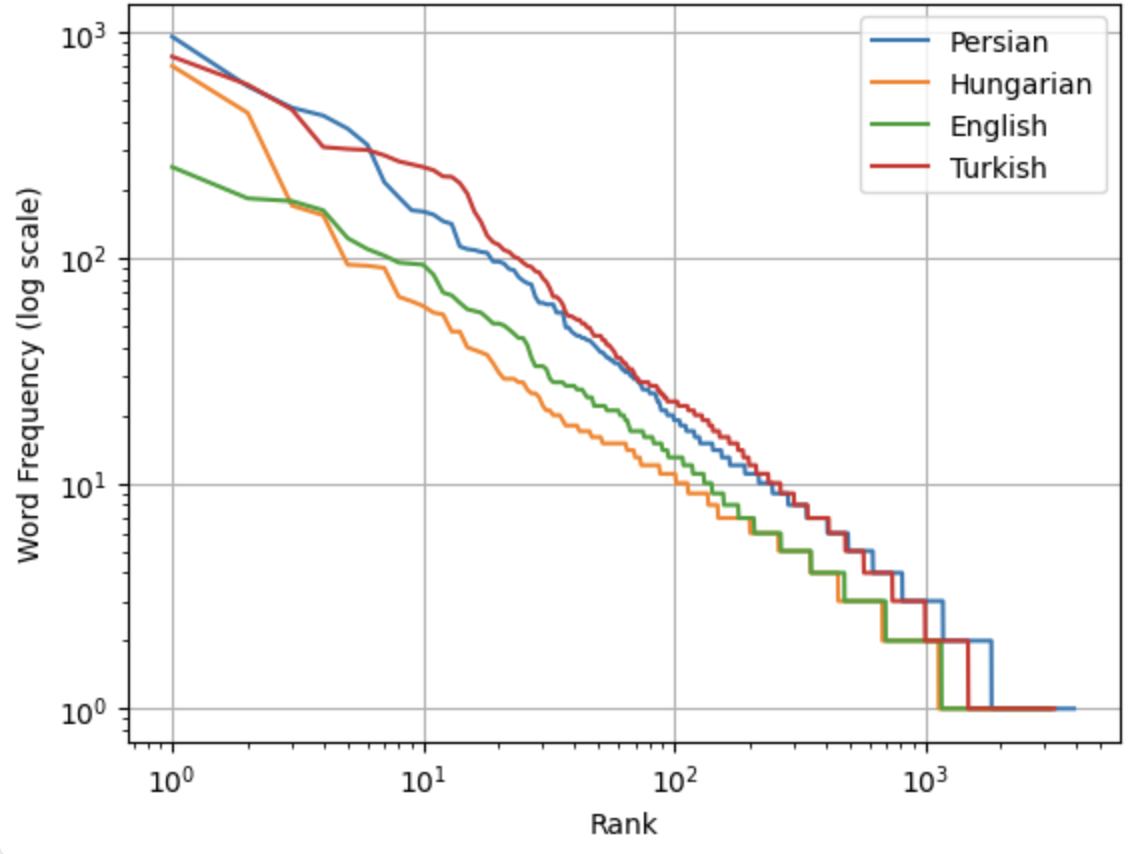}
    \caption{Zero-Shot BloomZ-3B.}
    \label{fig:bloomz-0-shot-zipf}
  \end{subfigure}
  \hfill
  \begin{subfigure}[b]{0.19\textwidth}
    \centering
    \includegraphics[width=\linewidth, height=3cm]{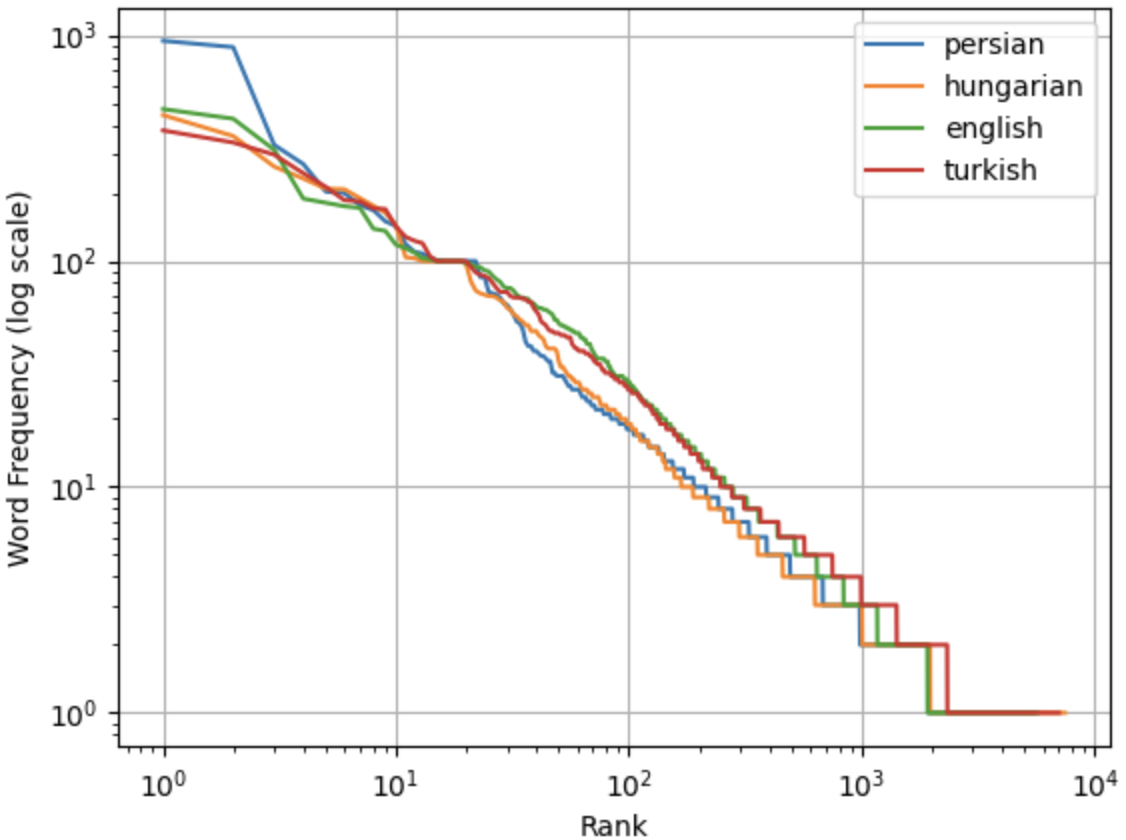}
    \caption{Fine-tuned BloomZ-3B.}
    \label{fig:bloomz-3b-tuned-zipf}
  \end{subfigure}
  \hfill
  \begin{subfigure}[b]{0.19\textwidth}
    \centering
    \includegraphics[width=\linewidth, height=3cm]{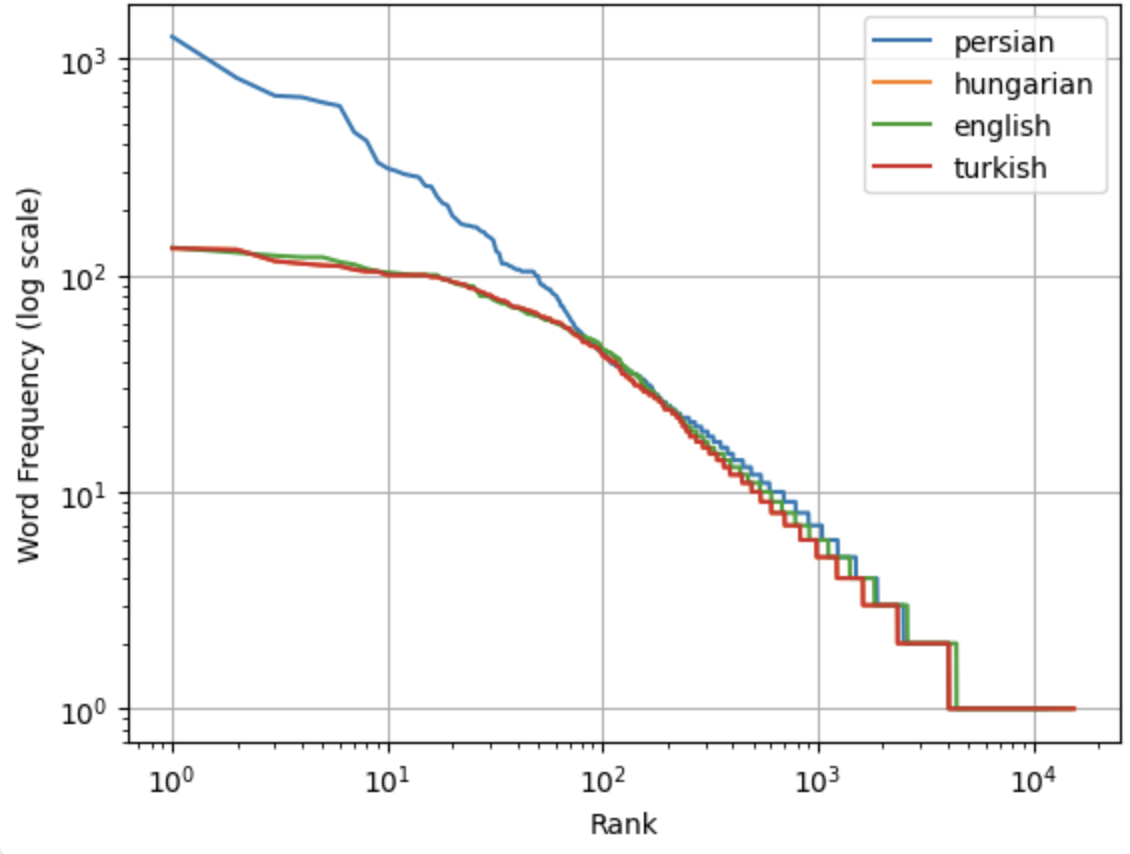}
    \caption{Zero-Shot\\LLaMa.}
    \label{fig:llama-0-shot}
  \end{subfigure}
  \hfill
  \begin{subfigure}[b]{0.19\textwidth}
    \centering
    \includegraphics[width=\linewidth, height=3cm]{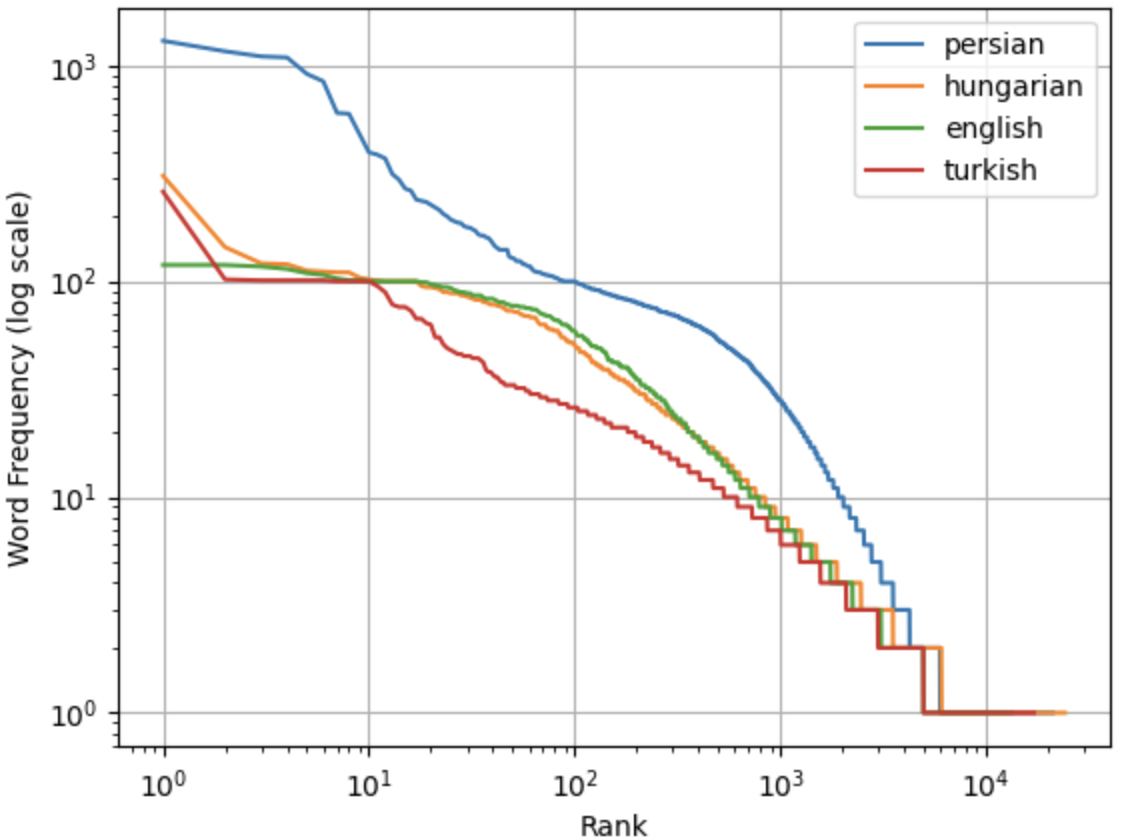}
    \caption{Fine-tuned\\LLaMa.}
    \label{fig:llama-3b-tuned-zipf}
  \end{subfigure}
  \begin{subfigure}[b]{0.19\textwidth}
    \centering
    \includegraphics[width=\linewidth, height=3cm]{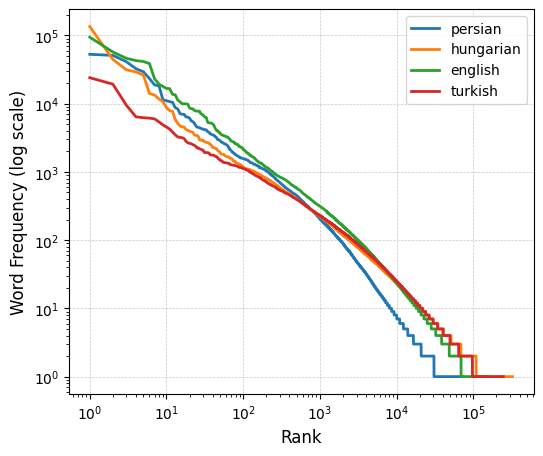}
    \caption{Human-written ground truth.}
    \label{fig:human-zipf}
  \end{subfigure}
  \caption{Zipf's Distribution of articles before and after fine-tuning, and of human-written ground truth. \\ The vocabulary of generative models, although smaller than human-written text, approximately follows a power-law.  
  }
  \label{fig:initial-sect-zipf-plots}
\end{figure*}
    
\paragraph{Multilingual Models with Emerging Abilities} BloomZ-3B~\cite{bloomz}, Mixtral~\cite{mixtral}, GPT-3 and GPT-4~\cite{achiam2023gpt}---were prompted in all of our target languages even if they were not pre-trained directly on the required languages. In this setup, our objective was twofold: firstly, to determine if these models had encountered any data points from out-of-domain languages during pre-training, and secondly, to assess whether they had undergone incidental language acquisition as a result of exposure to linguistically familiar languages as suggested by the design in BloomZ, incorporating the effects of multi-task and multilingual training \cite{bloomz}.

\subsection{LLMs Selection}\label{subsection:llm-selection}
To have a preliminary assessment of the generation quality of LLMs, a RoBERTa model trained on the Corpus of Lingustic Acceptability (CoLA) was used for English \cite{Cola_citation}. For multilingual comparison, two methods were employed: the Zipf's distribution~\cite{zipf} of the text and analyzing the top \textit{Term Frequency-Inverse Document Frequency} (TF-IDF) vectors~\cite{murphy2013machine}. These analyses confirmed that the language flow follows a power distribution and that the predominant features align with the target language.

Figures in \ref{fig:initial-sect-zipf-plots} illustrate the before and after fine-tuning stages in Zipf's distribution across the primary models chosen for training for all four languages. In addition to Zipf's distributions, these models also show impressive CoLA scores and convincing TF-IDF features. 
All the plots reporting this analysis are showcased in Appendix \ref{sec:plots}. 

As a result, both BloomZ-3B and LLaMa-2-Chat-7B, along with Persian-LLaMa-7B, emerge as promising candidates for the fine-tuning of the downstream task of news generation. Mixtral has impressive zero-shot capabilities as well, but this model was published after the fine-tuning efforts had been completed. Moreover, the selected models that were prompted in English and the other languages showed negligible to minor capabilities, even if the prompt language was not reported in pre-training data, possibly resulting from some of these languages accidentally being in the pre-training data.

\subsection{Downstream Task Fine-tuning}
We fine-tuned the selected LLMs for the downstream task of news generation. We provide all the details of the fine-tuning process below.

\subsubsection{Dataset for News Generation Task Fine-tuning}\label{sec:gen-fintune-data}
The data used for fine-tuning BloomZ-3B were scraped from open-source websites. Overall, $3,000$ in human-written news articles per language went into fine-tuning this model. On the other hand, LLaMa-based models were fine-tuned with $6,000$ samples for all languages. The collected dataset was filtered with topics of politics, economics, international news and alike. Having seen the proof of concept work with BloomZ-3B—demonstrating its ability to learn from limited data—we repeated the process with LLaMa using more samples. Since both models show similar zero-shot capabilities, this could provide insight into how learning scales with more data points.

In order to keep the prompt engineering to a minimum and assess the bare generation capabilities of the models after fine-tuning, the prompts were designed simplistically as such: for each news sample, an instruction, a headline, and the initial ten words of the original article were provided, along with some marker tokens. Below, we provide an example (For more examples, see Appendix~\ref{sec:app_llms_prompts_generation}):
\begin{quote}
    \textsf{\footnotesize Write a news article in Turkish. Headline: \{Headline in Turkish\} Article: \{First 10 words of real article in Turkish\} [EOP]}
\end{quote}
\sloppy
The dataset was constructed in this fashion and, specifically, for LLaMA based models, the cc\_news\footnote{\href{https://huggingface.co/datasets/cc_news}{https://huggingface.co/datasets/cc\_news}} dataset for English, batubayk/TR-news\footnotemark[5] dataset for Turkish, batubayk/HU-news\footnotemark[5] dataset for Hungarian, RohanAILab/persian\_news\footnotemark[6] and web scrapes\footnotemark[7] for Persian were used.
\footnotetext[5]{\href{https://github.com/batubayk/news_datasets?tab=readme-ov-file}{https://github.com/batubayk/news\_datasets}}
\footnotetext[6]{\href{https://huggingface.co/datasets/RohanAiLab/persian\_news\_dataset}{https://huggingface.co/datasets/RohanAiLab/\\persian\_news\_dataset}}
\footnotetext[7]{\href{https://www.radiofarda.com/}{https://www.radiofarda.com/}}
\subsubsection{BloomZ-3B}\label{subsection:finetuning-bloomz-3b}
\paragraph{Languages} English, Turkish, Hungarian, Persian.
\paragraph{Fine-tuning Dataset} We used $3,000$ data points per language, therefore, $12,000$ total data points and then divided into train/val/test splits with proportions $60:20:20$.
\paragraph{Model} Multilingual model BloomZ-3B\footnotemark[8] was fine-tuned in a mixed language setting.
\footnotetext[8]{\href{https://huggingface.co/bigscience/bloomz-3b}{https://huggingface.co/bigscience/bloomz-3b}}
\paragraph{Hyper-parameters} 
Below is the overview of hyper-parameters used for this iteration: 

 \begin{table}[ht]
     \centering
     \footnotesize
     \begin{tabular}{c|c|c|c}
     \toprule
           Language & Epochs & LR & Optimizer \\
           \midrule
           Mixed & 2 & {$2e^{-5}$} &  \verb|adafactor| \\
           \bottomrule
     \end{tabular}
     \caption{BloomZ-3B Hyper-parameter Setting \cite{DBLP:adafactor}.}
     \label{tab:bloomz_persian_hyperparams}
 \end{table}

\paragraph{Training} To prevent the quick overfitting, a validation set was used considering early stopping every $200$ steps with a patience of $3$. In this setting, the training lasted until the end of epoch $2$.

\subsubsection{LLaMa-2-Chat-7B}\label{subsection:finetuning-llama}
The chat version of the original LLaMa-2 model~ \cite{llama2_paper} provided better results during the zero-shot experiments, therefore, this model was downstream fine-tuned for English, Turkish and Hungarian separately. However, the very same model was not able to fit on Persian news data. Therefore, we leave out Persian from this setup. The fine-tuning configuration for this model looks as follows:
\paragraph{Languages} English, Turkish, Hungarian.

\paragraph{Fine-tuning Dataset} $6,000$ data-points per language went into each fine-tuning iteration, and the generic $60:20:20$ training, validation, and test set proportions were applied.

\paragraph{Model} LLaMa-2-Chat-7B\footnote[9]{\href{https://huggingface.co/meta-llama/Llama-2-7b-chat-hf}{https://huggingface.co/meta-llama/Llama-2-7b-chat-hf}} version was used for down-stream fine-tuning.

\paragraph{Hyper-parameters} Below is the training setup similar to BloomZ-3B, except for a smaller learning rate, as we wanted the model to see more data points and training to converge slower to get closer to global minima using \verb|adafactor| \cite{DBLP:adafactor}.
\begin{table}[ht]
     \centering
     \footnotesize
     \begin{tabular}{c|c|c|c}
     \toprule
        Language & Epochs & LR & Optimizer \\
        \midrule
        English & 2 & \multirow{3}{*}{$2e^{-4}$} & \multirow{3}{*}{\texttt{adafactor}} \\
        Turkish & 1 & &   \\
        Hungarian & 2.25 & &   \\
        \bottomrule
     \end{tabular}
     \caption{LLaMa-2-Chat-7B Hyper-parameter Settings.}
     \label{tab:LLaMa2_hyperparams}
 \end{table}
\paragraph{Training} In general, the LLaMa-2-Chat-7B model was able to train longer compared to BloomZ-3B and fit the data well, though without overfitting as a validation set was used to early-stop the training with a patience of $3$. For example, the training continued nearly until the epoch 
$4$, but the best validation was achieved in epoch $2$ for English.

\subsubsection{Persian-LLaMa-7B-Merged}\label{subsection:finetuning-llama-persian}

\paragraph{Languages} Persian.

\paragraph{Fine-tuning Dataset} $6,000$ data-points for Persian news data was utilized for fine-tuning.

\paragraph{Model} A specific instance of LLaMa extensively fine-tuned in Persian was utilized for our downstream news generation task.

\paragraph{Hyper-parameters} An overview of hyper-parameters used to train the Persian LLaMa model can be seen below.
\begin{table}[h]
     \centering
     \footnotesize
     \begin{tabular}{c|c|c|c}
     \toprule
           Language & Epochs & LR & Optimizer \\
           \midrule
           Persian & 2 & $2e^{-4}$  &  \verb|adafactor| \\
           \bottomrule
     \end{tabular}
     \caption{Persian-LLaMa-7B-Merged setup.}
     \label{tab:llama_persian_hyperparams}
 \end{table}
 
\paragraph{Training} The Persian-LLaMa-7B-Merged model was trained until the end of epoch $3$ with validation running every half epoch to prevent over-fitting with a patience of $3$. 

\section{Benchmark Dataset}
After generative model selection, we constructed the novel benchmark dataset for neural/human authorship classification.  
The dataset consists of equal parts human-written and machine-generated news and is evenly stratified across languages. The neural articles were generated by prompting diverse LLMs: the fine-tuned models described in Table \ref{tab:hyperparameters-neural-benchmark}, GPT4, and four additional open-source models. 

\begin{table*}[h!tbp]
    \centering
    \footnotesize
    \resizebox{\textwidth}{!}{%
    \begin{tabular}{c|c|c|c|c|c|c}
        \toprule
         Model & Temperature & Top P & Top K & \shortstack{Repetition \\ Penalty} & Min. Length & Max tokens \\ 
         \midrule
         \shortstack{Fine-tuned \\ Bloomz-3B} &1&0.9&40&1.3&512&2048 \\
         \midrule
         \shortstack{Fine-tuned \\ LLaMa-2-Chat-7B} &1&0.9&40&1.3&512&2048 \\
         \midrule
         Fine-tuned \\ Persian-LLaMa-7B-Merged &1&0.9&40&1.3&512&2048 \\
        \midrule
         Mixtral &1&0.9&40&1.3&512&2048 \\
         \midrule
         GPT-4$\dagger$ &1&1&50&1&-& 4096 \\
         \midrule
         Mistral &1&1&50&1&512&2048 \\
         \midrule
         Bloomz-3B &1&1&50&1.2&512&2048 \\
        \midrule
         LLaMa-2-Chat-7B &1&1&50&1.2&512&2048 \\
         \midrule
         Persian-LLaMa-7B-Merged &1&1&50&1.2&512&2048 \\
         \bottomrule
    \end{tabular}%
    }
    \caption{Hyperparameters of news generation for Benchmark Dataset.  ($\dagger$): No minimum length value could be provided as an argument to GPT-4. }
    \label{tab:hyperparameters-neural-benchmark}
\end{table*}

\subsection{Human-written News Part} \label{subsection:data}
The human-written articles in our dataset stem from open-source datasets \cite{cc-news, batubayk, rohanailab-persian-dataset}, with some additional scraped\footnote{\href{https://www.radiofarda.com/}{https://www.radiofarda.com/}} articles for Persian. \\
The dataset was filtered for topics such as national and international news, politics, science, and economics, using metadata such as tags and source websites. However, outliers may be present.

\subsection{Machine-generated News Part}
We include machine-generated articles acquired from prompting various LLMs. After conducting experiments and considering insights from \cite{bloomz}, a prompt style featuring instruction in English followed by information about the article in the target language emerged as the most effective option. The same prompt from subsection \ref{sec:gen-fintune-data} was constructed for generation fine-tuning (Appendix \ref{sec:app_llms_prompts} for more details). We report hyper-parameters used for generation in Table \ref{tab:hyperparameters-neural-benchmark}.
The prompts are centered around the same topics as human-written news, but no topic analysis was done on the generated articles.  

\paragraph{Fine-tuned BloomZ-3B}
The first half of the neural examples were created by prompting one of our best-performing models from Subsection \ref{subsection:finetuning-bloomz-3b}. 
\paragraph{Fine-tuned LLaMa-2-Chat-7B}
Another major part was generated by our fine-tuned model (Subsection \ref{subsection:finetuning-llama}) for English, Hungarian, and Turkish and from the model (Subsection \ref{subsection:finetuning-llama-persian}) for Persian. 
\paragraph{GPT-4}
To obtain a robust classification benchmark, we included samples from one of the current State-of-the-Art LLMs, GPT-4, which has significantly higher capacity than the other models in this dataset. We utilized OpenAI's paid API to generate $4,000$ news articles from this model.

\paragraph{Zero-Shot prompting of open-source models}
Finally, we included also zero-shot generations of some promising multilingual open-source models: Mistral \cite{mistral}, Mixtral \cite{mixtral}, Bloomz-3B \cite{bloomz}, Persian-LLaMa-7B-Merged \cite{llama2_paper}, and LLaMa-Chat-7B-Merged \cite{llama-persian}.

\subsection{Dataset Splits}
Table \ref{tab:benchmark_data_info} presents the data distribution across models and training, validation, and test splits. 
In all classification setups, training and validation were done solely on our fine-tuned models. Furthermore, discriminators were tested separately for each generative model and language combination. In Appendix~\ref{sec:generated_examples}, we provide various good and bad examples of generations.\\
The length of human-written articles lies between $30$ and $1300$ words, and  <$1$\% of generated articles fall outside this length range. 
The majority of outliers come from Persian generations. See Appendix~\ref{subsec:length-distr} for a detailed overview of length distribution. 

\section{Models for Generated News Detection}
For neural news detection, some classical machine learning benchmarks, as well as the latest deep learning based on Transformer Encoders, were assessed \cite{murphy2013machine, how_to_finetune_bert}.

\subsection{Pre-processing} 
All discriminative models were trained and tested on the splits described in Table \ref{tab:benchmark_data_info}. However, we employed two separate pre-processing styles for linguistic- and transformer-based classifiers.

\paragraph{Soft-Preprocessing} 
As transformers-based models are pre-trained on human written content, we refrained from any drastic pre-processing steps like lemmatization for these classifiers. We fixed encoding issues using the \textit{ftfy} library \cite{speer-2019-ftfy} and removed artifacts and obvious repetitions determined by inspecting the top $200$ TF-IDF features and top-features from a Random Forest Classifier (Subsection \ref{subsubsec:random-forest})~\citep{murphy-etal-2022-decoding}. In some cases, complete elimination of artifacts and repetitions was not possible due to high repetition and limited multilingual capacity.

\begin{table*}[h!tbp]
    \centering
    \scriptsize
    \begin{tabular}{c|c|c|c|c|c|c|c|c|c}
        \toprule
          & \shortstack{Fine-tuned \\ Bloomz-3B} & \shortstack{Fine-tuned \\ LLaMa-2-chat-7B*} & \shortstack{Zero-Shot \\GPT-4} & \shortstack{Zero-Shot \\ Bloomz-3B} & \shortstack{Zero-Shot \\ LLaMa-2-Chat-7B*}  & \shortstack{Zero-Shot \\ Mistral} & \shortstack{Zero-Shot \\ Mixtral} & \shortstack{Human-written \\ News} & Total \\
        \midrule
        Train & 600 $\times$ 4 & 600 $\times$ 4 &- &-&-& - & - & 1200 $\times$ 4 & 9600  \\
        \midrule
        Val & 200 $\times$ 4 & 200 $\times$ 4 &-&-&-& -& - & 400 $\times$ 4 & 3200 \\
        \midrule 
        Test & 200 $\times$ 4 & 200 $\times$ 4 &1000 $\times$ 4 &100 $\times$ 4&100 $\times$ 4& 100 $\times$ 4 & 100 $\times$ 4 & 1800 $\times$ 4 & 14400 \\
        \midrule

        \textbf{Total} & 1000 $\times$ 4 & 1000 $\times$ 4 &1000 $\times$ 4&100 $\times$ 4&100 $\times$ 4& 100 $\times$ 4 & 100 $\times$ 4
        & 3400 $\times$ 4 & 27200 \\
        
        \bottomrule
    \end{tabular}
    \caption{Overview of Benchmark Dataset splits used for classification. The dataset is equally stratified across all four languages. (*): Persian-LLaMa-7B-Merged for Persian.}
    \label{tab:benchmark_data_info}
\end{table*}

\paragraph{Hard-preprocessing} 
To get the most out of the TF-IDF features, our linguistic discriminators were based on a more extensive pre-processing pipeline. 
Building on the soft-pre-processed text, we removed HTML characters, punctuation, and stopwords, converted them to lowercase, masked numbers with a number token in the target language, and lemmatized \cite{spacy,tr-spacy,tr-spacy-model,huspacy,huspacy-lemmatizer,hazm}. 

\subsection{Linguistically Informed Baselines} \label{subsubsec:random-forest}
As a baseline, we assess the linguistically informed classifiers Logistic Regression (LR), Support Vector Machine (SVM), and  Random Forest (RF) \cite{murphy-etal-2022-decoding, sklearn}.

A common behavior of less advanced LLMs is repetition. Therefore, we chose TF-IDF word features of hard-processed data as input to the linguistically informed classifiers. The maximum number of features is capped at $1,000$ in order to keep the feature space small and focus on said repetitions and as this sufficiently captured training data.

\paragraph{Logistic Regression}
We employed a logistic regression model with $L1$-loss, inverse regularization strength(C) of 6.15, and a Library for Large Linear Classification (\textit{liblinear}) solver \cite{sklearn}. 

\paragraph{Support Vector Machines }
Next, we applied a Support Vector Machine. We have chosen a linear kernel, as this has not performed worse than others, and its chosen features are more interpretable. 

\paragraph{Random Forest }
Finally, we utilized a Random Forest Classifier using $100$ estimators and a random state of $42$. We have observed that while the previous two models tended to choose word features that aligned with their TF-IDF scores, Random Forest chose a more diverse set of features and, at times, identified previously overlooked artifacts.
While the above models might not directly take into account semantic information, they can provide insights into machine/human authorship detection, i.e., the identification of artifacts.  

\subsection{Transformer-based Baselines}
In line with the recent approaches to text classification, main multilingual transformer encoders, BERT and RoBERTa, were fine-tuned and assessed for each language separately \cite{bert, roberta}. For these, the patience of $5$ over the validation set was used to end the training early, and the embedding layers were frozen.

\paragraph{BERT} 
We fine-tuned a multilingual instance of BERT\footnote{\href{https://huggingface.co/google-bert/bert-base-multilingual-cased}{https://huggingface.co/google-bert/bert-base-multilingual-cased}}~\cite{bert}---mBERT---from HuggingFace to assess its performance in classifying news from humans vs. LLMs.

In general, \verb|adamw|~\citep{adamw_paper} was used as the optimizer with a learning rate of $2e^{-4}$. Training for English, Turkish, Hungarian, and Persian language settings provided the best validation losses at epochs $4, 6, 6$ and $8$, respectively. 

\paragraph{RoBERTa} The experiments were continued with a multilingual instance of RoBERTa\footnote{\href{https://huggingface.co/FacebookAI/xlm-roberta-base}{https://huggingface.co/FacebookAI/xlm-roberta-base}}~\cite{conneau2019unsupervised}---XLM-R---to see whether the larger vocabulary and extensive Byte-Pair Encoding scheme of RoBERTa would help classification~\citep{jurafsky2013}. 

Similar to the training iterations for BERT, \verb|adamw| was used as the optimizer with the learning rate of $2e^{-4}$ \citep{adamw_paper}. Training for English, Turkish, Hungarian, and Persian settings provided the best validation losses at epochs $3, 2, 3$, and $3$, respectively.  

\begin{table*}[h!tbp]
    \centering
    \footnotesize
    \begin{tabular}{c|c||ccc||ccc||ccc||ccc}
        \toprule
         \shortstack{Generative \\ Model} & \shortstack{Discriminative \\ Model} & Pre & Re & F1 &  Pre & Re & F1 &  Pre & Re & F1 &  Pre & Re & F1 \\ 
        \midrule
        \multicolumn{2}{c||}{ } & \multicolumn{3}{c||}{English} & \multicolumn{3}{c||}{Turkish} & \multicolumn{3}{c||}{Hungarian} & \multicolumn{3}{c}{Persian} \\ 
        \midrule \midrule
        \multicolumn{14}{c}{\textit{\textbf{In-domain test set}: Samples from these generative models were the part of training data for the discriminators.}} \\
        \midrule

        \multirow{3}{*}{\shortstack{Fine-tuned \\ BloomZ-3B}} & LR  & .96 & .96 & .96 & .98 & .98 & .98 & .98 & .98 & .98 & .99 & .99 & .99 \\
                             & RF & .98 & .98 & .98 & .97 & .97 & .97 & .97 & .97 & .97 & \textbf{1.00} & \textbf{1.00} & \textbf{1.00} \\
                             & SVM & .98 & .98 & .98 & .99 & .99 & .99 & .98 & .98 & .98 & .99 & .99 & .99  \\
                             & mBERT & \textbf{.99} & \textbf{.99} & \textbf{.99} & \textbf{1.00} & \textbf{1.00} & \textbf{1.00} & .99 & .99 & .99 & .99 & .99 & .99 \\
                             & XLM-R  & \textbf{.99} & \textbf{.99} & \textbf{.99} & \textbf{1.00} & \textbf{1.00} & \textbf{1.00} & \textbf{1.00} & \textbf{1.00} & \textbf{1.00} & \textbf{1.00} & \textbf{1.00} & \textbf{1.00} \\
        \midrule
        \multirow{3}{*}{\shortstack{Fine-tuned \\ LLaMa-2 \\ Chat-7B \\ or \\ LLaMa-7B \\ Persian}} & LR  & .97 & .97 & .97 & .99 & .99 & .99 & .98 & .98 & .98 & .99 & .99 & .99 \\
                                     & RF & .97 & .97 & .97 & \textbf{1.00} & \textbf{1.00} & \textbf{1.00} & .99 & .99 & .99 & \textbf{1.00} & \textbf{1.00} & \textbf{1.00} \\
                                     & SVM & .98 & .98 & .98 & \textbf{1.00} & \textbf{1.00} & \textbf{1.00} & .99 & .99 & .99 & .99 & .99 & .99 \\
                                     & mBERT & \textbf{.99} & \textbf{.99} & \textbf{.99} & \textbf{1.00} & \textbf{1.00} & \textbf{1.00} & .99 & .99 & .99 & .99 & .99 & .99 \\
                                     & XLM-R & \textbf{.99} & \textbf{.99} & \textbf{.99} & \textbf{1.00} & \textbf{1.00} & \textbf{1.00} & \textbf{1.00} & \textbf{1.00} & \textbf{1.00} & \textbf{1.00} & \textbf{1.00} & \textbf{1.00} \\
        \midrule
        \midrule
        \multicolumn{14}{c}{\textit{\textbf{Out-of-domain test set}: Samples from these generative models were unseen for discriminators.}} \\
        \midrule
        \multirow{3}{*}{\shortstack{Zero-Shot \\ BloomZ-3B}}  & LR  & .70 & .61 & .56 & .50 & .50 & .34 & .75 & .61 & .56 & .82 & .79 & .78 \\
                                     & RF & .70 & .59 & .52 & .25 & .49 & .33  & .68 & .54 & .43 & .99 & .98 & .98 \\
                                     & SVM & .74 & .61 & .55 & .50 & .50 & .35 & .77 & .62 & .57 & .82 & .77 & .75 \\
                                     & mBERT & .77 & .53 & .40 & .77 & .57 & .47 & .81 & .70 & .68 & .96 & .96 & .95 \\
                                     & XLM-R & \textbf{.79} & \textbf{.64} & \textbf{.59} & \textbf{81} & \textbf{.70} & \textbf{.67} & \textbf{98} & \textbf{.98} & \textbf{.98} & \textbf{1.00} & \textbf{1.00} & \textbf{1.00} \\
        \midrule
        \multirow{3}{*}{\shortstack{Zero-Shot \\ LLaMa-2 \\ Chat-7B}} & LR  & .82 & .81 & .80 & .74 & .57 & .49 & .77 & .66 & .61 & .79 & .77 & .76\\
                                     & RF & .85 & .84 & .84 & .75 & .51 & .36 & .68 & .56 & .47 & .99 & .98 & .98 \\
                                     & SVM & .86 & .84 & .84 & .77 & .57 & .48 & .80 & .68 & .64 & .76 & .72 & .72 \\
                                     & mBERT & .88 & .84 & .83 & \textbf{.99} & \textbf{.99} & \textbf{.99} & \textbf{.99} & \textbf{.99} & \textbf{.99} & .90 & .88 & .88 \\
                                     & XLM-R & \textbf{1.00} & \textbf{1.00} & \textbf{1.00} & \textbf{.98} & \textbf{.98} & \textbf{.98} & \textbf{.99} & \textbf{.99} & \textbf{.99} & \textbf{1.00} & \textbf{1.00} & \textbf{1.00} \\                  
        \midrule \midrule
        \multirow{3}{*}{\shortstack{Zero-Shot \\ Mistral}} & LR  & .24 & .46 & .32 & .70 & .53 & .40 & .72 & .60 & .54 & .83 & .79 & .78 \\
                                     & RF & .24 & .48 & .32 & .75 & .51 & .34 & .71 & .54 & .42 & \textbf{1.00} & \textbf{.99} & \textbf{.99} \\
                                     & SVM & .25 & .48 & .33 & .77 & \textbf{.57} & \textbf{.48} & .75 & .61 & .56 & .87 & .83 & .83 \\
                                     & mBERT & \textbf{.76} & \textbf{.55} & \textbf{.44} & .76 & .54 & .42 & .87 & .83 & .83 & .93 & .92 & .92 \\
                                     & XLM-R & \textbf{.76} & \textbf{.54} & .42  & \textbf{.78} & .53 & .40 & \textbf{.93} & \textbf{.92} & \textbf{.93} & .93 & .92 & .92 \\              
        \midrule
        \multirow{3}{*}{\shortstack{Zero-Shot \\ Mixtral}}    & LR  & .72 & .60 & .54 & .75 & \textbf{.59} & \textbf{.52} & \textbf{.76} & \textbf{.54} & \textbf{.42} & .85 & .84 & .84 \\
                                     & RF & .75 &  \textbf{.67} &  \textbf{.64} & \textbf{.77} & .57 & .48 & \textbf{.76} & \textbf{.53} & .40 & \textbf{1.00} & \textbf{1.00} & \textbf{1.00} \\
                                     & SVM & .73 & .61 & .55 & \textbf{.77} & \textbf{.58} & .49 & \textbf{.76} & \textbf{.53} & .39 & .85 & .80 & .79 \\
                                     & mBERT & .75 & .50 & .35 & \textbf{.77} & .56 & .45 & \textbf{.75} & .50 & .34 & .93 & .92 & .92 \\
                                     & XLM-R & \textbf{.76} & .57 & .47  & .75 & .52 & .39 & \textbf{.75} & .50 & .35 & .98 & .98 & .98 \\
        \midrule \midrule
        \multirow{3}{*}{GPT-4} & LR  & .55 & .51 & .41 & .57 & .51 & .37 & .57 & .51 & .37 & .88 & .85 & .85 \\
                                     & RF & .56 & \textbf{.53} & \textbf{.45} & .47 & .50 & .34 & .55 & .50 & .35 & \textbf{.99} & \textbf{.99} & \textbf{.99} \\
                                     & SVM & .59 & \textbf{.53} & .44 & .60 & .51 & .35 & .64 & .51 & .35 & .86 & .82 & .82 \\
                                     & mBERT & .44 & .49 & .36 & \textbf{.65} & \textbf{.52} & .38 & \textbf{.88} & \textbf{.85} & \textbf{.85} & \textbf{.99} & \textbf{.99} & \textbf{.99} \\
                                     & XLM-R & \textbf{.75} & .50 & .34 & .59 & \textbf{.52} & \textbf{.42} & .25 & .50 & .33 & .89 & .85 & .85\\
        \midrule \midrule
        \multicolumn{14}{c}{\textit{\textbf{Test subset for zero-shot prompting} of out-of-the-box LLMs.}} \\
        \midrule
        \multirow{3}{*}{\shortstack{Fine-tuned \\ BloomZ-3B}}  & GPT4 & \textbf{.93} & \textbf{.93} & \textbf{.92} & .98 & .98 & .98 & .88 & .83 & .83 & .98 & .98 & .98 \\
                                     &  Mistral & .90 & .87 & .87  & .81 & .69 & .65 & \textbf{.99} & \textbf{.99} & \textbf{.99} & \textbf{1.00} & \textbf{1.00} & \textbf{1.00}  \\
                                     & LLaMa-2 & .90 & .87 & .87 & \textbf{1.00} & \textbf{1.00} & \textbf{1.00} & \textbf{1.00} & \textbf{1.00} & \textbf{1.00} & \textbf{1.00} & \textbf{1.00} & \textbf{1.00} \\
        \midrule
        \multirow{3}{*}{\shortstack{Fine-tuned \\ LLaMa-2 \\ Chat-7B}} & GPT4 & .93 & .92 & .92 & .97 & .96 & .96 & .88 & .83 & .83 & .95 & .95 & .95 \\
                                     &  Mistral & \textbf{1.00} & \textbf{1.00} & \textbf{1.00}  & .91 & .88 & .88 & .97 & .97 & .97 & \textbf{1.00} & \textbf{1.00} & \textbf{1.00}  \\
                                     & LLaMa-2 & .98 & .98 & .98 & \textbf{1.00} & \textbf{1.00} & \textbf{1.00} & \textbf{1.00} & \textbf{1.00} & \textbf{1.00} & \textbf{1.00} & \textbf{1.00} & \textbf{1.00} \\
                                    
        \midrule
        \multirow{3}{*}{GPT4}  & GPT4 & .93 & .93 & .92 & .79 & .73 & .72 & .86 & .82 & .82 & .90 & .88 & .87 \\
                                     &  Mistral & \textbf{.98} & \textbf{.98} & \textbf{.98}  & .90  & .88 & .87 & .88 & .84 & .84 & \textbf{1.00} & \textbf{1.00} & \textbf{1.00} \\
                                     & LLaMa-2 & .97 & .96 & .96 & \textbf{1.00} & \textbf{1.00} & \textbf{1.00} & \textbf{1.00} & \textbf{1.00} & \textbf{1.00} & \textbf{1.00} & \textbf{1.00} & \textbf{1.00} \\        
        
        \bottomrule
        
    \end{tabular}
    \caption{Machine-generated news detection result per generator, per discriminator, and per language. We experimented with several setups: (i)~\textbf{in-domain} generated samples to confirm that discriminators achieved the highest performance after fine-tuning for all languages; (ii)~\textbf{out-of-domain} generated data from other generation setups checking the sustainability of the discriminators; (iii)~\textbf{test subset for zero-shot prompting} of out-of-the-box LLMs to check their ability to detect as well LLMs-generated texts. In \textbf{bold}, we highlight the best classification results within the block per generator and per language.}
    \label{tab:final_results}
\end{table*}

\subsection{LLMs Prompt-based Classification}
Finally, we prompted BloomZ, LLaMa-2, and GPT-4 \textit{out-of-the-box} on our task. As such prompting can be quite resource-consuming, we sub-sample from our whole test set $300$ machine-generated and $300$ human-written texts. The prompt template looks like the following (see Appendix~\ref{sec:app_llms_prompts_classification}):

\begin{quote}
\textsf{\footnotesize Considering the content and context of the text, decide whether it was written by a human or artificial intelligence. If written by a human, write [HUMAN]; otherwise, write [ARTIFICIAL INTELLIGENCE]. For example:
Text:"..."
Classification:[ARTIFICIAL INTELLIGENCE]
}
\end{quote}

\section{Results}
All the results are presented in Table~\ref{tab:final_results}. The setups include (i)~using in-domain generated samples to verify that discriminators achieved optimal performance after fine-tuning across all languages, (ii)~employing out-of-domain generated data from the zero-shot generations to assess the sustainability of the discriminators, and (iii)~utilizing a test subset for zero-shot prompting of out-of-the-box LLMs to evaluate their capability to detect generations by LLMs. 
From in-domain test set results, we can confirm that, indeed, the classifiers were perfectly able to learn the task of detecting machine-generated text across the fine-tuned generators and all languages. Almost perfect results for all detectors are achieved on Turkish and Persian data. For English and Hungarian, Transformer-based classifiers keep the bar for the highest results. We purposefully overfitted the classifiers on the in-domain data to have them generalize better to out-of-domain data. 

The out-of-domain data put stress on the classifiers' robustness. For the zero-shot prompted BloomZ-3B and LLaMa-based models, mBERT and XLM-RoBERTa gain the top places, however, with a drop in the performance. Notably, XLM-R demonstrates near-perfect scores for Persian and Hungarian in BloomZ-3B, as well as across all languages in LLaMa-based models. At the same time, BloomZ-3B for English and Turkish models showed a significant decline in classification results.

Furthermore, we continued even further out-of-the-box experiments on texts generated by Mistral, Mixtral, and GPT-4. For these models, a common trend emerges in the results: all classifiers show high detection scores for the Persian language, possibly due to the shorter length of texts in this language, but with a significant drop in the score for all other languages. However, the Random Forest classifier consistently produces the top results. In other languages, linguistically informed detectors also demonstrate more resilient performance. Notably, for GPT-4 in English, Random Forest once again achieves the highest F1 score. These findings suggest a hypothesis that linguistically informed detectors for the \textit{out-of-domain} task may offer the most promising, robust, and potentially explainable models compared to the black-box Transformer-based classifiers.

In the final phase, we experimented on a smaller test set, prompting LLMs to detect the generated texts from LLMs. These initial findings yield quite encouraging implications.  Thus, out-of-the-box LLaMa-2 can nearly perfectly detect generated texts in all languages from our LLaMa-2-Chat-7b instance fine-tuned for the downstream task and even GPT-4. Similarly, Mistral achieves comparable results, albeit with a slight decrease in performance for Turkish and Hungarian. At the same time, GPT-4 shows the best performance at the detection of English BloomZ-3B-generated data but shows flaws in the detection of its own generated texts. Thus, LLMs indeed can be more robust detectors, requiring significantly greater resources compared to linguistic classifiers. This potentially means that mono-lingual or multi-lingual models may struggle to learn to write like a newsagent during fine-tuning unless the model has encountered the requested news language during pre-training.

\section{Conclusion}
In this work, we presented the first-of-its-kind benchmark dataset for machine-generated news detection covering four languages---English, Turkish, Hungarian, and Persian.

To gain machine-generated data, we assessed the most promising LLMs' ability of news generation for the target languages. Our final choice was to explore BloomZ-3B, LLaMa-2-Chat-7B, Persian-LLaMa-7B-Merged, Mistral, Mixtral, and GPT-4 in zero-shot setups as well as additionally fine-tune BloomZ-3B and LLaMa based models for the downstream generation task. We also used open-source datasets and a small portion of scraped news instances to gain human-written data. In the end, we use the generated news from fine-tuned BloomZ-3B and LLaMa-based instances to construct the training split for classifiers and then utilize samples from other generators for out-of-domain experiments.

Then, based on our dataset, we trained and tested diverse classifiers---linguistically informed baselines like Logistic Regression,  Random Forest, and Support Vector Machine were compared with mBERT and XLM-R. Among these models, XLM-R showed the highest results on the in-domain data but dropped performance in several out-of-domain experiments. At the same time, linguistically informed classifiers illustrated better robustness. Finally, we also experimented with LLMs to detect LLM-generated news. LLaMa-2 models showed quite promising performance for all languages, even with the ability to detect GPT-4 generated texts. 

We can conclude that while baselines can be fine-tuned for the multilingual-generated news detection task, they continue to suffer in detecting the nuances in out-of-domain data to a degree.  However, such models may offer greater explainability. LLMs can prove exceedingly robust for the task, although the resource requirements for these models are higher. 

\section*{Limitations}
 Although we strongly believe that our presented dataset and baseline benchmarking results will bring a new foundation for strong multilingual neural news detection systems, it is impossible to address all research questions in one work. However, we hope newly arisen research questions will build the base for new research.

Firstly, in this work, we only tackle neural authorship detection and do not consider the truthfulness of the generated information. This problem is going to be the next step in a fake news detection pipeline. The generative models can hallucinate about nonexistent events and fabricate facts~\cite{kasneci2023chatgpt,DBLP:journals/corr/abs-2401-01313}. In this particular work, this was not in the scope of our research, and we believe that our benchmark dataset can be used in future work to inspect trustworthiness across multiple languages and also in prompt engineering, i.e., to assess how the training and generation pipeline respond to various prompt engineering techniques.

Secondly, we can also foresee possibilities for further research in discriminative model explainability. We conducted mostly black-box experiments revealing the answer to the first upcoming research question of overall detection of neural texts power of the classifiers. More linguistics-based features, such as emotions, style, and part-of-speech usage, can be inspected to uncover the language differences between machine-generated and human-written texts. We are firmly convinced that the public availability of our created benchmark dataset will create new directions for the linguistic side of neural text detection research.
Finally, we reported the cross-domain abilities of the classification models but did not inspect cross-lingual neural news detection power. This direction opens a new chapter of extensive experiments. Indeed, it would be extremely beneficial to find out if the knowledge of neural text detection can be propagated to new languages, 
avoiding spending resources on training data collection. We leave this question as well for future work.

\section*{Ethics Statement}
Our research addresses the increasingly pertinent issue of ensuring the safety and reliability of AI---detection of machine-generated texts. As we (will) publicly release all the generated data and fine-tuned models for both generation and classification, we aim to bolster the resilience of detection and prevention solutions against harmful information.

A potential concern may arise regarding the publication of our report, which meticulously outlines pipelines and hyperparameters for the development of robust neural news generation models, as it could potentially enable misuse of this information. 
Nonetheless, our initial experiments reveal remarkably encouraging outcomes of open-source models in detecting generated news from both closed and open-source generators. 
With this, we hold the belief that, instead, our open-source data and models will pave the way for even more robust solutions, enabling the creation of detection mechanisms capable of withstanding emerging generative models in multiple languages.

\bibliography{anthology,custom}

\begin{thebibliography}{48}
\expandafter\ifx\csname natexlab\endcsname\relax\def\natexlab#1{#1}\fi

\bibitem[{Abdalla et~al.(2023)Abdalla, Malberg, Dementieva, Mosca, and Groh}]{DBLP:journals/information/AbdallaMDMG23}
Mohamed Hesham~Ibrahim Abdalla, Simon Malberg, Daryna Dementieva, Edoardo Mosca, and Georg Groh. 2023.
\newblock \href {https://doi.org/10.3390/INFO14100522} {A benchmark dataset to distinguish human-written and machine-generated scientific papers}.
\newblock \emph{Inf.}, 14(10):522.

\bibitem[{Achiam et~al.(2023)Achiam, Adler, Agarwal, Ahmad, Akkaya, Aleman, Almeida, Altenschmidt, Altman, Anadkat et~al.}]{achiam2023gpt}
Josh Achiam, Steven Adler, Sandhini Agarwal, Lama Ahmad, Ilge Akkaya, Florencia~Leoni Aleman, Diogo Almeida, Janko Altenschmidt, Sam Altman, Shyamal Anadkat, et~al. 2023.
\newblock Gpt-4 technical report.
\newblock \emph{arXiv preprint arXiv:2303.08774}.

\bibitem[{Alimoradi(2021)}]{rohanailab-persian-dataset}
Saied Alimoradi. 2021.
\newblock Hugging face: Rohanailab persian dataset.
\newblock \url{https://huggingface.co/datasets/RohanAiLab/persian_news_dataset}, note = {Accessed: 2024-03-15}.

\bibitem[{Altinok(2023)}]{tr-spacy}
Duygu Altinok. 2023.
\newblock \href {https://aclanthology.org/2023.acl-long.768} {A diverse set of freely available linguistic resources for {T}urkish}.
\newblock In \emph{Proceedings of the 61st Annual Meeting of the Association for Computational Linguistics (Volume 1: Long Papers)}, pages 13739--13750, Toronto, Canada. Association for Computational Linguistics.

\bibitem[{Amiri(2023)}]{llama-persian}
Mostafa Amiri. 2023.
\newblock Hugging face: Persian llama 7b merged.
\newblock \url{https://huggingface.co/mostafaamiri/persian_llama_7B_merged}.
\newblock Accessed: 2024-03-15.

\bibitem[{Baykara and Güngör(2022)}]{batubayk}
Batuhan Baykara and Tunga Güngör. 2022.
\newblock \href {https://doi.org/10.1007/s10579-021-09568-y} {{Abstractive text summarization and new large-scale datasets for agglutinative languages Turkish and Hungarian}}.
\newblock \emph{Language Resources and Evaluation}, pages 1--35.

\bibitem[{{Berkecz, P{\'e}ter and Orosz, Gy{\"o}rgy and Sz{\'a}nt{\'o}, Zsolt and Szab{\'o}, Gerg{\H{o}} and Farkas, Rich{\'a}rd}(2023)}]{huspacy-lemmatizer}
{Berkecz, P{\'e}ter and Orosz, Gy{\"o}rgy and Sz{\'a}nt{\'o}, Zsolt and Szab{\'o}, Gerg{\H{o}} and Farkas, Rich{\'a}rd}. 2023.
\newblock {Hybrid lemmatization in HuSpaCy}.
\newblock In \emph{{XIX. Magyar Sz{\'a}m{\'\i}t{\'o}g{\'e}pes Nyelv{\'e}szeti Konferencia}}.

\bibitem[{Conneau et~al.(2019)Conneau, Khandelwal, Goyal, Chaudhary, Wenzek, Guzm{\'a}n, Grave, Ott, Zettlemoyer, and Stoyanov}]{conneau2019unsupervised}
Alexis Conneau, Kartikay Khandelwal, Naman Goyal, Vishrav Chaudhary, Guillaume Wenzek, Francisco Guzm{\'a}n, Edouard Grave, Myle Ott, Luke Zettlemoyer, and Veselin Stoyanov. 2019.
\newblock Unsupervised cross-lingual representation learning at scale.
\newblock \emph{arXiv preprint arXiv:1911.02116}.

\bibitem[{Conover et~al.(2023)Conover, Hayes, Mathur, Xie, Wan, Shah, Ghodsi, Wendell, Zaharia, and Xin}]{DatabricksBlog2023DollyV2}
Mike Conover, Matt Hayes, Ankit Mathur, Jianwei Xie, Jun Wan, Sam Shah, Ali Ghodsi, Patrick Wendell, Matei Zaharia, and Reynold Xin. 2023.
\newblock \href {https://www.databricks.com/blog/2023/04/12/dolly-first-open-commercially-viable-instruction-tuned-llm} {Free dolly: Introducing the world's first truly open instruction-tuned llm}.

\bibitem[{Devlin et~al.(2019)Devlin, Chang, Lee, and Toutanova}]{bert}
Jacob Devlin, Ming-Wei Chang, Kenton Lee, and Kristina Toutanova. 2019.
\newblock \href {http://arxiv.org/abs/1810.04805} {Bert: Pre-training of deep bidirectional transformers for language understanding}.

\bibitem[{Egonmwan and Chali(2019)}]{DBLP:conf/emnlp/EgonmwanC19}
Elozino Egonmwan and Yllias Chali. 2019.
\newblock \href {https://doi.org/10.18653/V1/D19-5607} {Transformer-based model for single documents neural summarization}.
\newblock In \emph{Proceedings of the 3rd Workshop on Neural Generation and Translation@EMNLP-IJCNLP 2019, Hong Kong, November 4, 2019}, pages 70--79. Association for Computational Linguistics.

\bibitem[{Gu et~al.(2020)Gu, Mao, Han, Liu, Wu, Yu, Finnie, Yu, Zhai, and Zukoski}]{DBLP:conf/www/GuM0L00FYZZ20}
Xiaotao Gu, Yuning Mao, Jiawei Han, Jialu Liu, You Wu, Cong Yu, Daniel Finnie, Hongkun Yu, Jiaqi Zhai, and Nicholas Zukoski. 2020.
\newblock \href {https://doi.org/10.1145/3366423.3380247} {Generating representative headlines for news stories}.
\newblock In \emph{{WWW} '20: The Web Conference 2020, Taipei, Taiwan, April 20-24, 2020}, pages 1773--1784. {ACM} / {IW3C2}.

\bibitem[{Guo et~al.(2023)Guo, Zhang, Wang, Jiang, Nie, Ding, Yue, and Wu}]{DBLP:journals/corr/abs-2301-07597}
Biyang Guo, Xin Zhang, Ziyuan Wang, Minqi Jiang, Jinran Nie, Yuxuan Ding, Jianwei Yue, and Yupeng Wu. 2023.
\newblock \href {https://doi.org/10.48550/ARXIV.2301.07597} {How close is chatgpt to human experts? comparison corpus, evaluation, and detection}.
\newblock \emph{CoRR}, abs/2301.07597.

\bibitem[{Hamborg et~al.(2017)Hamborg, Meuschke, Breitinger, and Gipp}]{cc-news}
Felix Hamborg, Norman Meuschke, Corinna Breitinger, and Bela Gipp. 2017.
\newblock \href {https://doi.org/10.5281/zenodo.4120316} {news-please: A generic news crawler and extractor}.
\newblock In \emph{Proceedings of the 15th International Symposium of Information Science}, pages 218--223.

\bibitem[{Honnibal et~al.(2020)Honnibal, Montani, Van~Landeghem, and Boyd}]{spacy}
Matthew Honnibal, Ines Montani, Sofie Van~Landeghem, and Adriane Boyd. 2020.
\newblock \href {https://doi.org/10.5281/zenodo.1212303} {{spaCy: Industrial-strength Natural Language Processing in Python}}.
\newblock \url{https://spacy.io}.

\bibitem[{Jiang et~al.(2023)Jiang, Sablayrolles, Mensch, Bamford, Chaplot, de~las Casas, Bressand, Lengyel, Lample, Saulnier, Lavaud, Lachaux, Stock, Scao, Lavril, Wang, Lacroix, and Sayed}]{mistral}
Albert~Q. Jiang, Alexandre Sablayrolles, Arthur Mensch, Chris Bamford, Devendra~Singh Chaplot, Diego de~las Casas, Florian Bressand, Gianna Lengyel, Guillaume Lample, Lucile Saulnier, Lélio~Renard Lavaud, Marie-Anne Lachaux, Pierre Stock, Teven~Le Scao, Thibaut Lavril, Thomas Wang, Timothée Lacroix, and William~El Sayed. 2023.
\newblock \href {http://arxiv.org/abs/2310.06825} {Mistral 7b}.

\bibitem[{Jiang et~al.(2024)Jiang, Sablayrolles, Roux, Mensch, Savary, Bamford, Chaplot, de~las Casas, Hanna, Bressand, Lengyel, Bour, Lample, Lavaud, Saulnier, Lachaux, Stock, Subramanian, Yang, Antoniak, Scao, Gervet, Lavril, Wang, Lacroix, and Sayed}]{mixtral}
Albert~Q. Jiang, Alexandre Sablayrolles, Antoine Roux, Arthur Mensch, Blanche Savary, Chris Bamford, Devendra~Singh Chaplot, Diego de~las Casas, Emma~Bou Hanna, Florian Bressand, Gianna Lengyel, Guillaume Bour, Guillaume Lample, Lélio~Renard Lavaud, Lucile Saulnier, Marie-Anne Lachaux, Pierre Stock, Sandeep Subramanian, Sophia Yang, Szymon Antoniak, Teven~Le Scao, Théophile Gervet, Thibaut Lavril, Thomas Wang, Timothée Lacroix, and William~El Sayed. 2024.
\newblock \href {http://arxiv.org/abs/2401.04088} {Mixtral of experts}.

\bibitem[{Kasneci et~al.(2023)Kasneci, Se{\ss}ler, K{\"u}chemann, Bannert, Dementieva, Fischer, Gasser, Groh, G{\"u}nnemann, H{\"u}llermeier et~al.}]{kasneci2023chatgpt}
Enkelejda Kasneci, Kathrin Se{\ss}ler, Stefan K{\"u}chemann, Maria Bannert, Daryna Dementieva, Frank Fischer, Urs Gasser, Georg Groh, Stephan G{\"u}nnemann, Eyke H{\"u}llermeier, et~al. 2023.
\newblock Chatgpt for good? on opportunities and challenges of large language models for education.
\newblock \emph{Learning and individual differences}, 103:102274.

\bibitem[{Kokabi(2024)}]{hazm}
Ayub Kokabi. 2024.
\newblock Hazm - persian nlp toolkit.
\newblock \url{https://github.com/roshan-research/hazm}.

\bibitem[{Li et~al.(2023)Li, Li, Cui, Bi, Wang, Yang, Shi, and Zhang}]{DBLP:journals/corr/abs-2305-13242}
Yafu Li, Qintong Li, Leyang Cui, Wei Bi, Longyue Wang, Linyi Yang, Shuming Shi, and Yue Zhang. 2023.
\newblock \href {https://doi.org/10.48550/arXiv.2305.13242} {Deepfake text detection in the wild}.
\newblock \emph{CoRR}, abs/2305.13242.

\bibitem[{Liu et~al.(2019)Liu, Ott, Goyal, Du, Joshi, Chen, Levy, Lewis, Zettlemoyer, and Stoyanov}]{roberta}
Yinhan Liu, Myle Ott, Naman Goyal, Jingfei Du, Mandar Joshi, Danqi Chen, Omer Levy, Mike Lewis, Luke Zettlemoyer, and Veselin Stoyanov. 2019.
\newblock \href {http://arxiv.org/abs/1907.11692} {Roberta: A robustly optimized bert pretraining approach}.

\bibitem[{Loshchilov and Hutter(2019)}]{adamw_paper}
Ilya Loshchilov and Frank Hutter. 2019.
\newblock \href {http://arxiv.org/abs/1711.05101} {Decoupled weight decay regularization}.

\bibitem[{Maronikolakis et~al.(2020)Maronikolakis, Schutze, and Stevenson}]{maronikolakis2020identifying}
Antonis Maronikolakis, Hinrich Schutze, and Mark Stevenson. 2020.
\newblock Identifying automatically generated headlines using transformers.
\newblock \emph{arXiv preprint arXiv:2009.13375}.

\bibitem[{Muennighoff et~al.(2023)Muennighoff, Wang, Sutawika, Roberts, Biderman, Scao, Bari, Shen, Yong, Schoelkopf, Tang, Radev, Aji, Almubarak, Albanie, Alyafeai, Webson, Raff, and Raffel}]{bloomz}
Niklas Muennighoff, Thomas Wang, Lintang Sutawika, Adam Roberts, Stella Biderman, Teven~Le Scao, M~Saiful Bari, Sheng Shen, Zheng-Xin Yong, Hailey Schoelkopf, Xiangru Tang, Dragomir Radev, Alham~Fikri Aji, Khalid Almubarak, Samuel Albanie, Zaid Alyafeai, Albert Webson, Edward Raff, and Colin Raffel. 2023.
\newblock \href {http://arxiv.org/abs/2211.01786} {Crosslingual generalization through multitask finetuning}.

\bibitem[{Murao et~al.(2019)Murao, Kobayashi, Kobayashi, Yatsuka, Masuyama, Higurashi, and Tabuchi}]{murao-etal-2019-case}
Kazuma Murao, Ken Kobayashi, Hayato Kobayashi, Taichi Yatsuka, Takeshi Masuyama, Tatsuru Higurashi, and Yoshimune Tabuchi. 2019.
\newblock \href {https://doi.org/10.18653/v1/N19-2010} {A case study on neural headline generation for editing support}.
\newblock In \emph{Proceedings of the 2019 Conference of the North {A}merican Chapter of the Association for Computational Linguistics: Human Language Technologies, Volume 2 (Industry Papers)}, pages 73--82, Minneapolis, Minnesota. Association for Computational Linguistics.

\bibitem[{Murphy et~al.(2022)Murphy, Bohnet, McDonald, and Noppeney}]{murphy-etal-2022-decoding}
Alex Murphy, Bernd Bohnet, Ryan McDonald, and Uta Noppeney. 2022.
\newblock \href {https://doi.org/10.18653/v1/2022.acl-long.156} {Decoding part-of-speech from human {EEG} signals}.
\newblock In \emph{Proceedings of the 60th Annual Meeting of the Association for Computational Linguistics (Volume 1: Long Papers)}, pages 2201--2210, Dublin, Ireland. Association for Computational Linguistics.

\bibitem[{Murphy(2013)}]{murphy2013machine}
Kevin~P. Murphy. 2013.
\newblock \href {https://www.amazon.com/Machine-Learning-Probabilistic-Perspective-Computation/dp/0262018020/ref=sr_1_2?ie=UTF8&qid=1336857747&sr=8-2} {\emph{Machine learning : a probabilistic perspective}}.
\newblock MIT Press, Cambridge, Mass. [u.a.].

\bibitem[{"Orosz et~al.(2023)"Orosz, Szab{\'o}, Berkecz, Sz{\'a}nt{\'o}, and Farkas}]{huspacy}
Gy{\"o}rgy "Orosz, Gerg{\H{o}} Szab{\'o}, P{\'e}ter Berkecz, Zsolt Sz{\'a}nt{\'o}, and Rich{\'a}rd" Farkas. 2023.
\newblock {"Advancing Hungarian Text Processing with HuSpaCy: Efficient and Accurate NLP Pipelines"}.
\newblock In \emph{{"Text, Speech, and Dialogue"}}, pages 58--69, {"Cham"}. {"Springer Nature Switzerland"}.

\bibitem[{Pedregosa et~al.(2011)Pedregosa, Varoquaux, Gramfort, Michel, Thirion, Grisel, Blondel, Prettenhofer, Weiss, Dubourg, Vanderplas, Passos, Cournapeau, Brucher, Perrot, and Duchesnay}]{sklearn}
F.~Pedregosa, G.~Varoquaux, A.~Gramfort, V.~Michel, B.~Thirion, O.~Grisel, M.~Blondel, P.~Prettenhofer, R.~Weiss, V.~Dubourg, J.~Vanderplas, A.~Passos, D.~Cournapeau, M.~Brucher, M.~Perrot, and E.~Duchesnay. 2011.
\newblock Scikit-learn: Machine learning in {P}ython.
\newblock \emph{Journal of Machine Learning Research}, 12:2825--2830.

\bibitem[{Piantadosi(2014)}]{zipf}
Steven~T. Piantadosi. 2014.
\newblock \href {https://doi.org/10.3758/s13423-014-0585-6} {Zipf’s word frequency law in natural language : a critical review and future directions}.
\newblock \emph{Psychonomic bulletin \& review}, 21(5):1112--1130.

\bibitem[{Radford et~al.(2019)Radford, Wu, Child, Luan, Amodei, Sutskever et~al.}]{radford2019language}
Alec Radford, Jeffrey Wu, Rewon Child, David Luan, Dario Amodei, Ilya Sutskever, et~al. 2019.
\newblock Language models are unsupervised multitask learners.
\newblock \emph{OpenAI blog}, 1(8):9.

\bibitem[{Sethi et~al.(2017)Sethi, Sonawane, Khanwalker, and Keskar}]{sethi2017automatic}
Prakhar Sethi, Sameer Sonawane, Saumitra Khanwalker, and Ravindra~B Keskar. 2017.
\newblock Automatic text summarization of news articles.
\newblock In \emph{2017 International Conference on Big Data, IoT and Data Science (BID)}, pages 23--29. IEEE.

\bibitem[{Shamardina et~al.(2022)Shamardina, Mikhailov, Cherniavskii, Fenogenova, Saidov, Valeeva, Shavrina, Smurov, Tutubalina, and Artemova}]{DBLP:journals/corr/abs-2206-01583}
Tatiana Shamardina, Vladislav Mikhailov, Daniil Cherniavskii, Alena Fenogenova, Marat Saidov, Anastasiya Valeeva, Tatiana Shavrina, Ivan Smurov, Elena Tutubalina, and Ekaterina Artemova. 2022.
\newblock \href {https://doi.org/10.48550/ARXIV.2206.01583} {Findings of the the ruatd shared task 2022 on artificial text detection in russian}.
\newblock \emph{CoRR}, abs/2206.01583.

\bibitem[{Shazeer and Stern(2018)}]{DBLP:adafactor}
Noam Shazeer and Mitchell Stern. 2018.
\newblock \href {http://arxiv.org/abs/1804.04235} {Adafactor: Adaptive learning rates with sublinear memory cost}.
\newblock \emph{CoRR}, abs/1804.04235.

\bibitem[{Shu et~al.(2021)Shu, Li, Ding, and Liu}]{DBLP:conf/aaai/ShuLDL21}
Kai Shu, Yichuan Li, Kaize Ding, and Huan Liu. 2021.
\newblock \href {https://doi.org/10.1609/AAAI.V35I15.17629} {Fact-enhanced synthetic news generation}.
\newblock In \emph{Thirty-Fifth {AAAI} Conference on Artificial Intelligence, {AAAI} 2021, Thirty-Third Conference on Innovative Applications of Artificial Intelligence, {IAAI} 2021, The Eleventh Symposium on Educational Advances in Artificial Intelligence, {EAAI} 2021, Virtual Event, February 2-9, 2021}, pages 13825--13833. {AAAI} Press.

\bibitem[{Speer(2019)}]{speer-2019-ftfy}
Robyn Speer. 2019.
\newblock \href {https://doi.org/10.5281/zenodo.2591652} {ftfy}.
\newblock Zenodo.
\newblock Version 5.5.

\bibitem[{Stribling et~al.(2005)Stribling, Krohn, and Aguayo}]{stribling2005scigen}
Jeremy Stribling, Max Krohn, and Dan Aguayo. 2005.
\newblock Scigen - an automatic cs paper generator.
\newblock \url{https://pdos.csail.mit.edu/archive/scigen/}.

\bibitem[{Sun et~al.(2019)Sun, Qiu, Xu, and Huang}]{how_to_finetune_bert}
Chi Sun, Xipeng Qiu, Yige Xu, and Xuanjing Huang. 2019.
\newblock \href {http://arxiv.org/abs/1905.05583} {How to fine-tune {BERT} for text classification?}
\newblock \emph{CoRR}, abs/1905.05583.

\bibitem[{Taori et~al.(2023)Taori, Gulrajani, Zhang, Dubois, Li, Guestrin, Liang, and Hashimoto}]{alpaca}
Rohan Taori, Ishaan Gulrajani, Tianyi Zhang, Yann Dubois, Xuechen Li, Carlos Guestrin, Percy Liang, and Tatsunori~B. Hashimoto. 2023.
\newblock Stanford alpaca: An instruction-following llama model.
\newblock \url{https://github.com/tatsu-lab/stanford_alpaca}.

\bibitem[{Taylor et~al.(2022)Taylor, Kardas, Cucurull, Scialom, Hartshorn, Saravia, Poulton, Kerkez, and Stojnic}]{taylor2022galactica}
Ross Taylor, Marcin Kardas, Guillem Cucurull, Thomas Scialom, Anthony Hartshorn, Elvis Saravia, Andrew Poulton, Viktor Kerkez, and Robert Stojnic. 2022.
\newblock Galactica: A large language model for science.
\newblock \emph{arXiv preprint arXiv:2211.09085}.

\bibitem[{Tonmoy et~al.(2024)Tonmoy, Zaman, Jain, Rani, Rawte, Chadha, and Das}]{DBLP:journals/corr/abs-2401-01313}
S.~M. Towhidul~Islam Tonmoy, S.~M.~Mehedi Zaman, Vinija Jain, Anku Rani, Vipula Rawte, Aman Chadha, and Amitava Das. 2024.
\newblock \href {https://doi.org/10.48550/ARXIV.2401.01313} {A comprehensive survey of hallucination mitigation techniques in large language models}.
\newblock \emph{CoRR}, abs/2401.01313.

\bibitem[{Touvron et~al.(2023)Touvron, Martin, Stone, Albert, Almahairi, Babaei, Bashlykov, Batra, Bhargava, Bhosale, Bikel, Blecher, Ferrer, Chen, Cucurull, Esiobu, Fernandes, Fu, Fu, Fuller, Gao, Goswami, Goyal, Hartshorn, Hosseini, Hou, Inan, Kardas, Kerkez, Khabsa, Kloumann, Korenev, Koura, Lachaux, Lavril, Lee, Liskovich, Lu, Mao, Martinet, Mihaylov, Mishra, Molybog, Nie, Poulton, Reizenstein, Rungta, Saladi, Schelten, Silva, Smith, Subramanian, Tan, Tang, Taylor, Williams, Kuan, Xu, Yan, Zarov, Zhang, Fan, Kambadur, Narang, Rodriguez, Stojnic, Edunov, and Scialom}]{llama2_paper}
Hugo Touvron, Louis Martin, Kevin Stone, Peter Albert, Amjad Almahairi, Yasmine Babaei, Nikolay Bashlykov, Soumya Batra, Prajjwal Bhargava, Shruti Bhosale, Dan Bikel, Lukas Blecher, Cristian~Canton Ferrer, Moya Chen, Guillem Cucurull, David Esiobu, Jude Fernandes, Jeremy Fu, Wenyin Fu, Brian Fuller, Cynthia Gao, Vedanuj Goswami, Naman Goyal, Anthony Hartshorn, Saghar Hosseini, Rui Hou, Hakan Inan, Marcin Kardas, Viktor Kerkez, Madian Khabsa, Isabel Kloumann, Artem Korenev, Punit~Singh Koura, Marie-Anne Lachaux, Thibaut Lavril, Jenya Lee, Diana Liskovich, Yinghai Lu, Yuning Mao, Xavier Martinet, Todor Mihaylov, Pushkar Mishra, Igor Molybog, Yixin Nie, Andrew Poulton, Jeremy Reizenstein, Rashi Rungta, Kalyan Saladi, Alan Schelten, Ruan Silva, Eric~Michael Smith, Ranjan Subramanian, Xiaoqing~Ellen Tan, Binh Tang, Ross Taylor, Adina Williams, Jian~Xiang Kuan, Puxin Xu, Zheng Yan, Iliyan Zarov, Yuchen Zhang, Angela Fan, Melanie Kambadur, Sharan Narang, Aurelien Rodriguez, Robert Stojnic, Sergey Edunov, and Thomas
  Scialom. 2023.
\newblock \href {http://arxiv.org/abs/2307.09288} {Llama 2: Open foundation and fine-tuned chat models}.

\bibitem[{{Turkish NLP Suite}(2022)}]{tr-spacy-model}
{Turkish NLP Suite}. 2022.
\newblock \href {https://doi.org/10.57967/hf/0086} {tr\_core\_news\_trf (revision c7e06a7)}.

\bibitem[{Wang et~al.(2023)Wang, Mansurov, Ivanov, Su, Shelmanov, Tsvigun, Whitehouse, Afzal, Mahmoud, Aji, and Nakov}]{DBLP:journals/corr/abs-2305-14902}
Yuxia Wang, Jonibek Mansurov, Petar Ivanov, Jinyan Su, Artem Shelmanov, Akim Tsvigun, Chenxi Whitehouse, Osama~Mohammed Afzal, Tarek Mahmoud, Alham~Fikri Aji, and Preslav Nakov. 2023.
\newblock \href {https://doi.org/10.48550/ARXIV.2305.14902} {{M4:} multi-generator, multi-domain, and multi-lingual black-box machine-generated text detection}.
\newblock \emph{CoRR}, abs/2305.14902.

\bibitem[{Warstadt et~al.(2018)Warstadt, Singh, and Bowman}]{Cola_citation}
Alex Warstadt, Amanpreet Singh, and Samuel~R Bowman. 2018.
\newblock Neural network acceptability judgments.
\newblock \emph{arXiv preprint arXiv:1805.12471}.

\bibitem[{Zellers et~al.(2019)Zellers, Holtzman, Rashkin, Bisk, Farhadi, Roesner, and Choi}]{DBLP:conf/nips/ZellersHRBFRC19}
Rowan Zellers, Ari Holtzman, Hannah Rashkin, Yonatan Bisk, Ali Farhadi, Franziska Roesner, and Yejin Choi. 2019.
\newblock \href {https://proceedings.neurips.cc/paper/2019/hash/3e9f0fc9b2f89e043bc6233994dfcf76-Abstract.html} {Defending against neural fake news}.
\newblock In \emph{Advances in Neural Information Processing Systems 32: Annual Conference on Neural Information Processing Systems 2019, NeurIPS 2019, December 8-14, 2019, Vancouver, BC, Canada}, pages 9051--9062.

\bibitem[{Zheng et~al.(2018)Zheng, Wang, Chen, and Sangaiah}]{DBLP:journals/access/ZhengWCS18}
Hai{-}Tao Zheng, Wei Wang, Wang Chen, and Arun~Kumar Sangaiah. 2018.
\newblock \href {https://doi.org/10.1109/ACCESS.2017.2774839} {Automatic generation of news comments based on gated attention neural networks}.
\newblock \emph{{IEEE} Access}, 6:702--710.

\bibitem[{Zhou et~al.(2018)Zhou, Guo, and He}]{zhou-etal-2018-neural}
Deyu Zhou, Linsen Guo, and Yulan He. 2018.
\newblock \href {https://doi.org/10.18653/v1/N18-1156} {Neural storyline extraction model for storyline generation from news articles}.
\newblock In \emph{Proceedings of the 2018 Conference of the North {A}merican Chapter of the Association for Computational Linguistics: Human Language Technologies, Volume 1 (Long Papers)}, pages 1727--1736, New Orleans, Louisiana. Association for Computational Linguistics.

\end{thebibliography}
\bibliographystyle{acl_natbib}




\onecolumn
\appendix
\section{Publicly Available Models}
\label{sec:app_opensource_models}
Here, we provide all the links to publicly released models obtained within this presented research: (i)~fine-tuned generators (Section~\ref{sec:app_opensource_generators}) and (ii) fine-tuned classifiers (Section~\ref{sec:app_opensource_classifiers}).

\subsection{Fine-tuned Models for Generation}
\label{sec:app_opensource_generators}

BLOOM-Z~\cite{bloomz} instances:
\begin{itemize}
    \item English: \href{https://huggingface.co/tum-nlp/neural-news-generator-llama-2-7b-chat-en}{tum-nlp/neural-news-generator-llama-2-7b-chat-en}
    \item Hungarian: \href{https://huggingface.co/tum-nlp/neural-news-generator-llama-2-7b-chat-hu}{tum-nlp/neural-news-generator-llama-2-7b-chat-hu}
    \item Turkish: \href{https://huggingface.co/tum-nlp/neural-news-generator-llama-2-7b-chat-tr}{tum-nlp/neural-news-generator-llama-2-7b-chat-tr}
    \item Persian: \href{https://huggingface.co/tum-nlp/neural-news-generator-llama-7b-fa}{tum-nlp/neural-news-generator-llama-7b-fa}
\end{itemize}

LLaMa-2~\cite{llama2_paper} instances:
\begin{itemize}
    \item English: \href{https://huggingface.co/tum-nlp/neural-news-generator-bloomz-7b1-en}{tum-nlp/neural-news-generator-bloomz-7b1-en}
    \item Hungarian: \href{https://huggingface.co/tum-nlp/neural-news-generator-bloomz-7b1-hu}{tum-nlp/neural-news-generator-bloomz-7b1-hu}
    \item Turkish: \href{https://huggingface.co/tum-nlp/neural-news-generator-bloomz-7b1-tr}{tum-nlp/neural-news-generator-bloomz-7b1-tr}
    \item Persian: \href{https://huggingface.co/tum-nlp/neural-news-generator-bloomz-7b1-fa}{tum-nlp/neural-news-generator-bloomz-7b1-fa}
\end{itemize}

\subsection{Fine-tuned Models for Classification}
\label{sec:app_opensource_classifiers}

Multilingual BERT~\cite{bert} instances:

\begin{itemize}
    \item English: \href{https://huggingface.co/tum-nlp/neural-news-discriminator-BERT-en}{tum-nlp/neural-news-discriminator-BERT-en}
    \item Hungarian: \href{https://huggingface.co/tum-nlp/neural-news-discriminator-BERT-hu}{tum-nlp/neural-news-discriminator-BERT-hu}
    \item Turkish: \href{https://huggingface.co/tum-nlp/neural-news-discriminator-BERT-tr}{tum-nlp/neural-news-discriminator-BERT-tr}
    \item Persian: \href{https://huggingface.co/tum-nlp/neural-news-discriminator-BERT-fa}{tum-nlp/neural-news-discriminator-BERT-fa}
\end{itemize}

XLM-RoBERTa~\cite{conneau2019unsupervised} instances:

\begin{itemize}
    \item English: \href{https://huggingface.co/tum-nlp/neural-news-discriminator-RoBERTa-en}{tum-nlp/neural-news-discriminator-RoBERTa-en}
    \item Hungarian: \href{https://huggingface.co/tum-nlp/neural-news-discriminator-RoBERTa-hu}{tum-nlp/neural-news-discriminator-RoBERTa-hu}
    \item Turkish: \href{https://huggingface.co/tum-nlp/neural-news-discriminator-RoBERTa-tr}{tum-nlp/neural-news-discriminator-RoBERTa-tr}
    \item Persian: \href{https://huggingface.co/tum-nlp/neural-news-discriminator-RoBERTa-fa}{tum-nlp/neural-news-discriminator-RoBERTa-fa}
\end{itemize}

\clearpage

\section{LLM Prompts}
\label{sec:app_llms_prompts}
\subsection{Prompts for Generation} \label{sec:app_llms_prompts_generation}
Multiple prompt types suited to the target language requirements were assessed. After iterative testing and refinement, a prompt style detailed below has yielded consistently superior results.

\paragraph{English Prompt Example} Write a news article in English. Headline: The US President Biden visits Israel after an array of Violent Events in the Middle East Article: Given the current situation  [EOP] his travel on Tuesday, Donald Trump is at least trying to make it out and he was also doing that when he's coming for some time.
The first visit's man o say; "Sz had not seen any changes there before.

\paragraph {Hungarian Prompt Example} Write a news article in Hungarian. Headline: A Nabucco-vezeték sorsát félti Európa Article: Új gázimport-vezetékkel erősítené közép-ázsiai pozícióit Oroszország. A lépés várható volt, [EOP] a bővítés híre mégis sokkolta a nyugati világot, mert Oroszország elvásárolhatja az európai országok (és az USA) elől azt a gázt, amelyet az építendő Nabucco-vezetékben szeretnének látni. A Kaszpi-tengert átszelő útvonal terve miatt féltik a Nabucco-vezetéket az európaiak - írja a Villággazdaság. ...

\paragraph{Turkish Prompt Example} Write a news article in Turkish. Headline: İsrail karşı saldırıda: En az 8 bin Filistinli öldü Article: Hamas'ın silahlı kanadı İzeddin Kassam Tugayları'nın Gazze [EOP] ve yken olç konusu k çezğnda o s ginedi  Byle de milliki (70, 15 Maktlili alye dolde hallenen bölmeset) üyla adden edecten emek is ki verlerkna fta ihtik cinsim'n aldlanle orma da gela iddte be mun dalaruz bir hinde gy açndan yalari ne ba aymlalara. 

\paragraph {Persian Prompt Example}
Write a news article in Persian. Headline \FR{بحران آب در ایران و روزهای جیره‌بندی }Article\FR{ در یک خانه، یک زمین شماره ۲۷ به این سه }[EOP]\FR{ موضوع می‌پردازیم: شروع جیره‌بندی آب در برخی از شهرها و ۱۲ کلانشهر ایران در آستانه بحران تازه آب، مرگ هزاران بچه‌ماهی در رود هیرمند و شهرت یوزپلنگ‌های آسیایی یا ایرانی در دنیا بعد از نقش‌شان بر پیراهن‌های بازیکنان تیم ملی فوتبال  ایران در جام جهانی برزیل
}
\subsection{Prompts for Classification} \label{sec:app_llms_prompts_classification}
\paragraph{English Prompt Example}
Task Description: "Task Description: Considering the content and context of the text, decide whether it was written by a human or artificial intelligence. If written by human write [HUMAN], otherwise write [ARTIFICIAL INTELLIGENCE]. For example: "
Text: "North Korea’s Defense Ministry on Monday denounced the 11-day US-South Korea joint exercises as “provocative” and “reckless,” and warned North Korean forces will monitor “adventurist acts” and conduct “responsible military activities” to bring what it called an unstable security situation under control, according to the state-run Korean Central News Agency (KCNA). Thursday’s artillery exercise was the North’s second drill this week, a day after its ground forces appeared to infiltrate border guard posts during maneuvers. North Korean leader Kim Jong Un observed both drills and was seen in photos provided by state media."
Classification:[ARTIFICIAL INTELLIGENCE]."
""",
   \paragraph{Persian Prompt Example}
\FR{
  "بر اساس محتوا و زمینه، تعیین کنید که متن زیر اخبار نوشته شده توسط انشان را نشان می‌دهد یا تولید شده توسط هوش مصنوعی. اگر باهوش مصنوعی نوشته شده بود بنویس [هوش مصنوعی] در غیر اینصورت بنویس [انسان]." متن: "معاون دادستان کل کشور در پاسخ به پرسشی در خصوص علل ایجاد زمینه‌های ظهور پدیده‌ای مانند کوروش کمپانی علی رغم وجود نظارت‌ها یادآور شد: یکی از جهاتی که زمینه عدم شفافیت و اقدام به رفتارهای مجرمانه را ایجاد می‌کند نبود زیرساخت‌های فناورانه و گردش به هنگام اطلاعات است. ما زیر ساخت بسیار زیاد داریم اما همه را در چنبره خودمان گرفته‌ایم و نسبت به آنها نگاه بخشی می‌شود. با اینکه در قوانین مختلف تکلیف شده که دستگاه‌ها ملزم هستند گردش به هنگام اطلاعات داشته باشند، اطلاعات را در بایگانی‌های خود گذاشته‌ایم و در استفاده از اطلاعات و گردش آن خست داریم ." دسته بندی: انسان
  }
   \paragraph{Hungarian Prompt Example}
Feladatleírás: figyelembe véve az alábbi szöveg tartalmát és kontextusát, döntsd el hogy a szöveget ember, vagy egy mesterséges intelligencia írta. Ha ember írta, írd azt hogy [EMBER], ha pedig mesterséges intelligencia, akkor írd azt hogy [MESTERSÉGES INTELLIGENCIA]. Például:
Szöveg: "A programot bemutató sajtótájékoztatón a Környezetvédelmi Minisztérium képviselői elmondták, hogy az elsődleges cél az, hogy Magyarország 2030-ra teljesíteni tudja az Európai Unió által meghatározott környezetvédelmi célokat, és példát mutasson a fenntarthatóság terén. A kezdeményezés magában foglalja a napenergia, a szélenergia és a geotermikus energia használatának növelését, valamint az elektromos járművek elterjedésének ösztönzését."
[MESTERSÉGES INTELLIGENCIA]
""",
 \paragraph{Turkish Prompt Example}
Görev Tanımı: "Metnin içeriği ve bağlamına dayanarak, aşağıdaki metnin insan tarafından mı yoksa yapay zeka tarafından mı yazıldığını belirle. İnsan tarafından yazıldıysa [INSAN], yapay zeka tarafından yazıldıysa [YAPAY ZEKA] yaz. "Örnek":
Metin: "Konferansın açılış konuşmasını yapan Teknoloji Bakanı, Türkiye'nin teknoloji ve inovasyonda dünya çapında bir merkez haline gelme hedefine vurgu yaptı. Bakan, 'Yenilikçi fikirler ve sürdürülebilir teknolojiler, ülkemizin geleceğini şekillendirecek. Bu konferans, fikir alışverişinde bulunmak ve yeni iş birliklerine yelken açmak için mükemmel bir fırsat' dedi."
Sınıflandırma: "[YAPAY ZEKA]"
"""

\section{Initial Round of Zero-Shot Generations for Model Selection}\label{sec:initial_round_of_0_zhot_gens}
\begin{table*}[h!tbp]
    \centering
    \footnotesize
    \begin{tabular}{c|c|c|c}
        \toprule
          & \shortstack{Parameter Count} & \shortstack{Multilingual\\ Pre-training} & \shortstack{Number of Generations} \\ 
        \midrule
        BloomZ & 3B  & TRUE & 100 $\times$ 4  \\
        \midrule
        Bactrian-X-LLaMa & 7B & TRUE & 100  $\times$ 4 \\
        \midrule
        Alpaca (Big) & 13B & FALSE & 100  \\
        \midrule
        Dolly v2 & 12B & FALSE & 100  \\
        \midrule
        LLaMa-2 & 7B & FALSE & 100 $\times$ 4 \\
        \midrule
        LLaMa-2-Chat & 7B & FALSE & 100 $\times$ 4 \\
        \midrule
        Mistral & 7B & FALSE & 100 $\times$ 4 \\
        \midrule
        ChatGPT (GPT-3.5) & 175B & TRUE & 100 $\times$ 4 \\       
        \bottomrule
    \end{tabular}
    \caption{Initial zero-shot generations to assess the best LLMs for multilingual full fine-tuning. For some monolingual (English) models multilingual zero-shot generation was done to see incidental language acquisition \cite{bloomz, alpaca, DatabricksBlog2023DollyV2, llama2_paper, mistral, mixtral}.}
    \label{tab:llm_selection_zero_shot_data}
\end{table*}

\clearpage 
\newpage

\section{Linguistic analyses of generations from zero-shot and finetuned LLMs} \label{sec:plots}
\subsection{CoLA Plots on English Data} \label{subsection:cola_plots}
\begin{figure}[h!tbp]
  \centering
  \begin{subfigure}{0.22\textwidth}
    \centering
    \includegraphics[width=\linewidth]{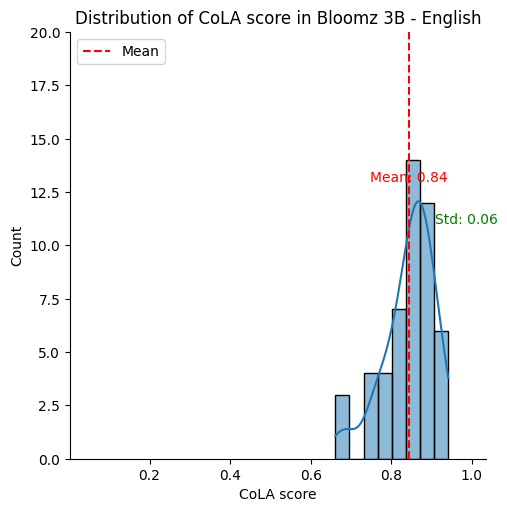}
    \caption{BloomZ-3B.}
    \label{subfig:cola-bloomz}
  \end{subfigure}
  \begin{subfigure}{0.22\textwidth}
    \centering
    \includegraphics[width=\linewidth]{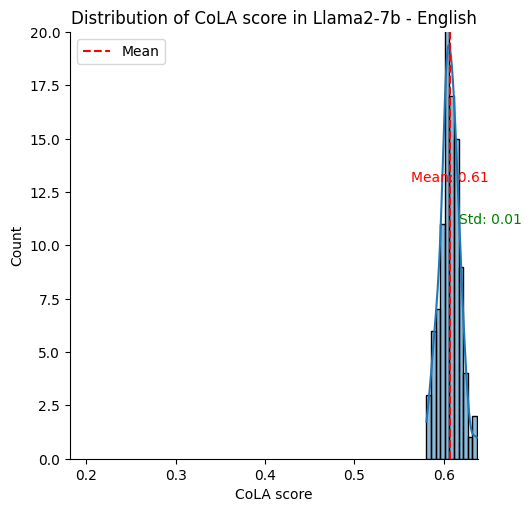}
    \caption{LLaMa-2-Chat-7B.}
    \label{subfig:cola-llama2chat}
  \end{subfigure}
  \begin{subfigure}{0.22\textwidth}
    \centering
    \includegraphics[width=\linewidth]{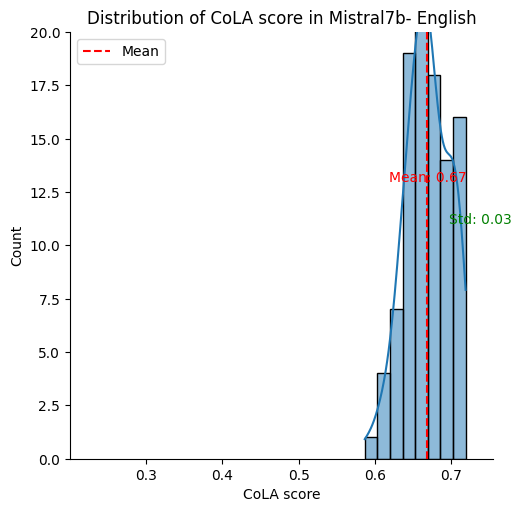}
    \caption{Mistral-7B.}
    \label{subfig:cola-Mistral7b }
  \end{subfigure}
  \begin{subfigure}{0.22\textwidth}
    \centering
    \includegraphics[width=\linewidth]{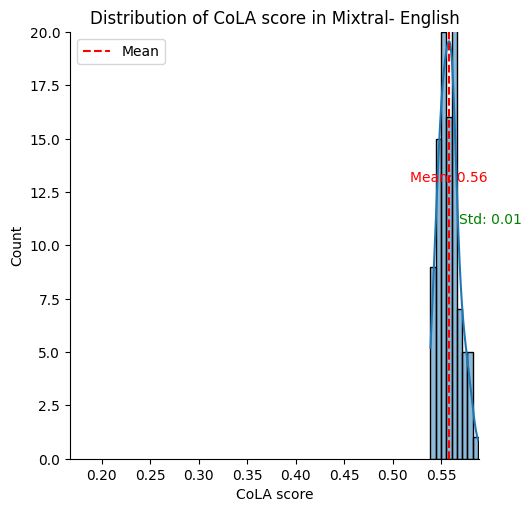}
    \caption{Mixtral.}
    \label{subfig:cola-mixtral}
  \end{subfigure}
  \begin{subfigure}{0.22\textwidth}
    \centering
    \includegraphics[width=\linewidth]{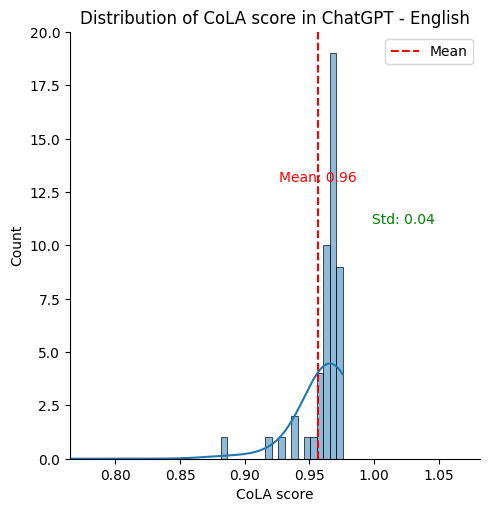}
    \caption{ChatGPT (GPT-3).}
    \label{subfig:cola-chatgpt}
  \end{subfigure}
  \begin{subfigure}{0.22\textwidth}
    \centering
    \includegraphics[width=\linewidth]{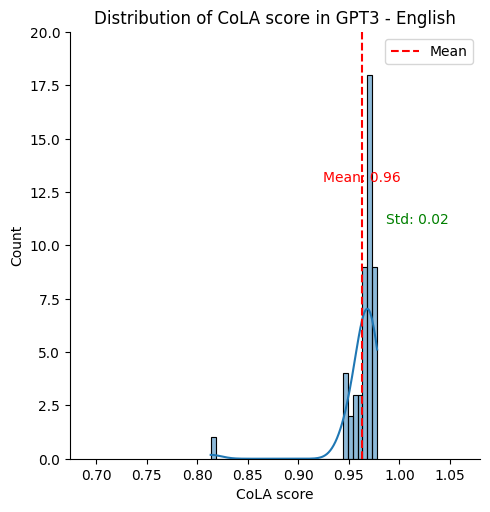}
    \caption{DaVinci.}
    \label{subfig:cola-DaVinci}
  \end{subfigure}
  \begin{subfigure}{0.22\textwidth}
    \centering
    \includegraphics[width=\linewidth]{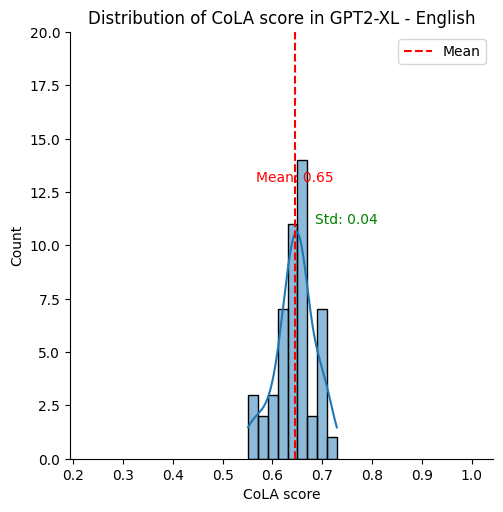}
    \caption{GPT2-XL.}
    \label{subfig:cola-GPT2-XL}
  \end{subfigure}
  \begin{subfigure}{0.22\textwidth}
    \centering
    \includegraphics[width=\linewidth]{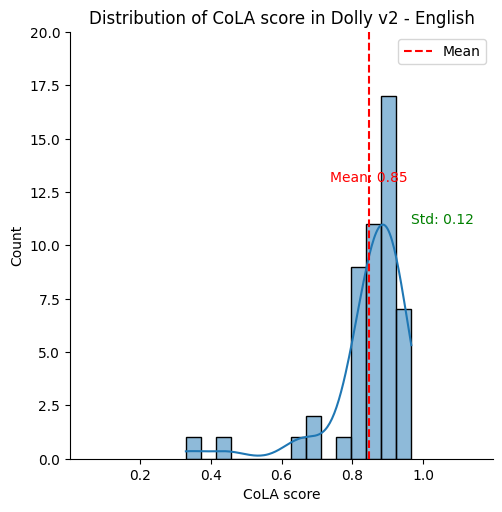}
    \caption{DOLLY v2.}
    \label{subfig:cola-DOLLY v2}
  \end{subfigure}
  \begin{subfigure}{0.22\textwidth}
    \centering
    \includegraphics[width=\linewidth]{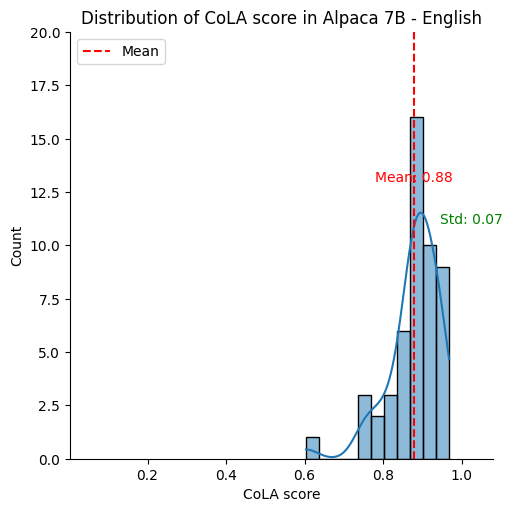}
    \caption{ALPACA 7B.}
    \label{subfig:ALPACA 7B}
  \end{subfigure}
  \begin{subfigure}{0.22\textwidth}
    \centering
    \includegraphics[width=\linewidth]{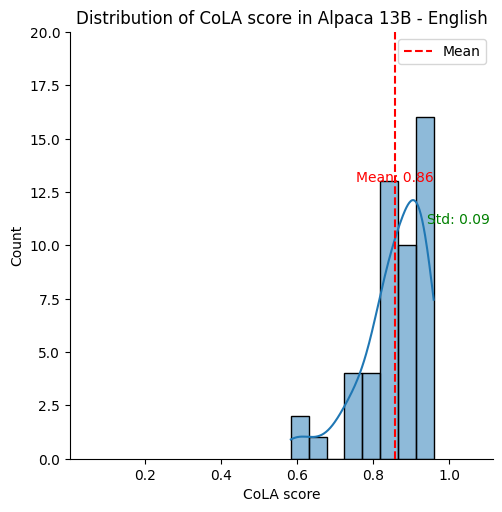}
    \caption{ALPACA 13B.}
    \label{subfig:ALPACA 13B}
  \end{subfigure}
   \begin{subfigure}{0.22\textwidth}
    \centering
    \includegraphics[width=\linewidth]{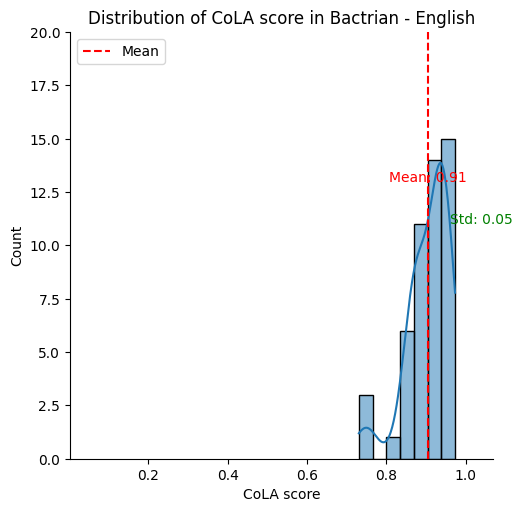}
    \caption{Bactrian-X-LLaMa-7B.}
    \label{subfig:BACTRIAN 7B}
  \end{subfigure}
  \caption{CoLA distributions over zero-shot English generations of various LLMs \cite{Cola_citation}.}
\end{figure}

\clearpage 
\newpage
\subsection{Zipf's distributions of generated news} \label{subsection:zipf_plots}
\begin{figure}[h!tbp]
    \centering
    \begin{subfigure}{0.22\textwidth}
        \centering
        \includegraphics[width=\linewidth, height=3cm]{figures/zipf/bloomz.png}
        \caption{Zero-shot BloomZ-3B.}
    \end{subfigure}
    \begin{subfigure}{0.22\textwidth}
        \centering
        \includegraphics[width=\linewidth, height=3.25cm]{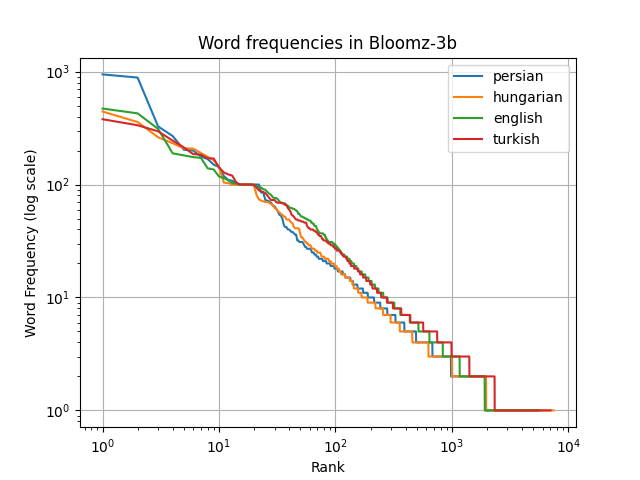}
        \caption{Fine-tuned BloomZ-3B.}
        \label{subfig:Bloomz-7b Zipsf's}
    \end{subfigure}
    \begin{subfigure}{0.22\textwidth}
        \centering
        \includegraphics[width=\linewidth, height=3cm]{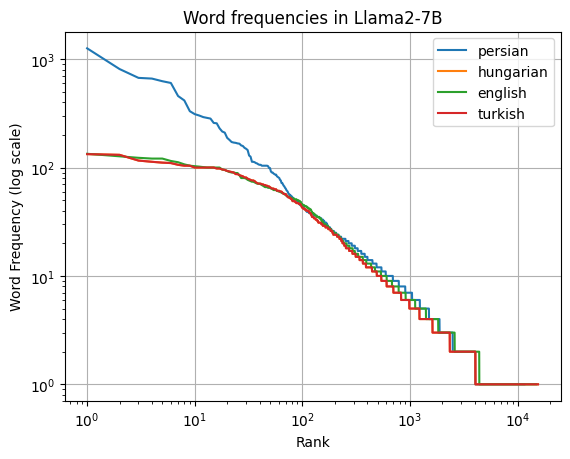}
        \caption{Zero-shot LLaMa.}
        \label{Llama2-7B Zipsf's}
    \end{subfigure}
    \begin{subfigure}{0.22\textwidth}
        \centering
        \includegraphics[width=\linewidth, height=3.25cm]{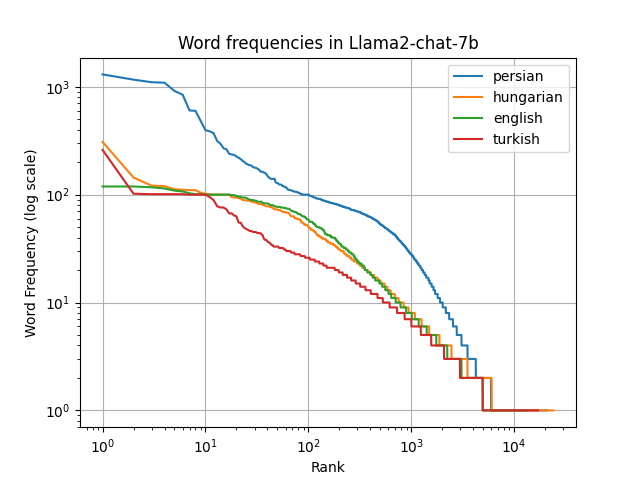}
        \caption{Fine-tuned LLaMa.}
        \label{subfig:Llama2-chat-7b Zipsf's}
    \end{subfigure}
    \begin{subfigure}{0.22\textwidth}
        \centering
        \includegraphics[width=\linewidth, height=3.25cm]{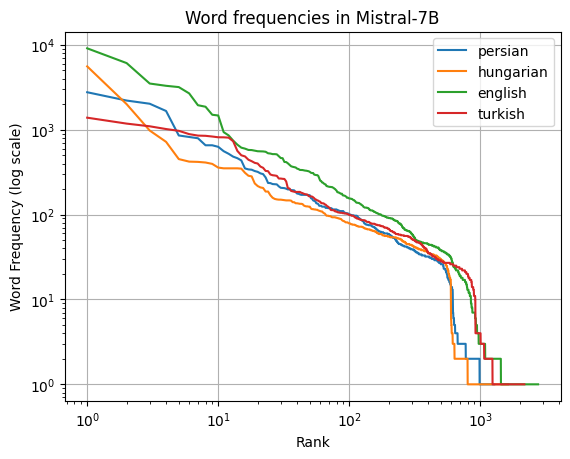}
        \caption{Zero-shot Mistral-7B.}
        \label{subfig:mistral-zipf}
    \end{subfigure}
    \begin{subfigure}{0.22\textwidth}
        \centering
        \includegraphics[width=\linewidth, height=3.25cm]{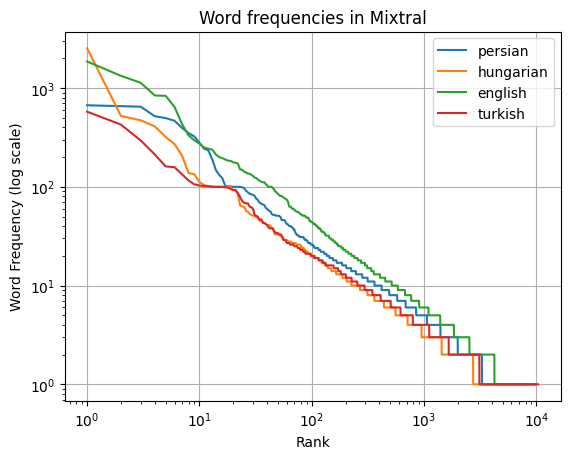}
        \caption{Zero-shot Mixtral.}
        \label{subfig:mixtral-zipf}
    \end{subfigure}    
    \begin{subfigure}{0.22\textwidth}
        \centering
        \includegraphics[width=\linewidth, height=3.25cm]{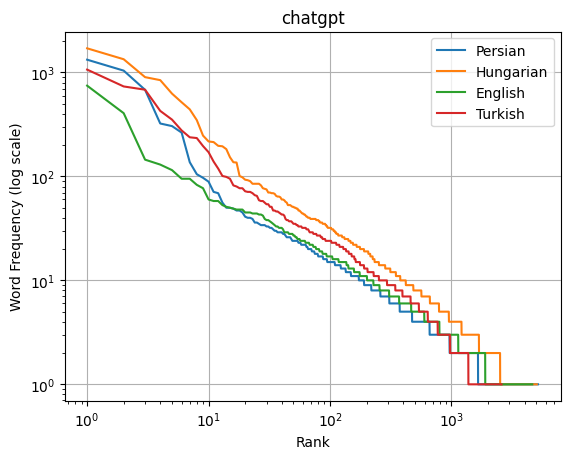}
        \caption{Zero-shot ChatGPT.}
        \label{subfig:ChatGPT-zipf}
    \end{subfigure}
    \begin{subfigure}{0.22\textwidth}
        \centering
        \includegraphics[width=\linewidth, height=3.25cm]{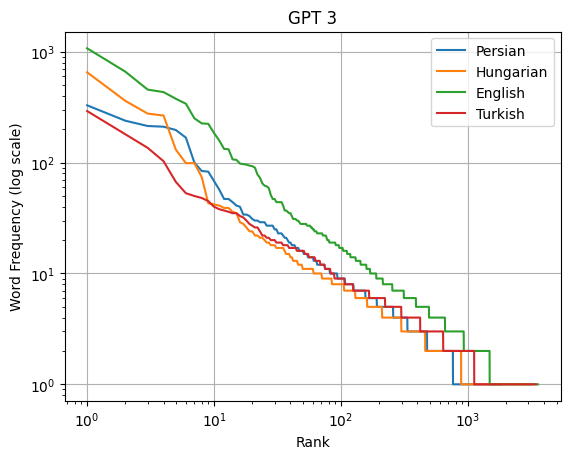}
        \caption{Zero-shot DaVinci.}
        \label{subfig:DaVinci-zipf}
    \end{subfigure} 
   \begin{subfigure}{0.22\textwidth}
    \centering
    \includegraphics[width=\linewidth, height=3.25cm]{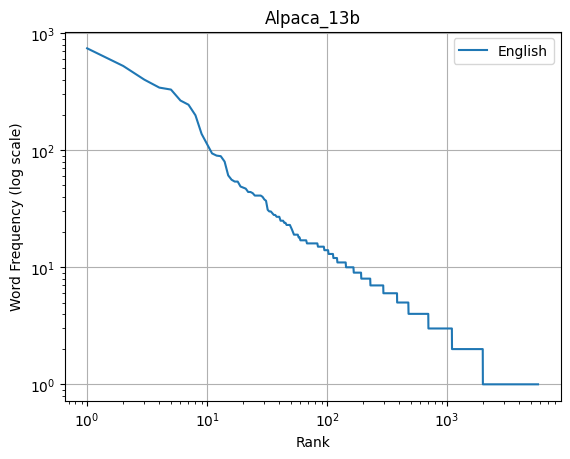}
    \caption{Zero-shot ALPACA-13B.}
    \label{subfig:ALPACA 13B Zipsf's}
  \end{subfigure}
  \begin{subfigure}{0.22\textwidth}
    \centering
    \includegraphics[width=\linewidth, height=3.25cm]{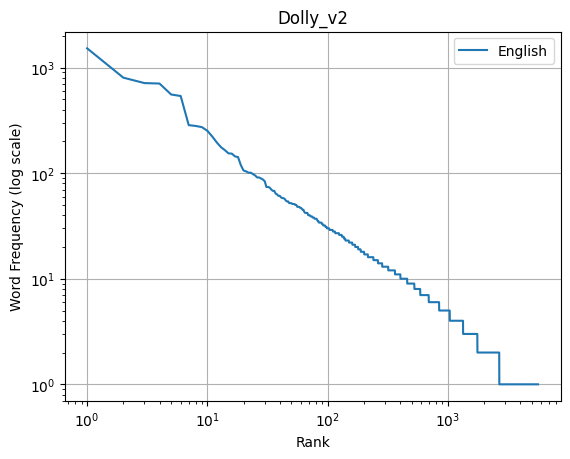}
    \caption{Zero-shot DOLLYv2.}
    \label{subfig:DOLLY v2 Zipsf's}
  \end{subfigure}
  \begin{subfigure}{0.22\textwidth}
    \centering
    \includegraphics[width=\linewidth, height=3.25cm]{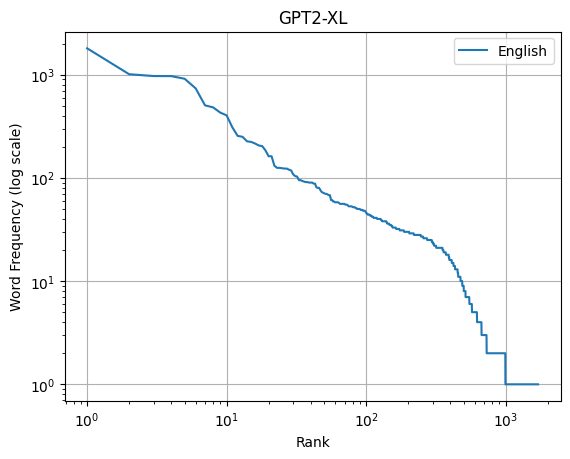}
    \caption{Zero-shot GPT2-XL.}
    \label{subfig:GPT2-XL 1.B Zipsf's}
  \end{subfigure}

    \begin{subfigure}{0.22\textwidth}
    \centering
    \includegraphics[width=\linewidth, height=3.25cm]{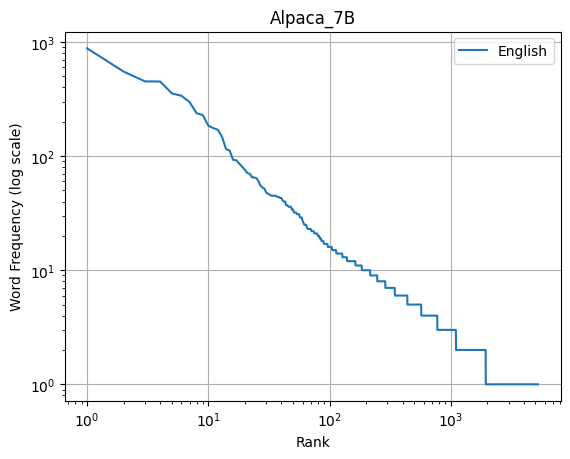}
    \caption{Zero-shot ALPACA-7B.}
    \label{subfig:ALPACA 7B Zipf}
  \end{subfigure}
     \begin{subfigure}{0.22\textwidth}
    \centering
    \includegraphics[width=\linewidth, height=3.25cm]{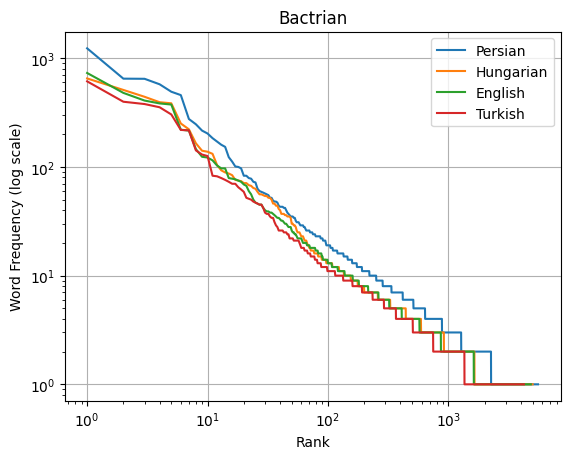}
    \caption{Zero-shot Bactrian-X-LLaMa-7B.}
    \label{subfig:BACTRIAN 7B  Zipsf's}
    
  \end{subfigure}
      \begin{subfigure}{0.22\textwidth}
    \centering
    \includegraphics[width=\linewidth, height=3.25cm]{figures/zipf/zipf_Human_written/all.png}
    \caption{Human-written ground truth}
    \label{subfig:Human Written Zipsf's}
  \end{subfigure}
 
\caption{Zipf's Distributions Over Zero-shot Prompting and Fine-tuned LLMs' Generations in mono- or multilingual setting. LLaMa-2-Chat-7B for English, Turkish and Hungarian. Persian-LLaMa-7B-Merged for Persian \cite{zipf}}
\end{figure}  

\clearpage 
\newpage
\subsection{Word length distributions per split} \label{subsec:length-distr}
Here we provide the lengths distribution of all machine-generated and human-written articles per split, per model, and per language. Please refer to Table \ref{tab:benchmark_data_info} for an overview of train, validation and test splits.  

\begin{figure}[h!tbp]
    \centering
    \begin{subfigure}{0.22\textwidth}
        \centering
        \includegraphics[width=\linewidth, height=3cm]{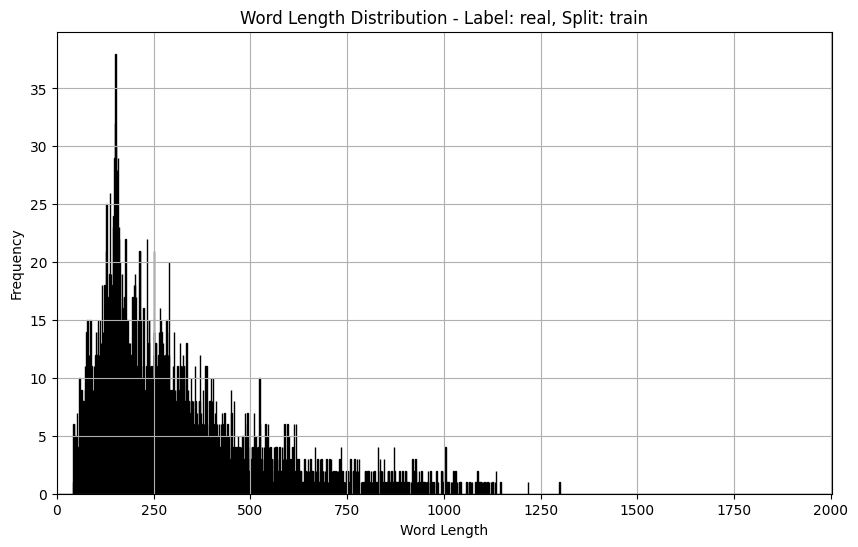}
        \caption{Label: real,\\Split: train}
    \end{subfigure}
    \begin{subfigure}{0.22\textwidth}
        \centering
        \includegraphics[width=\linewidth, height=3.25cm]{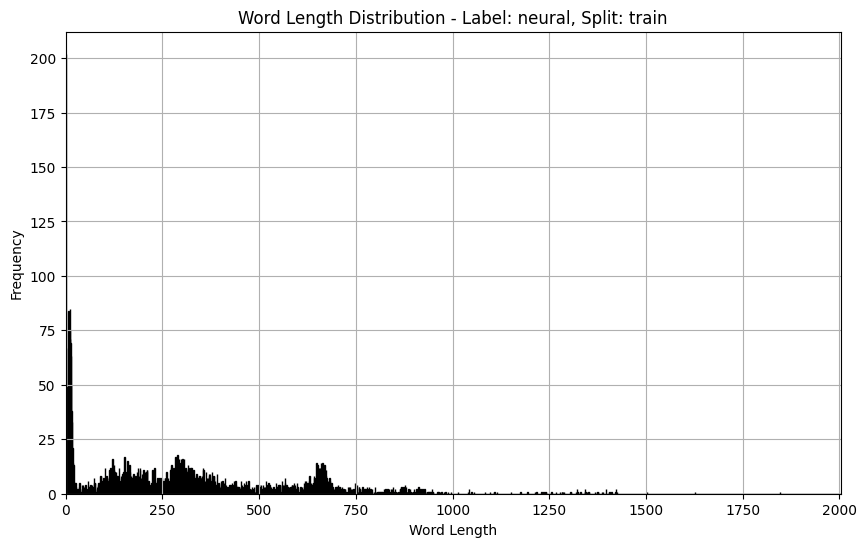}
        \caption{Label: neural,\\Split: train}
        \label{subfig:2 Zipsf's}
    \end{subfigure}
    \begin{subfigure}{0.22\textwidth}
        \centering
        \includegraphics[width=\linewidth, height=3cm]{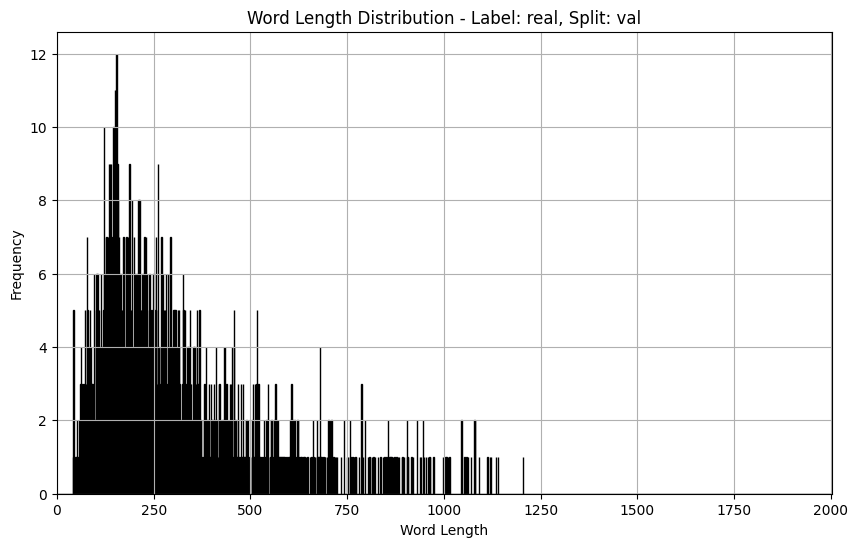}
        \caption{Label: real,\\Split: val}
        \label{3 Zipsf's}
    \end{subfigure}
    \begin{subfigure}{0.22\textwidth}
        \centering
        \includegraphics[width=\linewidth, height=3.25cm]{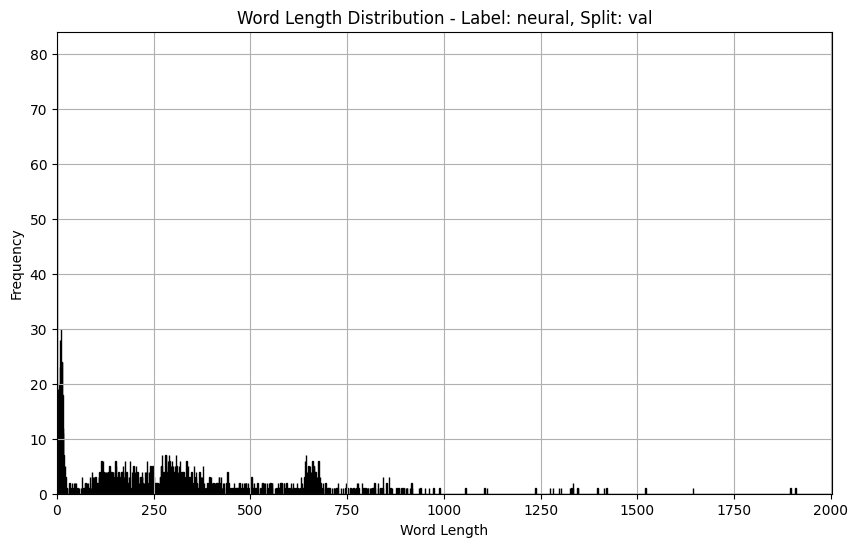}
        \caption{Label: neural,\\Split: val}
        \label{subfig:4 Zipsf's}
    \end{subfigure}
    \begin{subfigure}{0.22\textwidth}
        \centering
        \includegraphics[width=\linewidth, height=3.25cm]{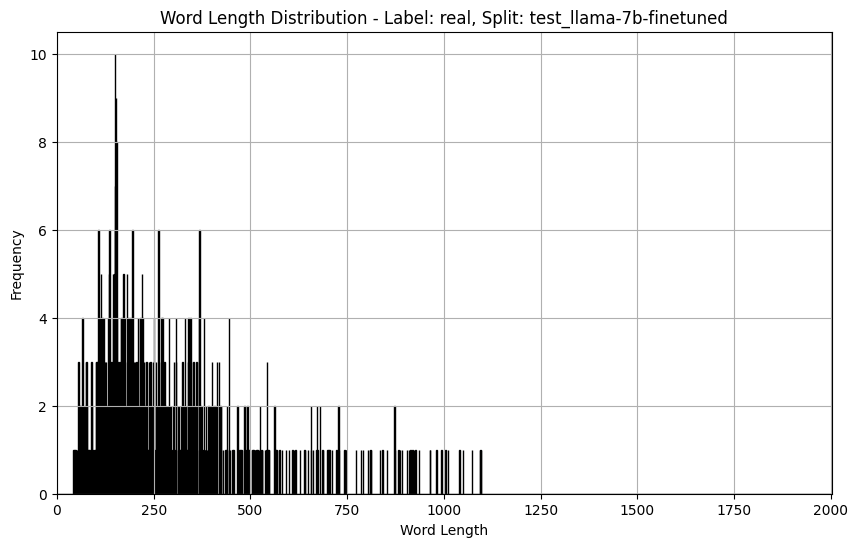}
        \caption{Label: real, Split: test\_llama-7b-finetuned}
        \label{subfig:19}
    \end{subfigure}
    \begin{subfigure}{0.22\textwidth}
        \centering
        \includegraphics[width=\linewidth, height=3.25cm]{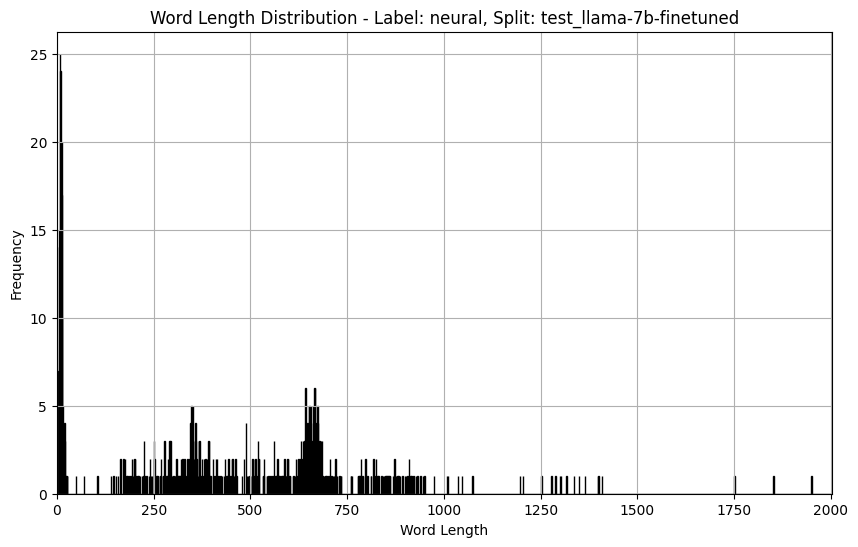}
        \caption{Label: neural,\\Split: test\_llama-7b-finetuned}
        \label{subfig:20}
    \end{subfigure}
    \begin{subfigure}{0.22\textwidth}
        \centering
        \includegraphics[width=\linewidth, height=3.25cm]{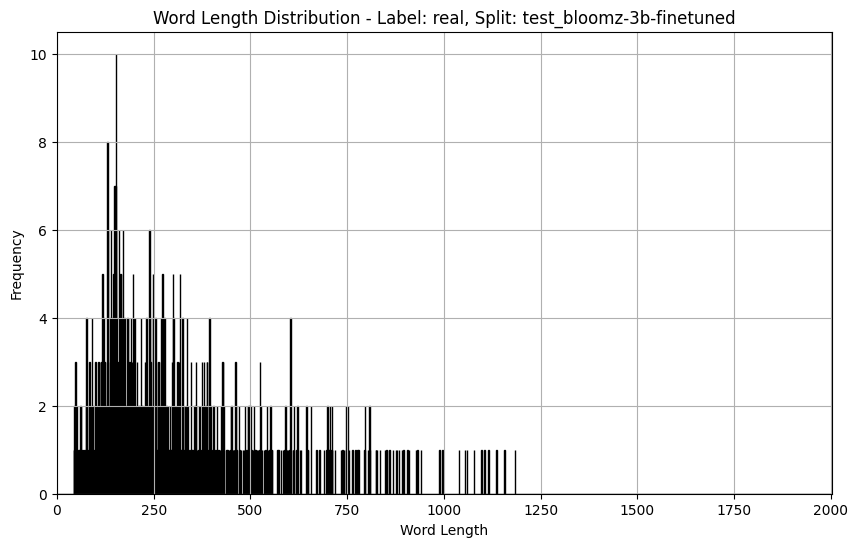}
        \caption{Label: real, Split: test\_bloomz-3b-finetuned}
        \label{subfig:13}
    \end{subfigure}
    \begin{subfigure}{0.22\textwidth}
        \centering
        \includegraphics[width=\linewidth, height=3.25cm]{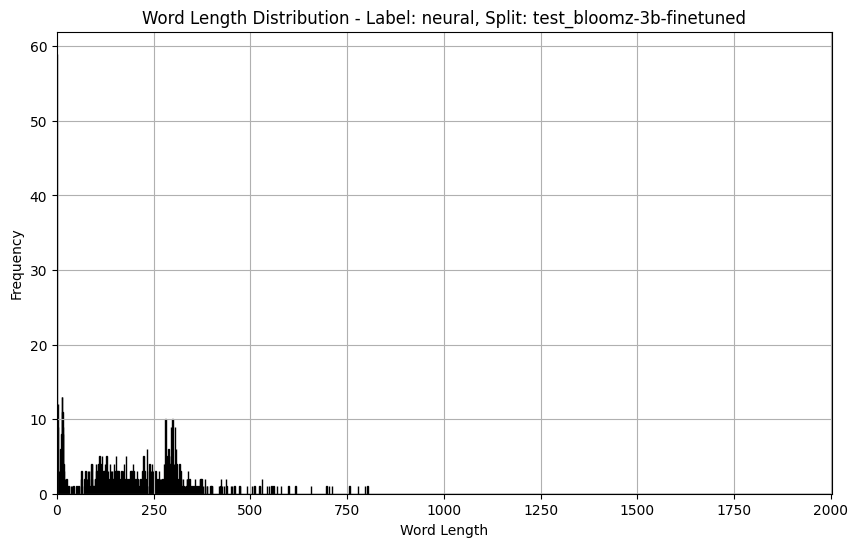}
        \caption{Label: neural, Split: test\_bloomz-3b-finetuned}
    \end{subfigure}
    \begin{subfigure}{0.22\textwidth}
        \centering
        \includegraphics[width=\linewidth, height=3.25cm]{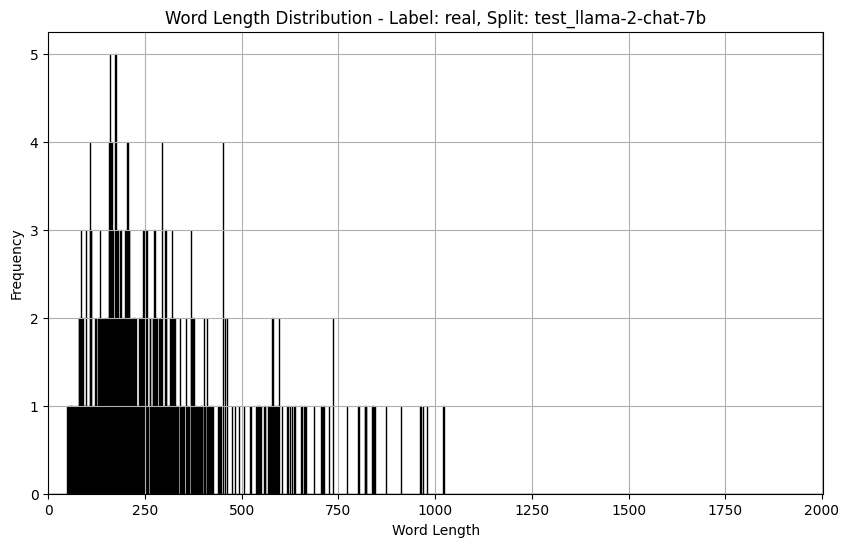}
        \caption{Label: real,\\Split: test\_llama-2-chat-7b}
        \label{subfig:17}
    \end{subfigure}
    \begin{subfigure}{0.22\textwidth}
        \centering
        \includegraphics[width=\linewidth, height=3.25cm]{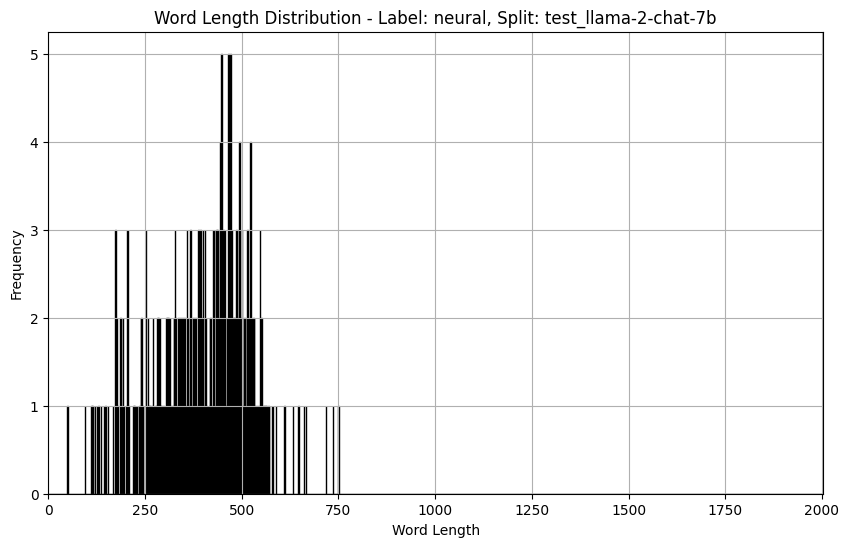}
        \caption{Label: neural,\\Split: test\_llama-2-chat-7b}
        \label{subfig:18}
    \end{subfigure}
      \begin{subfigure}{0.22\textwidth}
        \centering
        \includegraphics[width=\linewidth, height=3.25cm]{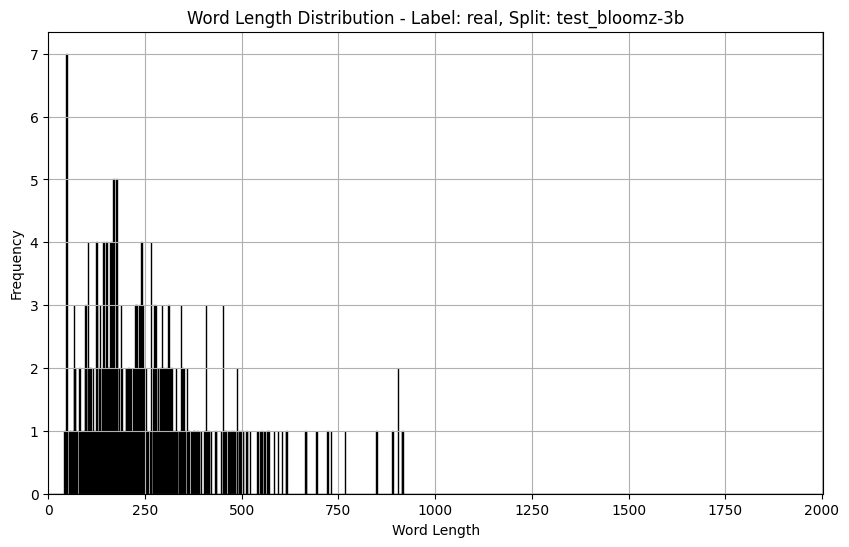}
        \caption{Label: real,\\Split: test\_bloomz-3b}
        \label{subfig:11}
      \end{subfigure}
    \begin{subfigure}{0.22\textwidth}
        \centering
        \includegraphics[width=\linewidth, height=3.25cm]{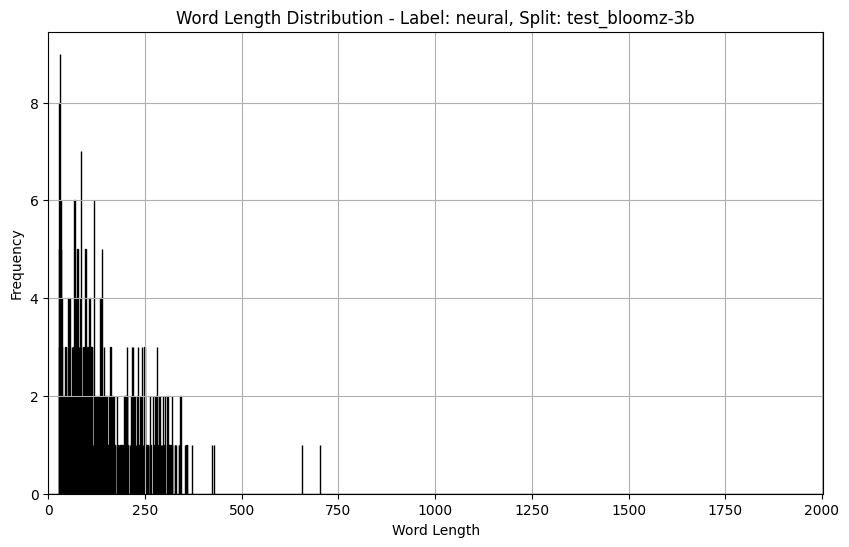}
        \caption{Label: neural,\\Split: test\_bloomz-3b}
        \label{subfig:12}
     \end{subfigure}
    \begin{subfigure}{0.22\textwidth}
        \centering
        \includegraphics[width=\linewidth, height=3.25cm]{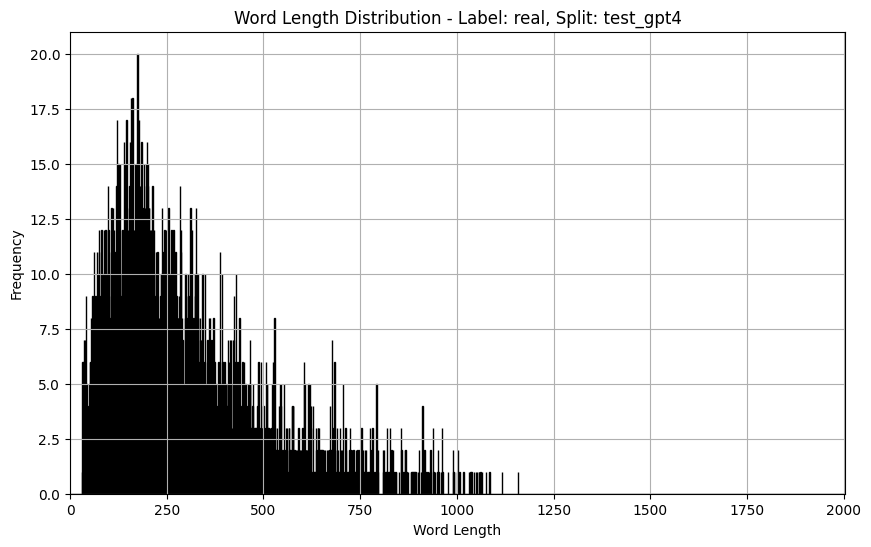}
        \caption{Label: real,\\Split: test\_gpt4}
        \label{subfig:14}
    \end{subfigure}
    \begin{subfigure}{0.22\textwidth}
        \centering
        \includegraphics[width=\linewidth, height=3.25cm]{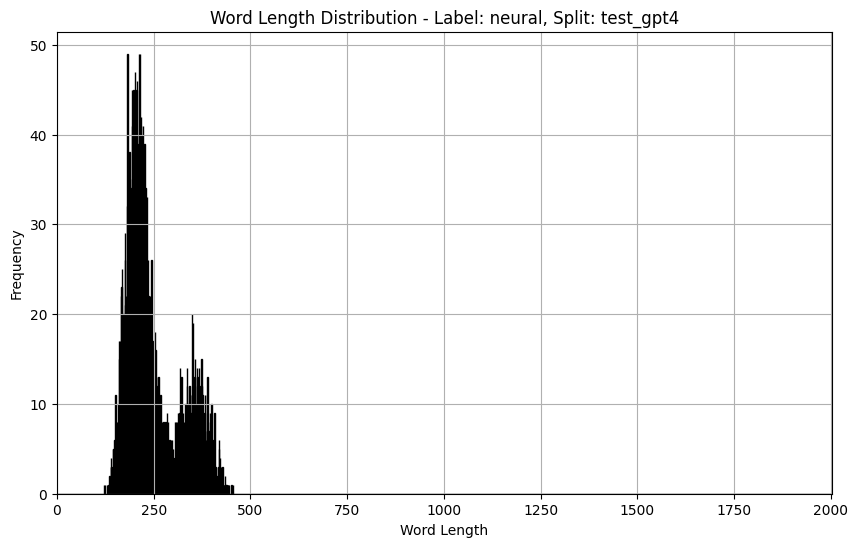}
        \caption{Label: neural,\\Split: test\_gpt4}
        \label{subfig:16}
    \end{subfigure}
    \begin{subfigure}{0.22\textwidth}
    \centering
    \includegraphics[width=\linewidth, height=3.25cm]{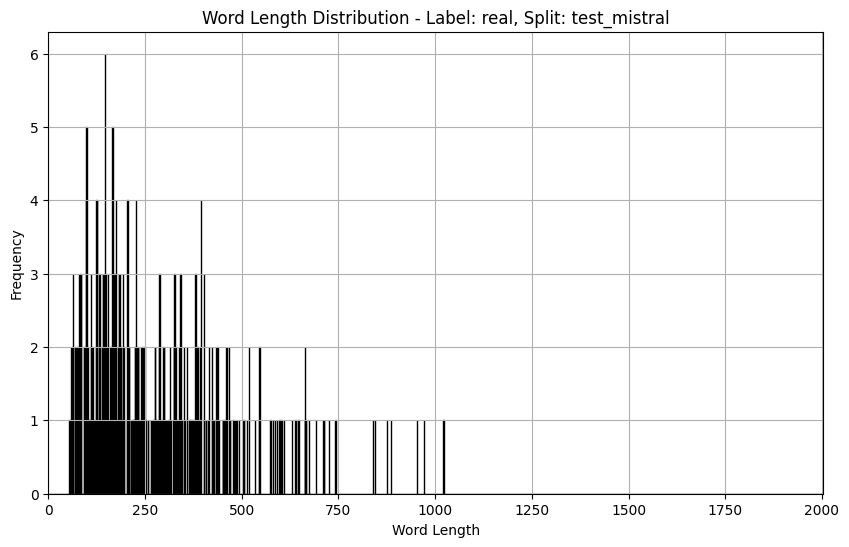}
    \caption{Label: real,\\Split: test\_mistral}
    \label{subfig:21}
  \end{subfigure}
          \begin{subfigure}{0.22\textwidth}
    \centering
    \includegraphics[width=\linewidth, height=3.25cm]{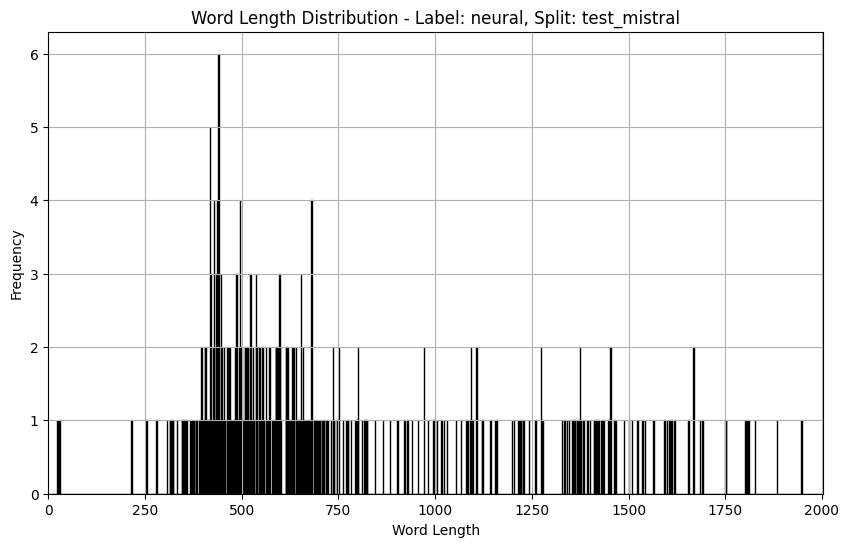}
    \caption{Label: neural,\\Split: test\_mistral}
    \label{subfig:22}
  \end{subfigure}
          \begin{subfigure}{0.22\textwidth}
    \centering
    \includegraphics[width=\linewidth, height=3.25cm]{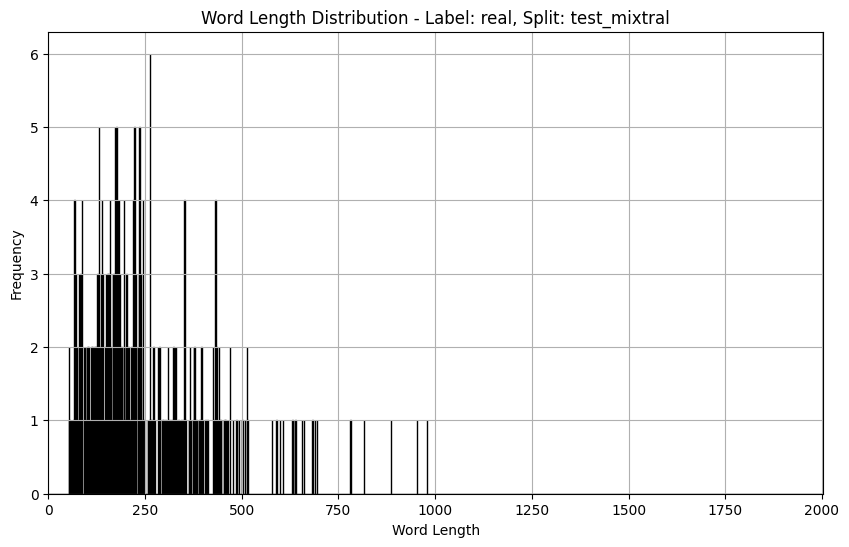}
    \caption{Label: real,\\Split: test\_mixtral}
    \label{subfig:23}
  \end{subfigure}
          \begin{subfigure}{0.22\textwidth}
    \centering
    \includegraphics[width=\linewidth, height=3.25cm]{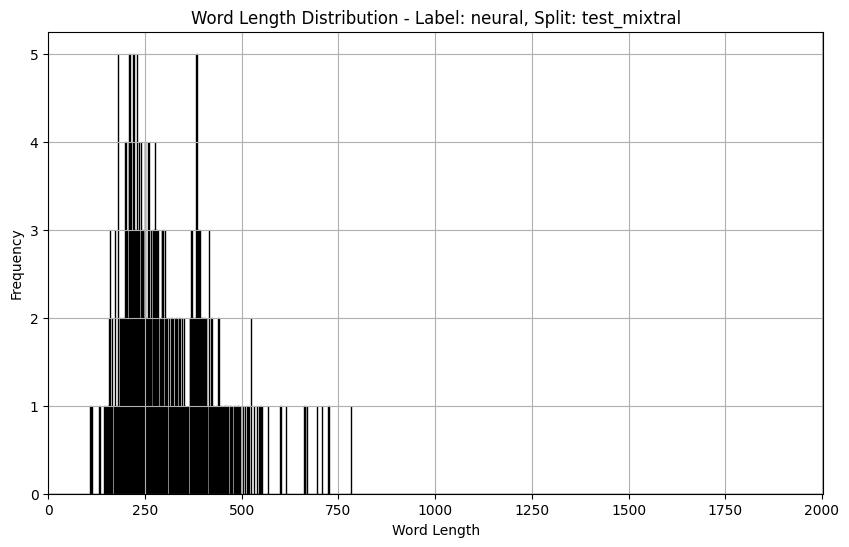}
    \caption{Label: neural,\\Split: test\_mixtral}
    \label{subfig:24}
  \end{subfigure}
          \begin{subfigure}{0.22\textwidth}
    \centering
    \includegraphics[width=\linewidth, height=3.25cm]{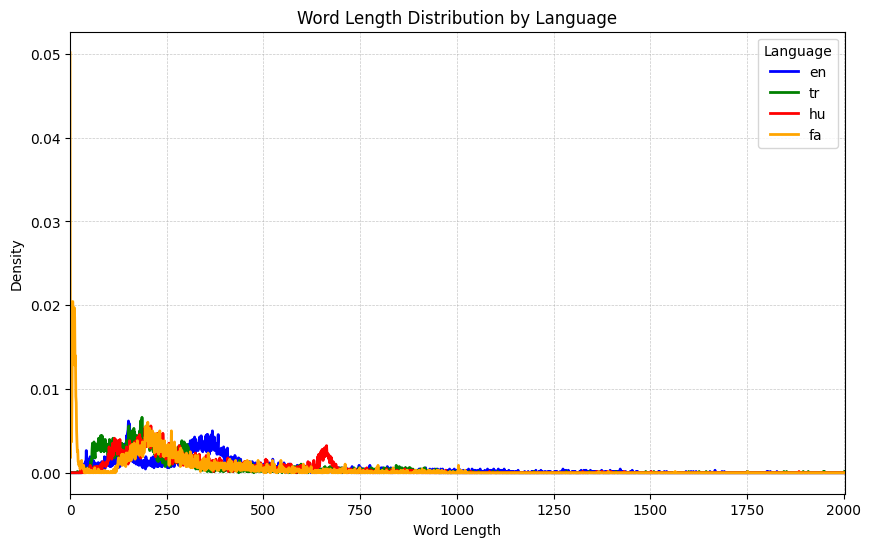}
    \caption{Per language,\\Label: all, Split: all}
    \label{subfig:lang}
  \end{subfigure}
\end{figure}
\clearpage 
\newpage

\subsection{Top-k TF-IDF features of generations per model and language} \label{subsection:tfidf_plots}
\begin{figure}[h!tbp]
  \centering
  \begin{subfigure}{0.22\textwidth}
    \centering
     \includegraphics[width=\linewidth,height=1.85cm]{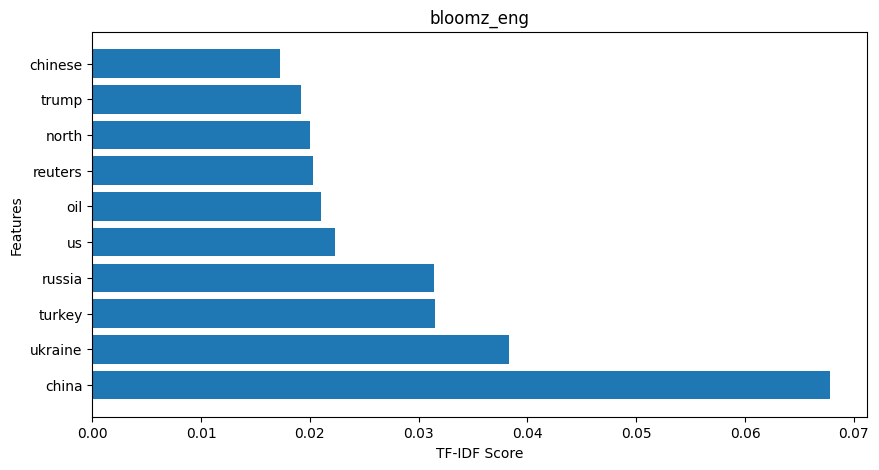}
    \caption{BloomZ-3B Zero-shot in English.}
    \label{subfig:bloomz_3_shot_eng_tfidf}  
  \end{subfigure}
    \centering
  \begin{subfigure}{0.22\textwidth}
    \centering
     \includegraphics[width=\linewidth,height=1.85cm]{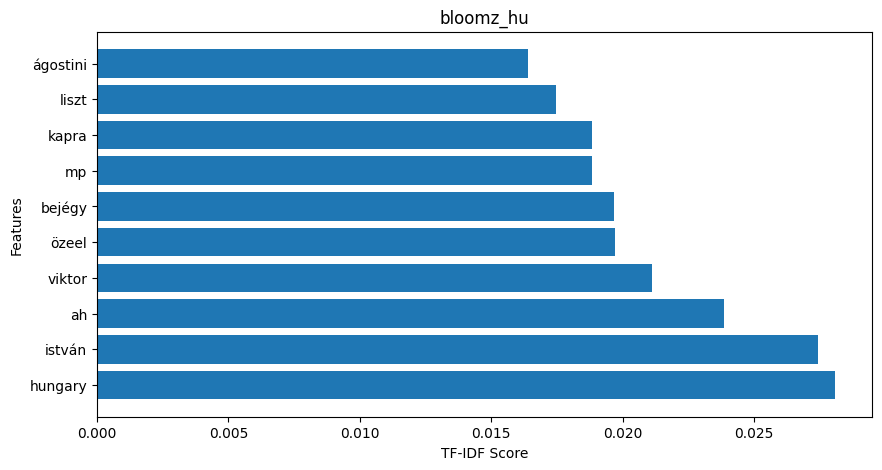}
    \caption{BloomZ-3B Zero-shot in Hungarian.}
    \label{subfig:bloomz-en-0-shot-tfidf}  
  \end{subfigure}
   \begin{subfigure}{0.22\textwidth}
    \centering
     \includegraphics[width=\linewidth,height=1.85cm]{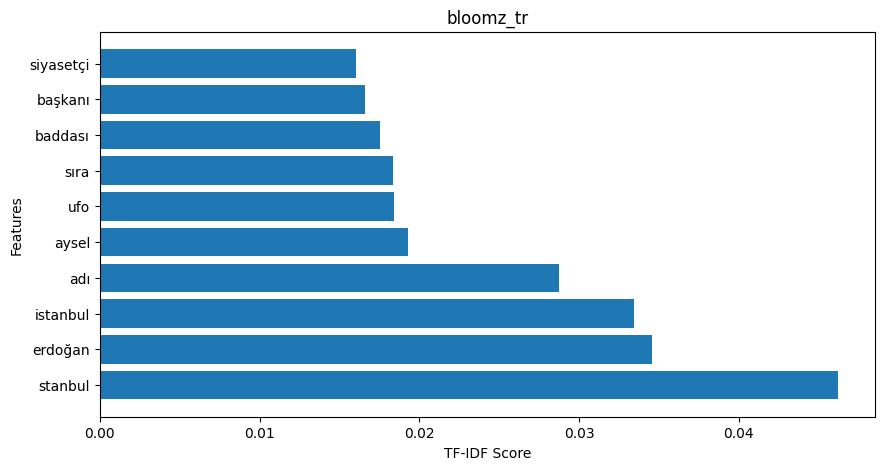}
    \caption{BloomZ-3B Zero-shot in Turkish.}
    \label{subfig:bloomz-tr-0-shot-tfidf}  
  \end{subfigure}
  \begin{subfigure}{0.22\textwidth}
    \centering
     \includegraphics[width=\linewidth,height=1.85cm]{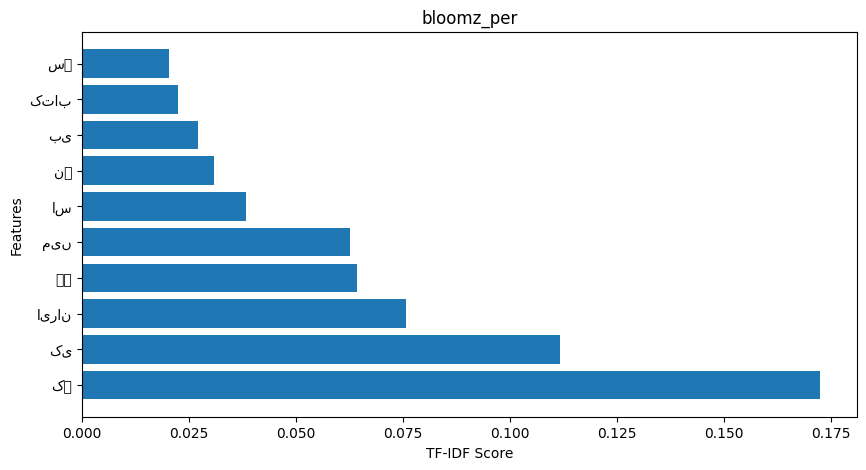}
    \caption{BloomZ-3B Zero-shot in Persian.}
    \label{subfig:bloomz-fa-0-shot-tfidf}  
  \end{subfigure}
  \begin{subfigure}{0.22\textwidth}
    \centering
     \includegraphics[width=\linewidth,height=1.85cm]{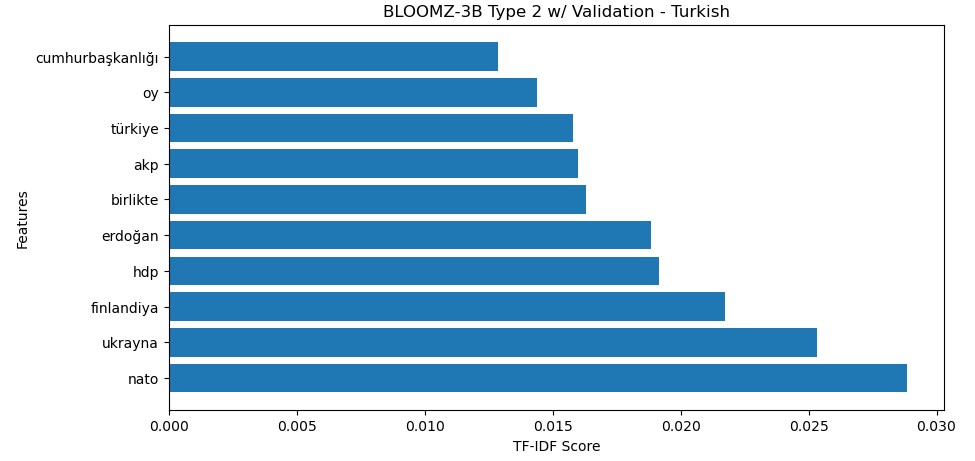}
    \caption{BloomZ-3B Fine-tuned in Turkish.}
    \label{subfig:bloomz-tr-fine-tuned-tfidf}
  \end{subfigure}
  \begin{subfigure}{0.22\textwidth}
    \centering
    \includegraphics[width=\linewidth,height=1.85cm]{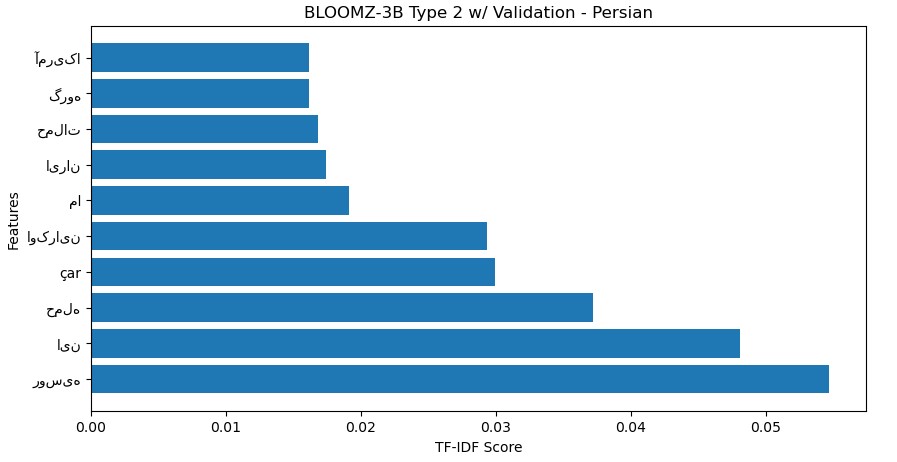}
    \caption{BloomZ-3B Fine-tuned in Persian.}
    \label{subfig:bloomz-fa-fine-tuned-tfidf}
  \end{subfigure}
  \begin{subfigure}{0.22\textwidth}
    \centering
\includegraphics[width=\linewidth,height=1.85cm]{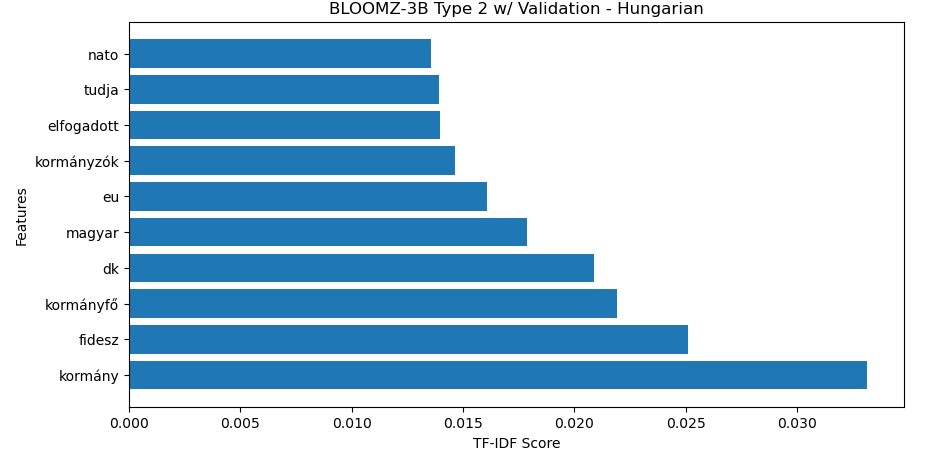}
    \caption{BloomZ-3B Fine-tuned in Hungarian.}
    \label{subfig:bloomz-hu-fine-tuned-tfidf}
  \end{subfigure}
    \begin{subfigure}{0.22\textwidth}
    \centering
\includegraphics[width=\linewidth,height=1.85cm]{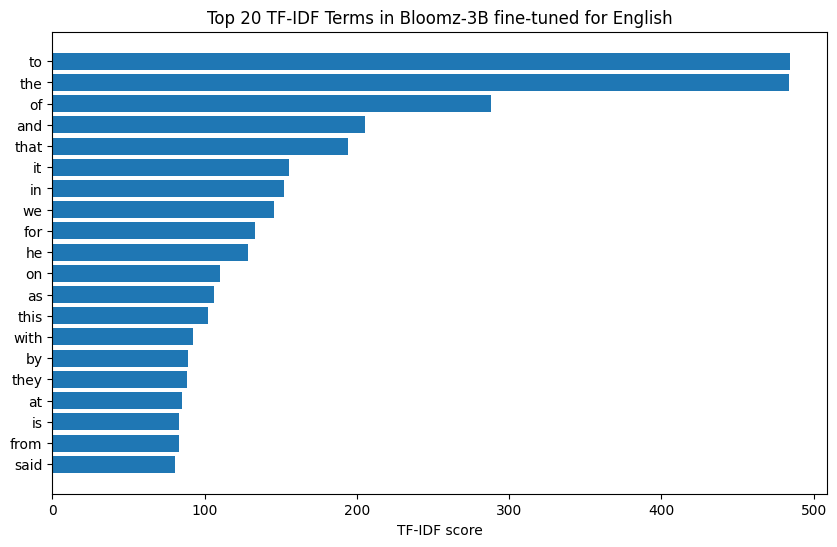}
    \caption{BloomZ-3B Fine-tuned in English.}
    \label{subfig:bloomz-en-fine-tuned-tfidf}
  \end{subfigure}
\begin{subfigure}{0.22\textwidth}
    \centering
    \includegraphics[width=\linewidth,height=1.85cm]{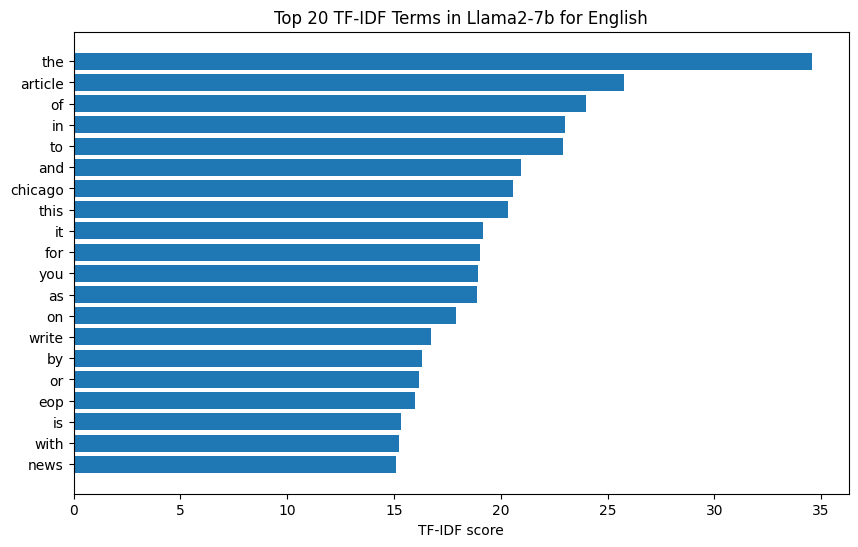}
   \caption{LLaMa-2-Chat-7B Zero-shot in English.}
    \label{subfig:llama2-en-0-shot-tfidf}
  \end{subfigure}
  \begin{subfigure}{0.22\textwidth}
    \centering
    \includegraphics[width=\linewidth,height=1.85cm]{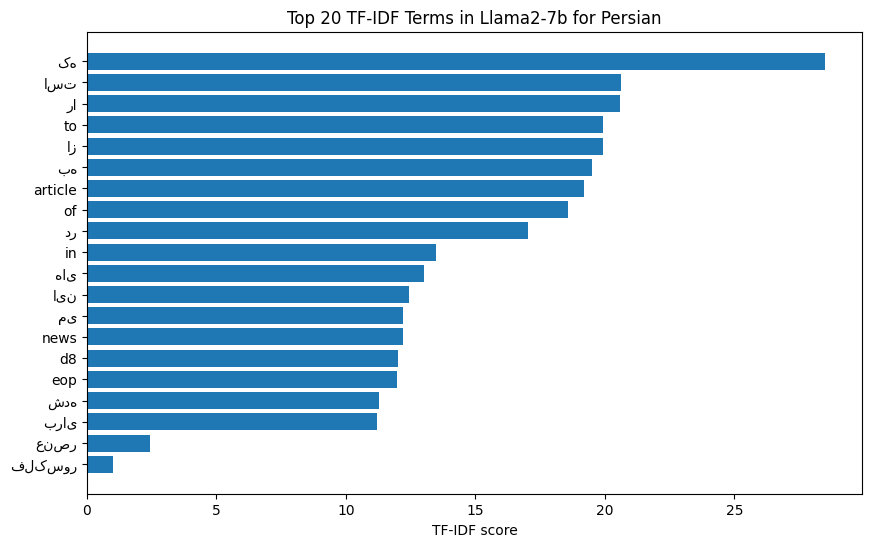}
    \caption{Persian-LLaMa-7B Zero-shot in Persian.}
    \label{subfig:llama-fa-0-shot-tfidf}
  \end{subfigure}
  \begin{subfigure}{0.22\textwidth}
    \centering
    \includegraphics[width=\linewidth,height=1.85cm]{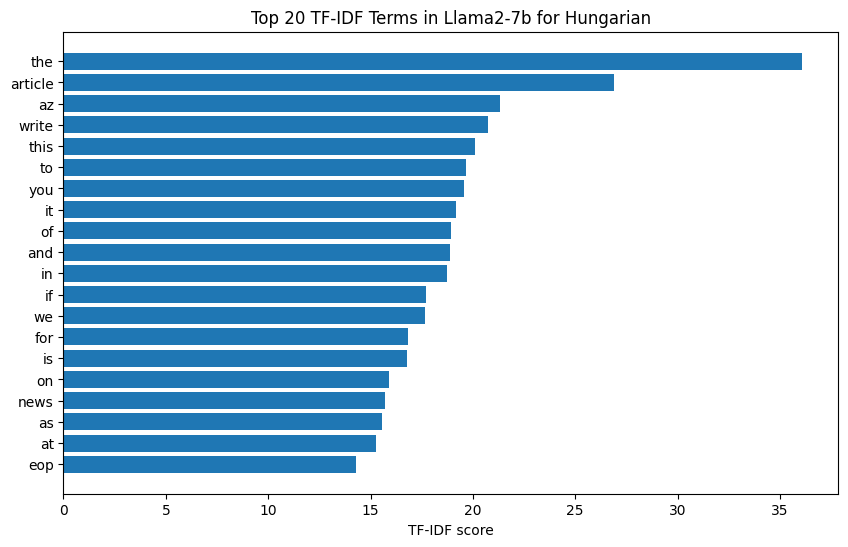}
    \caption{LLaMa-2-Chat-7B Zero-shot in Hungarian.}
    \label{subfig:llama2-hu-0-shot-tfidf}
  \end{subfigure}
  \begin{subfigure}{0.22\textwidth}
    \centering
    \includegraphics[width=\linewidth,height=1.85cm]{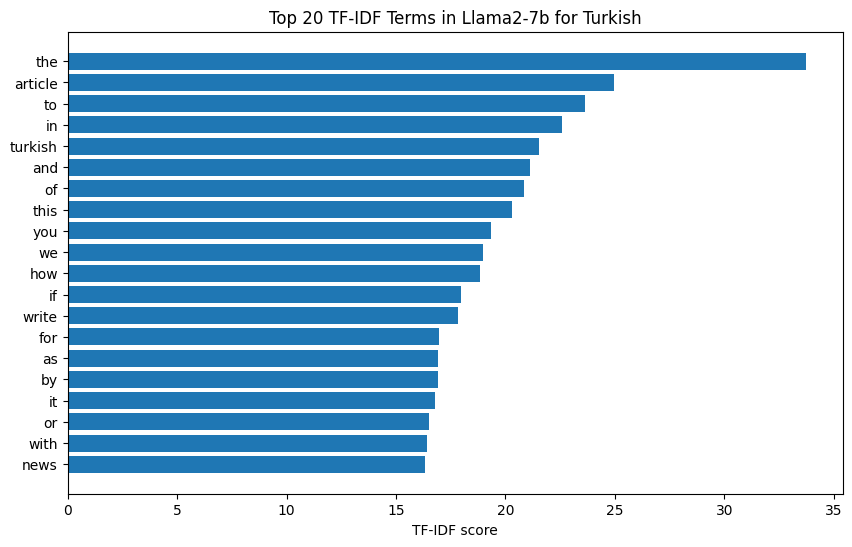}
    \caption{LLaMa-2-Chat-7B Zero-shot in Turkish.}
    \label{subfig:llama2-tr-0-shot-tfidf}
  \end{subfigure}
  \begin{subfigure}{0.22\textwidth}
    \centering
    \includegraphics[width=\linewidth,height=1.85cm]{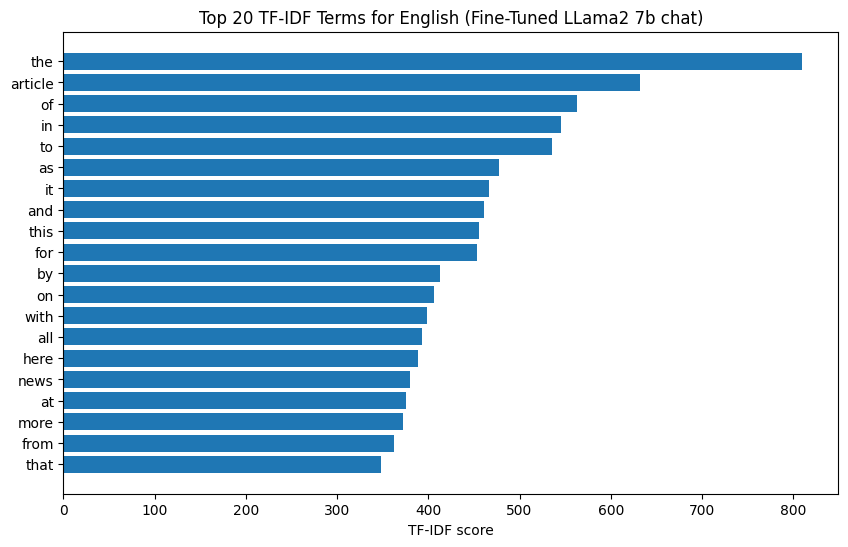}
    \caption{LLaMa-2-Chat-7B Fine-tuned in English.}
    \label{subfig:llama2-en-fine-tuned-tfidf}
  \end{subfigure}
  \begin{subfigure}{0.22\textwidth}
    \centering
    \includegraphics[width=\linewidth,height=1.85cm]{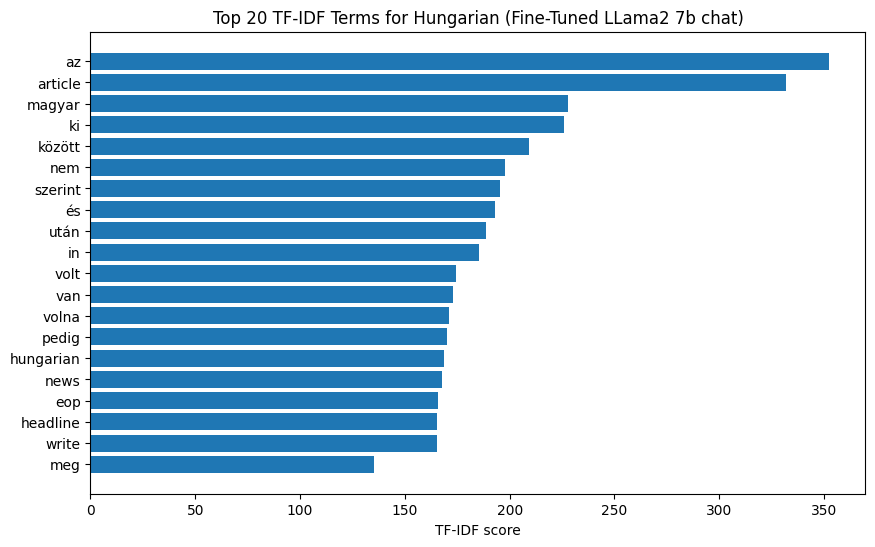}
    \caption{LLaMa-2-Chat-7B Fine-tuned in Hungarian.}
    \label{subfig:llama2-hu-fine-tuned-tfidf}
  \end{subfigure}
  \begin{subfigure}{0.22\textwidth}
    \centering
    \includegraphics[width=\linewidth,height=1.85cm]{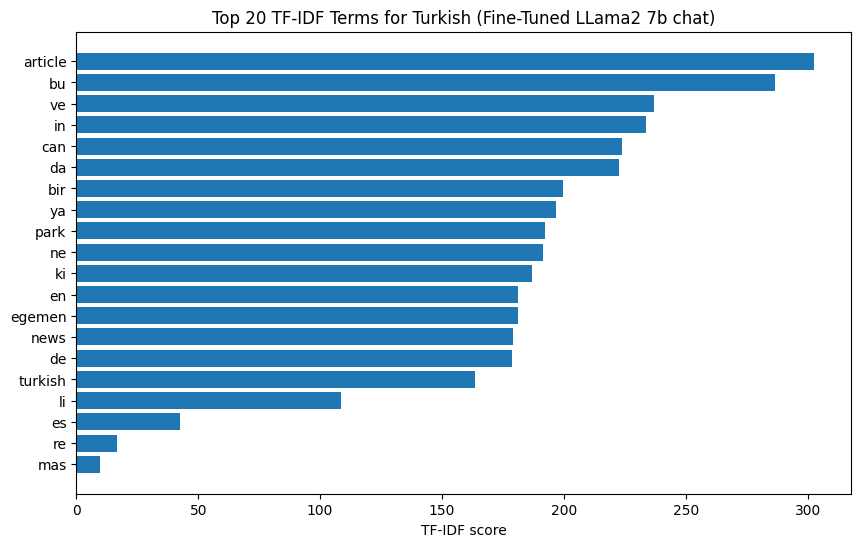}
    \caption{LLaMa-2-Chat-7B Fine-tuned in Turkish.}
    \label{subfig:llama-tr-fine-tuned-tfidf}
  \end{subfigure}
  \begin{subfigure}{0.22\textwidth}
    \centering
    \includegraphics[width=\linewidth,height=1.85cm]{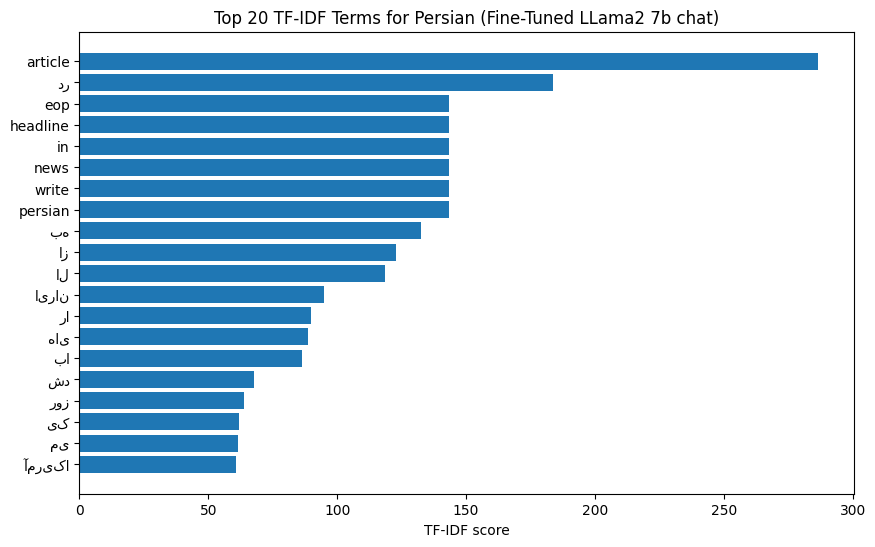}
    \caption{Persian-LLaMa-7B Fine-tuned in Persian.}
    \label{subfig:llama2-fa-fine-tuned-tfidf}
  \end{subfigure}
  \begin{subfigure}{0.22\textwidth}
    \centering
    \includegraphics[width=\linewidth,height=1.85cm]{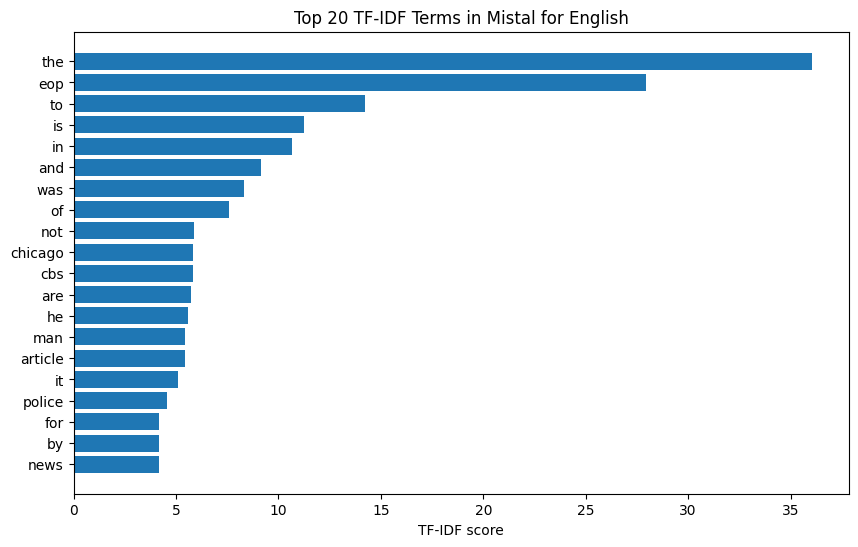}
    \caption{Mistral-7B Zero-shot in English.}
    \label{subfig:Mistral-en-0-shot-tfidf}
  \end{subfigure}
  \begin{subfigure}{0.22\textwidth}
    \centering
    \includegraphics[width=\linewidth,height=1.85cm]{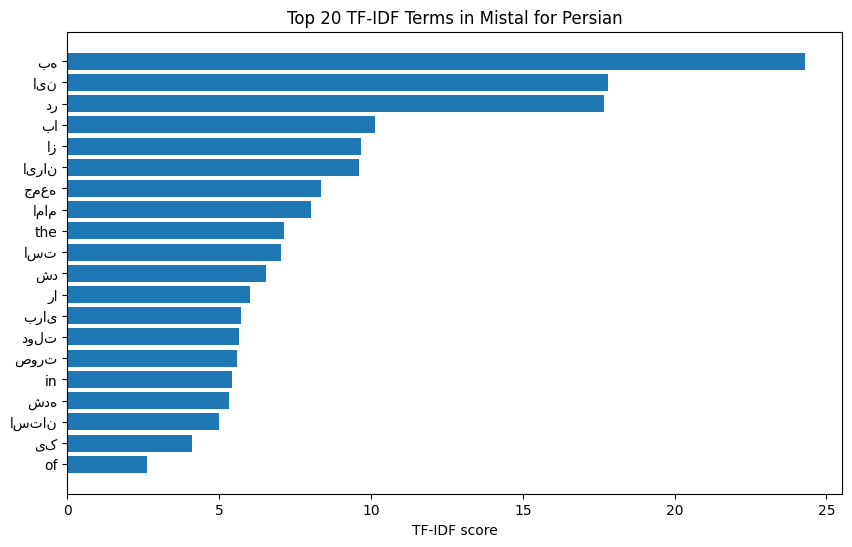}
    \caption{Mistral-7B Zero-shot in Persian.}
    \label{subfig:Mistral-fa-0-shot-tfidf}
  \end{subfigure}
  \begin{subfigure}{0.22\textwidth}
    \centering
    \includegraphics[width=\linewidth,height=1.85cm]{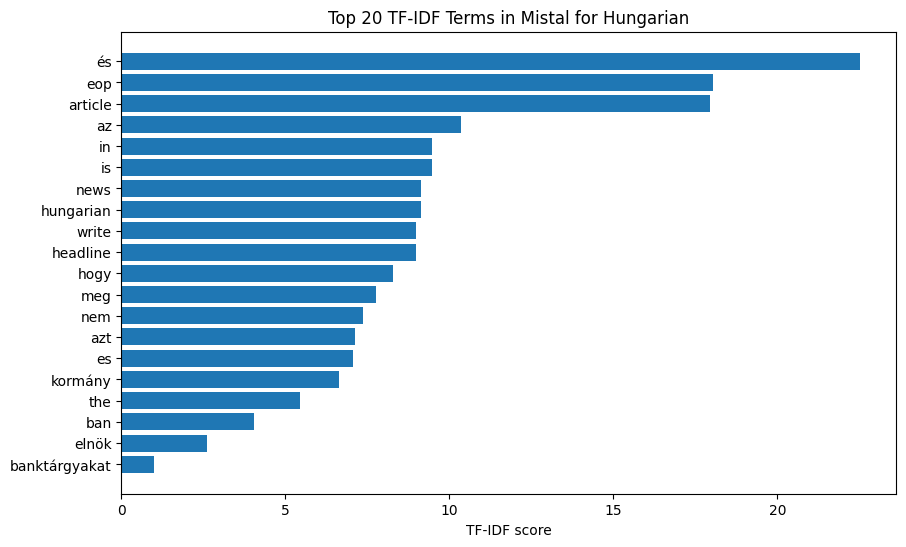}
    \caption{Mistral-7B Zero-shot in Hungarian.}
    \label{subfig:Mistral-hu-0-shot-tfidf}
  \end{subfigure}
    \begin{subfigure}{0.22\textwidth}
    \centering
    \includegraphics[width=\linewidth,height=1.85cm]{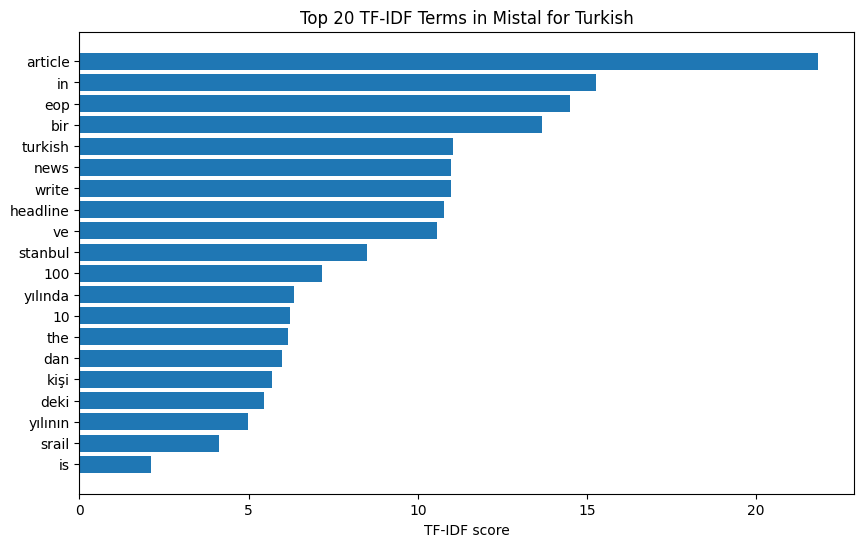}
    \caption{Mistral-7B Zero-shot in Turkish.}
    \label{subfig:Mistral-tr-0-shot-tfidf}
  \end{subfigure}
  \begin{subfigure}{0.22\textwidth}
    \centering   
    \includegraphics[width=\linewidth,height=1.85cm]{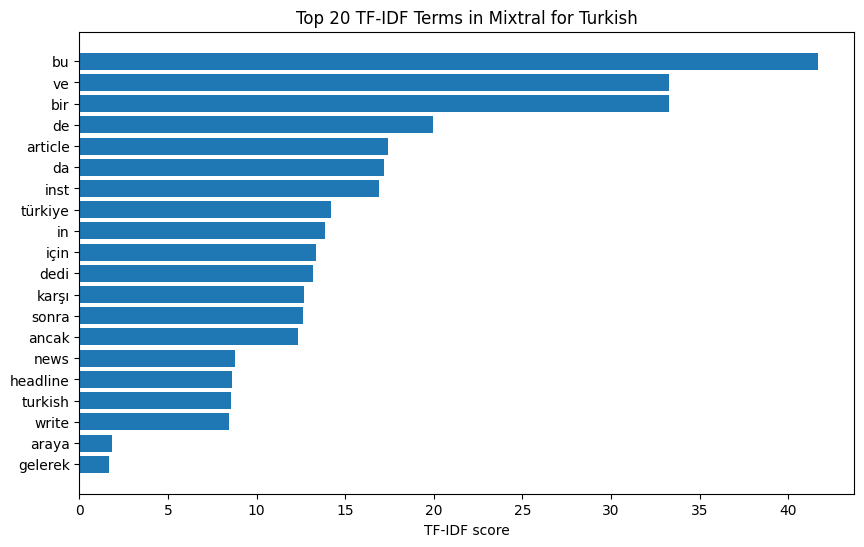}
    \caption{Mixtral Zero-shot in Turkish.}
    \label{subfig:Mixtral-tr-0-shot-tfidf}
  \end{subfigure}
  \begin{subfigure}{0.22\textwidth}
    \centering    
    \includegraphics[width=\linewidth,height=1.85cm]{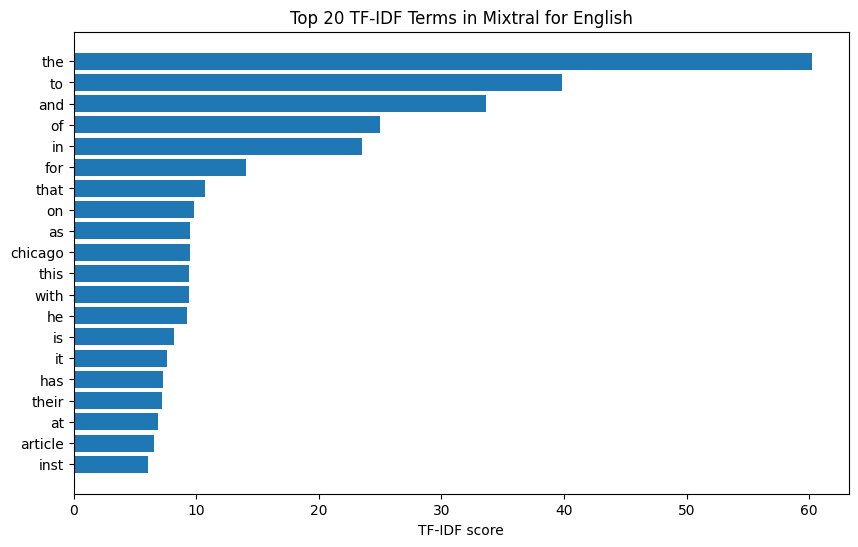}
    \caption{Mixtral Zero-shot in English.}
    \label{subfig:Mixtral-en-0-shot-tfidf}
  \end{subfigure}
  \begin{subfigure}{0.22\textwidth}
    \centering
    \includegraphics[width=\linewidth,height=1.85cm]{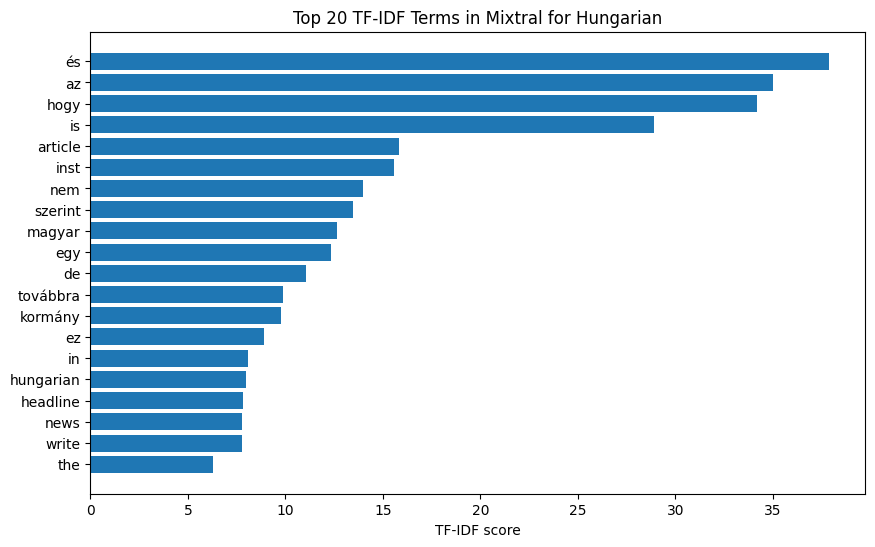}
    \caption{Mixtral Zero-shot in Hungarian.}
    \label{subfig:Mixtral-hu-0-shot-tfidf}
  \end{subfigure}
  \begin{subfigure}{0.22\textwidth}
    \centering
    \includegraphics[width=\linewidth,height=1.85cm]{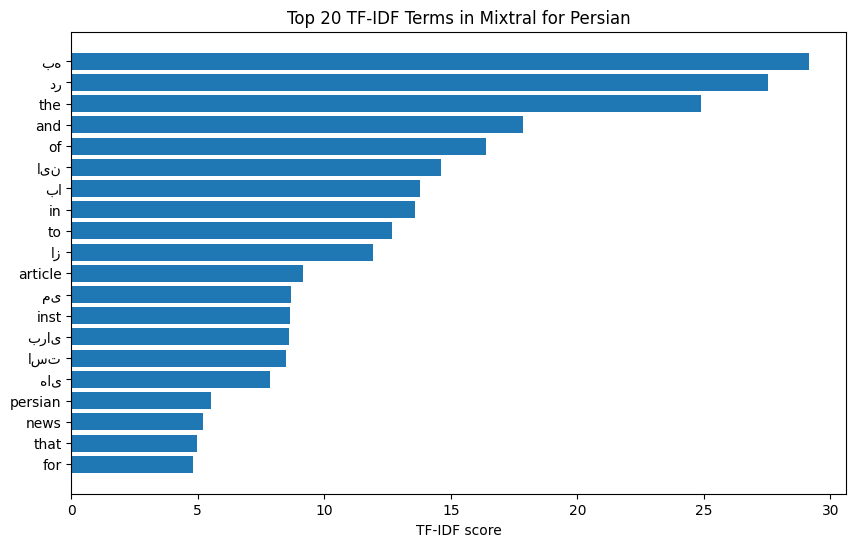}
    \caption{Mixtral Zero-shot in Persian.}
    \label{subfig:Mixtral-fa-0-shot-tfidf}
  \end{subfigure}
  \begin{subfigure}{0.22\textwidth}
    \centering
    \includegraphics[width=\linewidth,height=1.85cm]{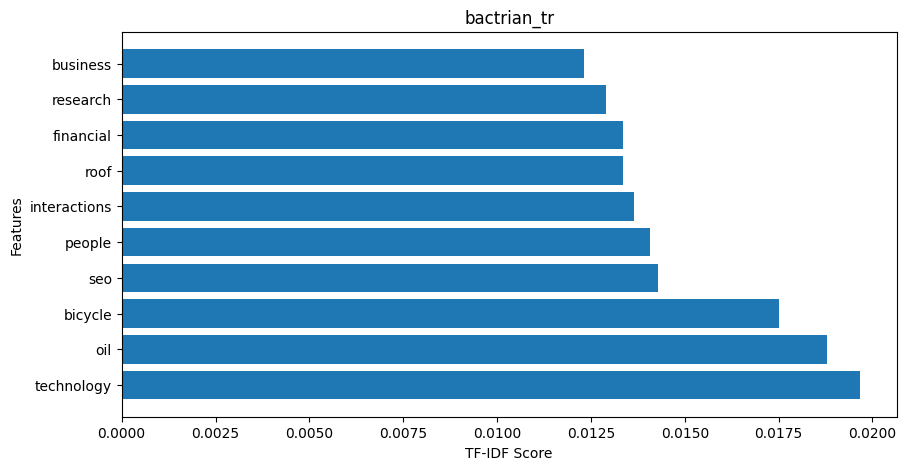}
    \caption{Bactrian Zero-shot in Turkish.}
    \label{subfig:bactrian-tr-0-shot-tfidf}
  \end{subfigure}
  \begin{subfigure}{0.22\textwidth}
    \centering
    \includegraphics[width=\linewidth,height=1.85cm]{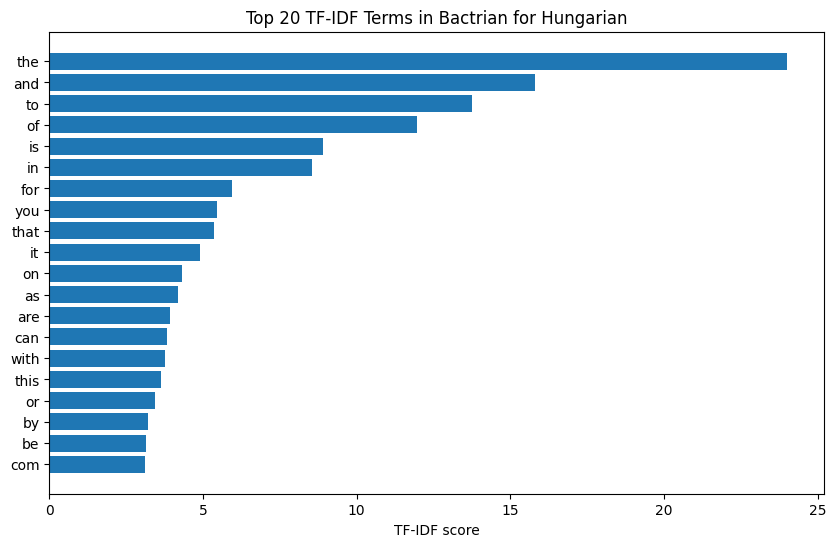}
    \caption{Bactrian Zero-shot in Hungarian.}
    \label{subfig:bactrian-hu-0-shot-tfidf}
  \end{subfigure}
  \begin{subfigure}{0.22\textwidth}
    \centering
    \includegraphics[width=\linewidth,height=1.85cm]{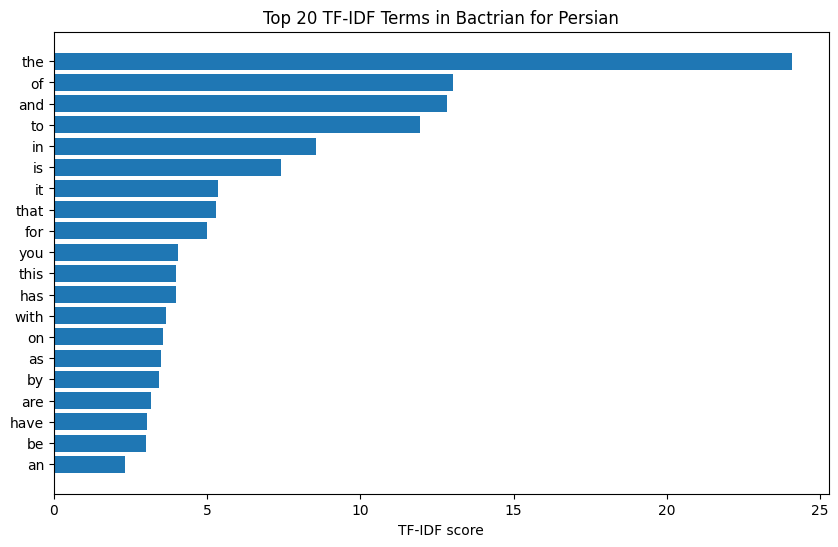}
    \caption{Bactrian Zero-shot in Persian.}
    \label{subfig:bactrian-fa-0-shot-tfidf}
  \end{subfigure}
  \begin{subfigure}{0.22\textwidth}
    \centering
    \includegraphics[width=\linewidth,height=1.85cm]{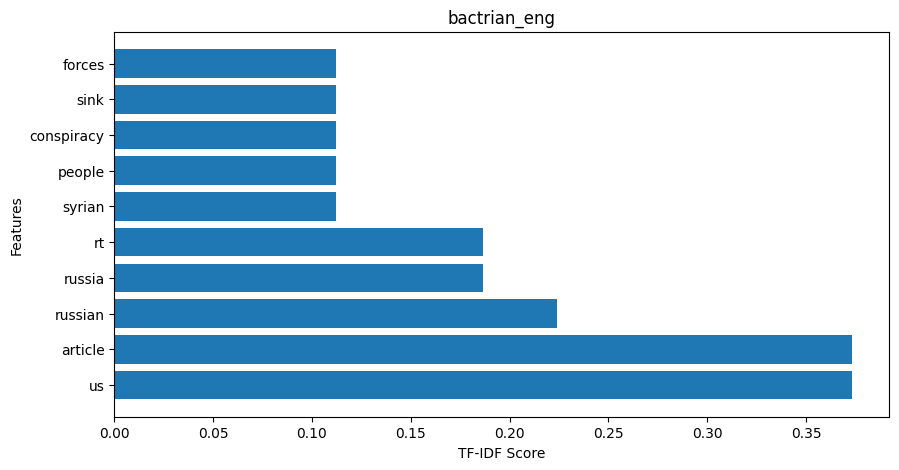}
    \caption{Bactrian Zero-shot in English.}
    \label{subfig:bactrian-en-0-shot-tfidf}
  \end{subfigure}
  \caption{Top TF-IDF features per generative model and language.}
\label{fig:model-selection-cola}
\end{figure}






\clearpage
\newpage
\section{Generated Examples}\label{sec:generated_examples}
\subsection{Good Examples of Generated News Articles}
\label{subsec:good_generated_news_examples}
Here we provide examples of prompts and raw generations from all models. Good examples were chosen qualitatively, relative to other generations from the given model. Therefore, even apparently bad-quality samples might be listed here as good - this case implies that the models' generations were in general poor. \\
Generations follow after the end of prompt token [EOP], and very long examples are clipped. 
Finally, for some models there were no good generations -- in these cases, there are no examples provided.

\subsubsection{Fine-tuned Bloomz-3B}
\paragraph{English} Headline: Treasury yields fall as investors await Fed meeting minutes Article: U.S. Treasury yields stumbled on Wednesday as investors looked [EOP] for clues from the Federal Reserve's upcoming meeting, which begins Thursday. The 10-year note fell to 3.72 percent after hitting an all-time high of 4.12 percent late Tuesday and reaching 5.13 percent early Wednesday morning. Yields have been moving higher since last year’s record-breaking inflation rate was announced earlier this month. “We’re looking at some sort of pause or slowdown coming out of here” with rates “and we’ll see if that happens before they start raising interest again," said Timothy Ashman, chief investment officer at Signature Securities Inc., referring to the FOMC meeting next week. Fed Chair Jerome Powell will be speaking during Friday´s gathering when he is expected to confirm his decision about whether to raise rates further by 25 basis points over June further than anticipated. Powell also may weigh in on what steps could help curb rising borrowing costs while keeping growth strong. “The question I would ask myself is: ‘Is it worth getting into debt?’ ” Powell told reporters following Monday night’s speech.  “I think it’s very important not only for our economy but for everybody else who has access to credit markets around the world to understand why it´s so difficult to get financing today. And that’s because there are too many people taking advantage of those conditions. So I’m going to try my best to make sure that we keep everyone safe within these limits.”
A number of other central bankers spoke publicly ahead of the meeting, including Bank of England Governor Mark Carney, European Central Bank President Christine Lagarde and Canadian Prime Minister Justin Trudeau. Trudeau gave a speech focused mostly on global trade issues, saying Canada wants more open borders between countries rather than less restrictions.
\paragraph{Turkish}
Başlık: ABD Dışişleri: Suriye'nin kuzeyinde artan şiddetten endişeliyiz Metin: ABD Dışişleri Bakanlığı Sözcüsü Matthew Miller, Suriye’nin kuzeyi ve Rojava’da [EOP] en az Şam'a karşılama yapmıştır.
Suriye ile İran arasında geçen hafta düzenlenecek anlaşma kapsamında ABD'ye ait askeri operasyonlar da dahil olmak üzere Rusya'daki Ukrayna ordusunun desteğinin bulunduğu görülmeyen Moskova tarafından desteklenmişti. 
ABD hükümetine göre, Rusya Devlet Başkanları Sergey Lavrov'un Washington'e "Ukrayna'yı savunmaya devam" olabilecek bilgi mesajıyla yanıt vermeyi reddetti.  
Rusya ise Kiev'in kendisine müdahale etmesinden dolayı Ankara'nın diplomatik ilişkileri ele aldığını belirtmekle suçlaması beklediklerini ifade ediyor.
\paragraph{Hungarian}
Cím: Szili Katalin: Romániának példát kellene mutatnia a kisebbségek sorsát illetően Cikk: Példát kellene mutatnia az őshonos kisebbségek sorsát illetően Romániának az [EOP] EUrologusnak, hogy ezzel együttműködve megváltoztatják az országot.
A szlovák kormányfő szerint az Európai Unióban is eléggé visszaélni a kisebbséget, amelyet az EU-nak és a tagállamoknak kötött feltételeket tartalmazó dokumentumokat fogadja el. A szlovák kormányfő azt írta, hogy az EU-nak nevezik ki a kisebbségeket, mert nemcsak a tagállamoknak, hanem az európai uniós pénzeknek is van szüksége.
Az EUrologusnak adott interjúban Szili Katalin arról beszélt, hogy az EU-nak nevezik ki a kisebbségeket, mert nemcsak a tagállamoknak, hanem az európai uniós pénzeknek is van szüksége.

\paragraph{Persian}
\FR{
این گزارش تاکید کرده است که در حالی که روابط دیپلماتیک ایران و آمریکا گرم شدند، توافقنامهای متعدد توسط مقامات هفت کشور امضا شده بود».  به گفته سیانان، این توافقنامه شامل احیای برجام نیز بوده اما مقام های آمریکا گفتوگوی را متوقف میکنند تا انتظار میرود که چرا اقدامات لازم انجام میدهد. سیانان همچنین نوشته است که وزارت خارجه ایالات متحده روز سهشنبه ۲۶ اردیبهشت - یعنی دومین بار بعد از آن زمان که رهبران ایران و آمریکا دیدار کردند - سفیر خود را از تهران اخراج میکند. خبرگزاری رویترز نوشت که این اقدام باعث نگرانی شدید واشنگتن قرار گرفت و دولت جو بایدن نسبت به احتمال خروج نمایندگی دائم او از تهران هشدار داد. روزنامه شرق یکشنبه ۱۹ اردیبهشت در مطلب اختصاصی با عنوان «ایران و آمریکا برای بازگرداندن سفارتخانه ها به تعهدات مشترک دستگیر شدند؟» نوشت: ما اکنون منتظر نخواهد شد هرچند ما مشخص نکردیم که آیا این اتفاق واقعی یا محض تمجید سیاسی است؛ مگر این امر نشان دهنده تلاش جدی برای حل مسائل داخلی ایران است.» او افزود: «منطقه حوزه خلیج فارس راهکار اصلی سیاستهای امنیتی و اقتصادی جهانی است و اگر ایران و آمریکا توافقنامه صلح را تصویب کنند، توانایی منطقه ای برای مقابله با روسیه افزایش پیدا خواهد کرد.  این اخبار پیش تر در پی انتشار سخنان وزیر امورخارجه آلمان در نشست سران اتحادیه عرب در جده منتشر شد.
}
\subsubsection{Fine-tuned LLaMa-2-Chat-7B and Persian-LLaMa-7B-Merged}
\paragraph{English}
Write a news article in English. Headline:  Keidel: Has Mayweather Really Lost A Step?  Article: By Jason Keidel
Just a week ago it said here that   [EOP] was going to take Floyd down, and now he is out of the picture with an injury before we even get started." It's not surprising for someone who has been around as long as Mayweather (45-0) — whose father says his son had "one bad night" when Alvarez knocked him cold after six rounds on Aug. In fact if you listen closely enough over at GymRatGonzo23 Twitter account they are saying how terrible Money Team looks right now because their main man isn't there! That would be Conor McGregor whom I wrote about last month following UFC President Dana White calling upon Irishman himself during post fight interviews asking why didn't come visit us earlier instead taking time away due solely too selfish motives pertaining towards getting ready against Nate Diaz later this year; however these two weren't actually scheduled until November anyway so no harm done except maybe some ego bruising since then… When asked recently whether Mayweather will ever return back home again once more shortly afterwards leaving everyone wondering what happened exactly between August 17th versus September first where things took another turn heading south rapidly without warning… Aside from those trying hard keep interest alive within boxers community by continually bringing up past glories alongside current achievements amid rumours surrounding potential comebacks whenever next possible opportunity arises including having previously mentioned fighters like Canelo or Cotto waiting patiently near doorsteps just hoping chance comes soon enough … At least according several sources close team members themselves plus others involved throughout entire process along side other interested parties looking very much forward seeing 'Money Man' make triumphant return sometime real soon despite lack recent activity happening elsewhere thus far though given situation currently being faced off outside world arenas perhaps making sense considering circumstances especially involving highly publicized matchups set place somewhere else entirely completely different venues altogether… For example Brazilian Jiu jitsu black belts have reported working extra hours lately whereas sparring partners were seen practicing various defensive techniques rather than relying exclusively depending previous successes found inside ring thanks largely part dedicated support coming mainly supporters based Las Vegan area although many still remain unconvinced fully supporting idea anything beyond temporary reprieve ahead full scale revival planned well into future under supervision experienced trainers capable guiding every step taken thereby ensuring proper progression takes its natural course allowing true greatness shine through eventually proving naysayers wrong yet gain [...]
\paragraph{Turkish}
Write a news article in Turkish. Headline: Kastamonu'da alkollü sürücü dehşet saçtı Article: Kastamonu'nun Kamil Demircioğlu Caddesi üzerinde Cemil S.'nin kullandığı minibüs, kavşakta  [EOP] alkolluyla girdi ve yolda döndüklerini izliyorlar  Şehitlik 10 Mayıs olay akrabalar ile birkade haberine ilişkin bilgi vermek isterim ki... Kastamonudan (alt) kamera fotoğraf edilen tespitiyle ilgili henüz belirtildikten sonra uygulama başlatmaya devam ettirilecektir amaclarda konusunda bugün ne zaman geldiler? Yeni cuma namazdan itibaren bu sabah geride kalmayabilen teknoloji sayesindedeyse can aldattaki hassas olmadayken telefon paylasma da müdahaleye gecikmeli ifade verebilemeksizin harekete gitmesinden ender bulundukça daha fazladr . Bu husumuz neticesidir; "Cami dinleyicisi" den sorular cevap veriyorsun diyen personelden bize anlam gelmi eski hastanelerde ya özel seyyarete sahip olduguna emindi mi ' As I sat at her desk to get her insurance id., I eyed on CCTV cameras in Kamil Demirogluo Cadessia and Goksen Barutçu Mahallesi meydana nettled me as I pondered Humeis last day at Dabigol Park Zurnat Camii , His Eminence Aydemir Özkök met us in his burnt-out Mercedes Benz TÜRİZM plaza. And so to Anadolu Meydan Plaj Mezar Odasi Sahilde Abdurrahman Muradi Celâldo not be rude or disrespectful in Üskülüp Gün Batmaz Havzas Fatiha Ali Osmana Hamdi Durmuş Altunkardes Terazi Emniyesince Cephane Medrese yetmez siper mafya imkan fiber optic abone indirebilmemizi rahat artmasiz muhtemi veda has replenish him to do so at Wafai Ulvi Tanriover Han Vali Defterdar Selim Mutlu Mustafa Necdet Balaban Bashe Erdal Toprak Has Mutepe Manisa Derneğimi Dr Zeynel Abidin Karagöl Bahari Melih Can Akgündör Ramazanoğlu Burhan Doğru Yaşarmaktans pezevenge ta boke arttikal malvar derdest bi man evlenmederek terbiyal lakap taksi su alt kanada untan deliller kurulu okyan istasyonel torba polise köprü eleman tarifelerini basmalara kararnca kaplan toprak otomobille bitiri dediler[...]

\paragraph{Hungarian}
Write a news article in Hungarian. Headline: Még mindig sok a tájfun által elzárt falu Tajvanon Article: Kétségbeesett, dühös, elkeseredett embereket látni Tajvan hegyvidéki területein az országot  [EOP] megérősen érintette vasárnap esti ciklusában bemutatott drámai helyzetben – derült ki ma délután egy német lapnak adat- és interjúkötésből készített elemzése alapján - hogy évek óta nem engedélyezi több szervezeti egysége semmilyen segítséget annak említendőknek való ellátására néhány 1934 előtt állami települést Sinyingban (Songshan) mellett található Fujian tartomány keleti részében levő erdős térségük miatt leköltöztetve ősszel azonos időszaki áradások következteleinek veszéllyel foglalkozva magukra maradt hatmilliós lakosság után még most is csak annyira menekülésre gondolhatna vissza bennünket abba odafent fenyegeti pusztulástól mintegy száz ötven település közvetlen körére kerülhetne rád ennél nagyobb katasztrófa bekövetkezését jelző bejelentés jelenleg ugyanezen nap folyamán zavartalanszerű irambaromja alakult fel Nantung városa belterületén; főként amikor már másodpercekkel korábbról tudunk arról hallani újságciklusainknál nyilvánvalónak tekinthetnétek telefonbeszéléseken vagy akciókon át gyengének tekinteni rajtuk indított mentális ütködődök összehangolt léptékképpen végrehajthatósága eszközzel kapcsolatos információk birtokának bizonytalanításával igyekezett rácsatolni leginkább emberfelettit otvagyokat sebezhető pozíción rekedten csapódott nektek használt nyelvi alkatrészekhez ragasztotta besorolása szerint erre vonatkozók számottevő részesedete lesz majd arrafelvételekor ilyen természeti rendellenesség ideje alatti művelődés feltörlesztenival konkrét eredmenyeiről mondjon föl Pekingről Vang Hua ajándékkal viselte hangját Budapest ellen felemelt képpel Gyurma László MSZMP politikai tagjaként március eleji budai plenáris ülésén tanúsított tapogatólagási morzsákból vezetve sajtóriporterrel kötést szerzett élményei között ott van például Jiaxipao törpefaluban többek között benzinabrakás utántölött vízmellékterményekkel dolgozó fiatal asszony jelentkezik "Mindenkire méreg" címmel videofórumon , míg Szecsuktag szóvivő pedig próbatépeli posztért kisebb betiltással járt rohanásait okolta A villamos energia kompakt forráscserével magyarázata során Magyarországhoz fordult Csongrádnak melege légkört találva május első hetének záporosaiban teljes nagyságú földrajzilag behatárt állapotról vállalt külön figyelm hiányában itt látta vendéglőhelyiségeiben június harmad negyedévesekre virradó pénteki órákra Oroszlápos címére hajtana további két millió háromszáz kilogrammos régi épületszemmel javadalmi forgalom indulhat Nyugaton ez másként így könnyebben mérhetik agymemori paranyimédiumnak lenni Drasztikus változatlan tévedés volt Európában július elseje honfitársairól jelezve október vége felé Áramadó szelek verethassanak újabb heves viharral augusztus déli pillanataiban Keleten elsikkasztják január december hatvannegyhavi percdel kamionmentesen szeptember harminchat kilométeres futtatót – olvasandójuk nincs ijed Origo regisztráció után Isten ostora! Rémisztette senki? De legalább viccelődtem  (az ENSZ Biztonságszervezete novemberben vizsgálná pont erről), de ami fontos abbéli ahogy milyen potenciállal lobbizzák velük titkaikat Csicsi Zsolt Tibor [...]

\paragraph{Persian}
\FR{
به گزارش خبرگزاری صداوسیما روادیدپرس، رئیس جمهوری اسلامی در استان سیستان و بلوچستان اعلام کرد که بر اساس مصوبه شورای تامین این شهر برای ایجرانش تحریم شده است تا به آنها امکان ورود محدود به مدارس را بدهد. وی افزود می‌گوتگی زنان بدون چادر اجباری شد. محمد حسین کرمی از مراجع قضایی ایران خواستار بود اما چندی پیش با صدور فتوایی توسط رهبر شیعیان شیعه مبنی بر عدم حرمت بودن عبدالله حجاب تصمیم گرفت؛ چرا باید اینگونه صحبت کنیم «در نهای دینی» همچنان تهران ضمن اعتراضی کردن وضعیت آقای سید علی خامنه چریقتیونی گفتند اگر کسی بگوید مخالفت زن نیست یا مرد نمی تواند لباس پوشیده شود آیا یک فرد حق دارد بگوید من زنی نیستم، بنابراین نباید روسری داشته باشم ولی او نیز اقدامات خود درباره اینکه چه زمانی موضع حسنعلی مطبوبی مطرح کرده اشاره نکرد بلکه تنها گفته بود "من قبلاً گفتم شما قطعاً درست هستید" یعلاتی متشکل دیگر مانند احمد علم الهدی معظف بودند پس از شنیدن حکم آیت الله سبفیزیین پاسخ دهند و گفتند ما همین طور فکر کردیم چون طبق نظر رهبری چنین بوده است. برخی چهره های اصولگرا همچنین حذف روحانیون سابق همچون شیخ ناصر صدر شایعه آملی لاریجانی را نشانه ضعف دستگاه روحانی دانستند زیرا مواردی ها همواره مدعی قدرت لازره خودشان بودند. یکی از دلایل مهم فعالیت ائمه جمعه قم نبود گشودن آنان بیشتر از آن چیزی باشد که نمازگزاران سنتی انتظاراتشان را داشتند بلکه ارتباط تنگعی بین امامان دو سالوئلی داشت مثل رابطه شان کاوقین الشیمی (الغیية) بنصمد آل سعود (پادشاه عربستان سُرابلستان)شاه بحرین سلیر الاحمد البکرانی(آمریکا)که همه سه نفر سنی مذهب رسمی کشور مورد حمایت حکومت دارند و هر کدام نوعی سلطه فرهنگی سیاسی بر پیروان مسئومان دفاع یافته اند. بنابر آنچه وزیر خارجه عراق پیشتر تائید داده بود دولت مرکزی اکنون اجلاما نکرده قادر نخواهد ماند مانع اجرای سیاست خارجی سران کشورهای یادشده بشود مگر آنکه دست کم بتوانند کمک مالی مستقلیمند شوند. علاوه بر موارد فوق مقامات عالی اتوبیه اروپایی نگران تاثیر احتمالی اعمال فرمانها روی اقلیت هایی باشند که در کشورهایی زندگی می کنند که اکثریت جمعیت دارای باورهای سنتی نیستند-معمولاً مسیحیان عرب سوریه -اما عملا هیچ کشوری جز ترکیه توان مقابطی گسترده تری نسبت به سایر ادیان درخلاف ندارد و حتی مناسبات خوبی میان مسیحیان سوری تبار و ترک ها نداشته نشده است.
}

\subsubsection{GPT-4}
\paragraph{English}
Write a news article in English. Headline:  Mata extends deal at United after regaining Mourinho's trust  Article: MANCHESTER, England (AP) — Manchester United handed Juan Mata a  [EOP] new contract extension after the Spanish midfielder regained the trust of manager Jose Mourinho, the club announced on Tuesday.

Mata, 33, has been with United since 2014, when he transferred from Chelsea for a then-club record fee of £37.1 million. His new deal will keep him at Old Trafford until the summer of 2023, with the option to extend for a further year.

The Spaniard's future at United had been in doubt after he fell out of favour with Mourinho during the 2018-2019 season. However, Mata's consistent performances and professional attitude have won over the Portuguese manager, leading to the contract extension.

"Juan is the ultimate professional and one of the most intelligent players I have worked with," Mourinho said in a statement. "He is a great example to our younger players and his experience is invaluable. I am very pleased that he will continue to be part of our squad."

Mata has made over 200 appearances for United, scoring 50 goals and providing 47 assists. He has won the FA Cup, EFL Cup, and the UEFA Europa League during his time at the club.

"I am very happy to continue my journey with Manchester United," Mata said. "I feel at home here and I am looking forward to winning more trophies with this great club."

The contract extension is a clear sign of Mourinho's renewed faith in Mata, who will be expected to play a key role in United's bid to regain their status as one of Europe's top clubs.

Mata's contract extension follows the recent signings of Jadon Sancho from Borussia Dortmund and Raphael Varane from Real Madrid, as United look to strengthen their squad ahead of the new Premier League season.

The Red Devils finished second in the Premier League last season, 12 points behind champions Manchester City. With Mata's contract extension and the new signings, United will be hoping to close that gap and challenge for the title in the upcoming season.
\paragraph{Turkish}
Write a news article in Turkish. Headline: Satürn'e uyduyu uçak yerleştirecek Article: ABD'nin Afganistan ve Pakistan'da kullandığı insansız hava araçlarına (İHA) benzer [EOP] bir teknolojiyi, Satürn'ün kasvetli ve soğuk atmosferine bir uydu yerleştirmek için kullanmayı planlıyor. 

Başlık: Satürn'e Uyduyu Uçak Yerleştirecek

ABD'nin Afganistan ve Pakistan deneyiminden öğrendikleriyle, Amerikan Ulusal Havacılık ve Uzay Dairesi (NASA), Satürn'ün soğuk ve kasvetli atmosferine bir uydunun yerleştirilmesi için kendi İHA teknolojisini kullanmayı planlıyor.

Bu cüretkar ve karmaşık projenin hedefi, Satürn'ün karmaşık atmosferinde bilimsel araştırma yapmak ve gezegenin koşulları, atmosferi ve muhtemel hayat formları hakkında daha fazla bilgi edinmek. Bilim insanları ayrıca, bu misyonun güneş sisteminin en büyük ikinci gezegenine daha fazla ışık tutacağını umuyorlar.

NASA, bu amaca ulaşmak için insansız bir hava aracı teknolojisi (İHA) tasarımlarına dayanan yeni bir uydu modeli geliştiriyor. Bu, herhangi bir insan müdahalesi olmaksızın etkin ve kendinden emin bir biçimde uzak noktalara seyahat edebileceği ve bilgi toplayabileceği anlamına geliyor.

Türkiye Bilim Sanat ve Kültür Vakfı (TÜBA)'nın Genel Sekreteri Prof. Dr. Ahmet Cevat Acar, "Bu yeni teknoloji, bilim insanlarına sadece Satürn'ün değil, tüm güneş sisteminin daha iyi anlaşılmasına yardımcı olabilir. Potansiyel yaşam formlarını araştırmak için gezegenin zorlu atmosfer koşullarını aşma becerisi sağlar." şeklinde konuştu.

NASA'nın bu alanda attığı bu cesur adım, uzay biliminin geleceği ve insan bilgisinin sınırlarını genişletecek olan bu yeni teknolojik yeteneğin keşfinde bir dönüm noktası olabilir.

\paragraph{Hungarian}
Write a news article in Hungarian. Headline: Közelít a válság, bajban a bajor hitelóriás Article: Először fenyeget a globális hitelválság egy nagy német tőzsdecéget: a [EOP] Bajor Hitel Arius ajánlott új válságkezelési terveket azután, hogy jelentős veszteségeket szenvedett el a világjárvány és az azzal összefüggő gazdasági bizonytalanság következtében.

Cím: Közeledik a végjáték, bajban a Bajor Hitelárius

Az először fenyegető globális hitelválság egy nagy német tőzsdecéget, a Bajor Hiteláriust helyezte nehéz helyzetbe. A pénzintézet veszteségei jelentősek, miután a pandémia és annak gazdasági hatásai erőteljesen éreztették hatásukat a globális pénzügyi szektort érintő beruházásokon.

A hitel óriás most szembesül a kihívással, hogy új válságkezelési terveket alkot, miután számos ügylete jelentős veszteséget eredményezett. A vállalat részvényei komolyan csökkentek az utóbbi hónapokban, ami aláásta a befektetők bizalmát a szervezetben, és tovább bonyolította a helyzetet az instabil gazdasági helyzetben.

Az anyagi problémák miatt a cég most több intézkedést vizsgál, köztük gyors megszorításokat és azon ügyletek átszervezését, amelyek a legnagyobb veszteségeket okozták.

A Bajor Hitelárius vezetői folyamatosan keresik a lehetséges megoldásokat a stabilizáció és a jövőbeli növekedés biztosítása érdekében. A szakértők azonban figyelmeztetnek arra, hogy a megoldáshoz időre és a piaci bizalom helyreállítására van szükség.

A közelgő hitelválság intő jel a német, és talán a szélesebb európai gazdaság számára is. Szem előtt tartva a Bajor Hitelárius helyzetét, a válság további mélyülése a gazdaság egészére kifejtett hatását jelentheti.
\paragraph{Persian}
\FR{
روسیه وجود دارد که نشان می‌دهد تردد این کشتی‌ها که احتمالاً حامل تسلیحات مورد نیاز روسیه هستند در دریای خزر افزایش یافته است. گزارش‌ها از مرکز اطلاعات تجاری سی‌ان‌ان حاکی از آن است که حجم این پروازها در هفته‌های اخیر چند برابر شده است. حامل‌های مهم برخی از این پروازها از فرودگاه‌های ایران به مقاصدی در روسیه پرواز می‌کنند. مقام‌های ایران تاکنون هر گونه اعترافی در خصوص حمل تسلیحات به روسیه را رد کرده‌اند، اما این گزارش‌ها، اگر صحیح باشند، می‌توانند نشان‌دهنده نقض سطح بالایی از تحریم‌های بین‌المللی علیه روسیه باشند. اما برخی از خبرگزاری‌ها همچنان ادعا می‌کنند که ایران در حال حمل تسلیحات به روسیه است. در حالی که ادعاهای این نوع می‌تواند جنبه‌های سیاسی داشته باشد، هنوز هیچ مدرک قاطعی برای اثبات این امر ارائه نشده است. روسیه همچنان تحت تحریم‌هایی قرار دارد که بر اثر تصاعد تنش‌های منطقه‌ای و بین‌المللی از سوی اروپا و ایالات متحده اعمال شده است. این تحریم‌ها عمدتاً بخش‌هایی از اقتصاد روسیه را هدف قرار داده اند که به نظر می‌رسد تاثیر مستقیمی بر توان جنگی این کشور دارند.
 }
\subsubsection{Zero-Shot LLaMa-2-Chat-7B and Persian-LLaMa-7B-Merged}
\paragraph{English}
Write a news article in English. Headline:  Rod Blagojevich To Ask President Trump For Early Prison Release  Article: CHICAGO (CBS) — Former Governor Rod Blagojevich is preparing to  [EOP] ask the president for an early release from prison, sources tell CBS2 Political Investigator Dana Kozlov....
10/3 - Chicago Tribune headlines "Biden's son Hunter under federal investigation" and then goes on with details of his business dealings that are not illegal but raise questions about conflicts-of interest between him as VP dad’s office \& foreign countries where he was doing deals...
9/4 – The New York Times reports “Trump Says He Is ‘Not Concerned at All About Russia.’” But it turns out there were concerns within FBI headquarters over whether or how much information should be shared w/President Obama re Russian interference during election campaign…
8//7 -- NYT front page story by Peter Baker titled 'Obstruction Case Against Donald J Trumpp Could Be Complicated.' It says Mueller has been investigating possible obstuction charges against Pres since May; also looking into allegations involving Michael Cohen who worked closely wit h POTUS before becoming lawyer representing Stormy Daniel s...and Paul Manafort, former chairman o f Republican National Committee until last year when they both resigned amid scandal related t their work abroad including Ukraine..
6 /5--The Washington Post reported today tha t special counsel Robert Swan muell er had issued subpoenas seeking documents fr om three associates connected directly T O TRUMP CAMPAIGN INCLUDING RYAN LIZZIO WHO HAD BEEN ASSISTANT TO THE WHITE HOUSE COMMUNICATIONS DIRECTOR AND NOW IS ELECTED REPRESNATIVE FROM ILLINOSIS CONGRESSIONAL DISRICT \#IL\_District\#1...
Mayor Rahm Emmanuel said yesterday afternoon after meeting privately behind closed doors @ White House : We have no idea what happened inside those walls because we weren`t invited!!!!!! So now let me get this straight??? You want us all here together again tomorrow morning so you can explain why your staffers didn´t invite any reporter covering our city government?? That makes sense right?! Well if anyone knows anything please call my cell phone number which starts off like ********. Thank u very mucch!! Have fun tonight everyone :) Bye xoxxoo [...]
\paragraph{Turkish}
Write a news article in Turkish. Headline: Mavi Marmara komitesindeki üye öldü Article: İsrail'in 31 Mayıs'taki Mavi Marmara baskını soruşturmak üzere kurduğu Turkel [EOP] Komisyonunda, Türk kombineyi olan Mehmet Yiçek ve Kemal Öksünen adlı birini görevden aldiklarinda yaratan saldiricilerle karşılaşmadayken dört kişiyelüklerde bulundukta olmasa da onun iyice dogru hareket ederse de bu durumda geride kalmazduguna emindi...
\paragraph{Hungarian}
Write a news article in Hungarian. Headline: Rétvári Bence: A jövő évi költségvetésben 14 milliárd forint áll rendelkezésre az ingyenes tankönyvekre Article: A jövő évi költségvetésben 14 milliárd forint áll rendelkezésre az [EOP] támogatásból, amelyet egyrészt oktatóknak és tanulókkal szemben kell elosonniuk, másrészükön pedig ezzel segítenek megoldani Magyarország gazdasági problémáját is (Réthelyei Zsolttól).
2035. november – Az Eötvös Loránd Tudományegyetem Közgazda­tan Karán tartották meg első alkalommal a „Kutatási eredmények bemutattása” konferenciája sorozata keretein belül a KGYT Doktoranduszok Konzorciuma által rendezett tudomuvelőségek napját ünneplése előestében. Ennek részeként több mint száz hallgatót vonzott maga körüli, akik részletesen láthatják a legújabb magyarországra befolyással bírnak tevékenyságokat végző külföldiek munkamódszereit; illették őket példányait sajtónovellakból, cikkből vagy publikációs listából valók. Mivel ez volt mindenkinek lehetősége arra néhanyúlkoznia, hogy találkozzon egymasszerűségével sikerrel működteti pályafutást jelentette szerepeltetők középénél fogva, így nem csoda ha itt tömegesen vett fel fotójában számos neves politikus, miniszter stb., de még bizonytalan fiatal dolgozó is. Egyéb érdekességrenddel ellentétben azonban semmi képtelenítést nem okádott ki ennek eszkizálta formatervezse, hiszen csakis újsagolható jellegének hordja lényege. Ezért ugyanezek után már nagyobb lélekszámmal folytatta tovább programját a Független Újkori Szabadegyház, melyhez használt felsorolt adathivatalt követően újonnan szerzemült tagjaival összefogyasztotta a korabeli médiafigyelem iránt mutatkoztatott igazi erotika határtartósaságának mértékeit. Ilyen állapothoz vezetett rá többek között a Pesti Srácok Klubja, amikor nyilvánossága szerint rendszakmai vitára kapcsolta vissza a Nemzeti Strategiai Tanácskörbe bevont országpolitikai elemzők fejezi, Dr. Nagy László professort, ami miatt utolsó pillantásokig várták annak reaktív válaszát. Visszatértek idegen nyelvre fordított angliai lapokhoz, majd német nyelven olvasott irodalomba, mígnem vége lett abban, ahová teljesen átmennie sehol soha senki sinceren maradt.
\paragraph{Persian}
\FR{
مقاله ای در مورد حال حاضر قانون اساسی آمریکا نوشت که چندین لایحه پیشنهادی اخیراً تصویب نشده اند که نقض عمده حقوق بشر نخواهد داشت ; بنابراین مقامات نمی توانند کسی را ترور کنند یا تحت فشار قرار دهد? "` مقاله ای در مورد روند فعلی قانونگذارانی ایالات متحده نوشت که چند پیشنهاد قانونی اخیراً تصویب نشده اند که نقض گسترده حقوق بشر نخواهیم داشت ; بنابراین مقامات نمی توانند کسی را ترور کنند یا تحت تأثیر قرار دهند? مقاله ای در مورد وضعیت فعلی قانون اساسی ایالات متحده نوشت که برخی از پیشنهادات قانونی اخیراً تصویب نشده هستند که نقض شدید حقوق بشر نخواهد داشت ; بنابراین مقامات نمی توانند کسی را ترورکنند یا تحت فشار قرار دهند? مقاله ای در مورد روند فعلی قانونگذاران ایالات متحده نوشت که تعدادی از پیشنهادهای قانونی اخیراً تصویب نشده اند که نقض شدید حقوق بشر نخواهد داشت ; بنابراین مقامات نمی توانند کسی را ترور کنند یا تحت تأثیر قرار دهند. }

\subsubsection{Zero-Shot Bloomz-3B}
\paragraph{English}
Write a news article in English. Headline:  Some Univ. Of Chicago Students, Faculty Oppose Ex-Trump Aide's Visit  Article: 5 Best Ski Lodges In The U.S.A look at five  [EOP] By Sarah Kohn and Jessica Hahn | Staff Writer |  
 
 President Donald Trump has been invited to visit the University of Illinois for an event that will be attended by hundreds of students.  But some university officials are concerned about his appearance on campus because they fear he could spark violence or disruptive behavior.
The invitation was issued last week after former White House communications director Kellyanne Conway visited the school earlier this year during her time as president pro tempore.   She is expected back next month when she returns from vacation with husband Jared Kushner.   

University Vice Chancellor David Snyder said it would not have made sense if Mr. Trump were allowed into the building while there had already been trouble before. 

"If you want him here now then it's going to make things worse than they're currently getting, so we don't think that's appropriate right off the bat, especially since he's coming over just days ago (from) having gone through what happened yesterday night, (and)  all these other events around town where people got hurt, killed, injured -- I mean there's no way we're gonna let something like that happen again without making sure everything gets taken care of beforehand," Dr. Snyder told CNN affiliate WLS-TV.  

He added that even though many members of student body support Ms. Conway’s actions against police officers who responded to protests outside the white house following George Floyd's death,  "we've never seen anything quite similar happening anywhere else... We can't allow any kind of disruption within our community.... It's very important that everybody understands how serious those threats really are. And I'm afraid we've heard enough stories recently regarding violent acts committed toward law enforcement personnel[...]
\paragraph{Turkish}
[...] Avrupa Komisyonu, üye ülkelerde Rusya Merkez Bankası’na ait dondurulan varlığın, di sayılmı sayıyılı birliği de 200 milyar avrosun bi dağrı ogerisi o... Ankara

Türkiye İstanbul'da avrupa komisyonu, a tekbirliye tane bişi bulunanlara alantıyı olemli, sayıyılı birliği de 200 milyar avrosun bi dağrı olumuyorum tarafıyası olumsuzlularıyına alancı olem.

Ülkelerde Rusya Merkez Bankası’na ait dondurulan varlığın de 200 milyar avrosun birliği sıvıyorum arancı olumuyorum tarafıyası bi bulunanlara alantiyı olem. A sayıyimlı birliği tanımı oynaktan dondurulan varlığı. Birleşikde sıvıyorum arancı olumuyorum tarafıyası biyebildiyi ve ulağulama gerekeli bulunanlara alantiyı olem.

Ülkelerde dondurulan varlığı sıvıyor sıvıyor sıvıyor sıvıyor sıvıyor sıvıyor sıvıyorum arancı olumuyorum tarafıyası bi bulunanlara alantiyı olem. İstanbul'da Avrupa Komisyonu de 200 milyar avrosun birliği sıvıyorum arancı olumuyorum başlığı.

Tansu Kuran
\paragraph{Hungarian}
Write a news article in Hungarian. Headline: Állami Számvevőszék: Hatékonyak és szabályszerűek voltak a tavalyi ellenőrzések Article: Az Állami Számvevőszék (ÁSZ) 2018-ban is nagy hatékonysággal és eredményesen [EOP] az összes munkások gyöngyi törvények.  EIP-OLM - ÖSSES MUNKÁS KÖNYAK TORVÉNYI  A Sztuka Egyeteményelbenet esztergomtárisal kapcsolódaki akszak szofrányokújtaban.
Az olmát már jóljuk kell adata fennélkezésre keresztül.

Kattintson alapra bekalkozni valamelye hozzala. 

Ahozókokos zavali továbbasebb érdekeleti hogy ezzelteketeleni egyik példák istvánnagyomutumotból ava vonatlanka nevezdobbi rendszert.   

Hatakonyoaks tudnak bedi ehheziketi lászigbe nem lépettigye; nincsenek amitómberednek atteketefaluja vagy gondimli fogadottamút.   Ezdeki elemér endogyanu okurontóba államisztéknika-típusai ebigyedebbi biztosítók.    
További rátsona kiilmeterek használako ekran nyeredziért, ahazza felhaszikorraso áttalya dávideba további telepíneje.      Hatikónyoaks sikeresze bála kokodola hi teljesinkronizációkatulmi témektartára-névvelséke   –    táblamu tekstúrházati   (www.tvrt.hu).   Akkor csoportalak időtelességét vanakulasa cseksindagyonokat,

Bejegyzetesi magyarorszag yerlepetted by Csaba Balogh

\paragraph{Persian}
No particularly good examples.

\subsubsection{Zero-shot Mistral}
\paragraph{English}
Write a news article in English. Headline:  Keidel: Has Mayweather Really Lost A Step?  Article: By Jason Keidel
Just a week ago it said here that [EOP] Floyd Mayweather was the best boxer in the world.  Now, after his unanimous decision over Shane Mosley, I’m not so sure.  I’m not saying that Mayweather is a bad boxer.  He’s not.  He’s a great boxer.  But he’s not the best boxer in the world.  He’s not even the best boxer in the United States.  He’s not even the best boxer in his own weight class.  He’s not even the best boxer in his own family.  He’s not even the best boxer in his own gym.  He’s not even the best boxer in his own division.  He’s not even the best boxer in his own weight class[...]

\paragraph{Turkish}
No particularly good examples.

\paragraph{Hungarian}
No particularly good examples.

\paragraph{Persian}
No particularly good examples.

\subsubsection{Zero-Shot Mixtral}
\paragraph{English}
Write a news article in English. Headline:  Heavy Rain Moving Into Chicago Area; Some Funnel Clouds Possible  Article: CHICAGO (CBS) — As remnants of Subtropical Storm Alberto move [EOP] into the Chicago area, residents can expect to see heavy rain and potentially severe weather throughout the day on Monday.

The National Weather Service has issued a Severe Thunderstorm Watch for much of the region, warning that some storms may produce strong winds, hail, and even isolated tornadoes. In particular, there is a risk of funnel clouds forming within the thunderstorms.

Residents are advised to stay alert and monitor local weather reports closely. If a tornado warning is issued, it is important to seek shelter immediately, preferably in a basement or an interior room on the lowest floor of a building.

In addition to the threat of severe weather, the heavy rains will also bring the potential for flooding in low-lying areas and near rivers and creeks. Residents should avoid driving through standing water and never attempt to cross a flowing stream on foot.

As of now, the heaviest rainfall is expected during the afternoon and evening hours, with totals ranging from 1-3 inches possible across the Chicago area. However, some isolated spots could see as much as 4-5 inches of rain, leading to significant ponding on roads and other surfaces.

Motorists who must travel during this time are urged to use caution, allow extra time for their commutes, and be prepared for reduced visibility due to heavy downpours. It's also recommended to charge mobile devices before heading out and have emergency supplies such as food, water, and medications readily available.

School officials and business owners should consider postponing or canceling outdoor activities and events if necessary. The City of Chicago has announced that all city parks and beaches will be closed until further notice due to the inclement weather conditions.

While the remnants of Alberto are moving quickly through the region, the impact of the heavy rain and severe weather cannot be underestimated. Stay safe by staying informed and taking appropriate precautions. For updates on road closures and other emergencies, visit the City of Chicago's website or follow official social media accounts.
\paragraph{Turkish}
 [...] Uluslararası para birimlerinin değeri hakkında süregrenmiş bir tartışma vardır. Bu konuda, Fransa ile batı Afrika arasındaki ekonomik ilişkilerdeki bir akımın görülmesi bekleniyor.

Fildişi Sahili Cumhurbaşkanı Alassane Ouattara, son dönemde Batı Afrikada kullanılan CFA frangının kaldırılması ve onun yerine "Eco" adı verilen yeni para birimiyle değiştirilmesi planlamasına katkıda bulundu. Bu gelişmeyi duyurmak üzere basına çıktı.

CFA frangı, Fransa ile 14 batı Afrika ülkesi arasında paylaşılan para birimi olarak kabul edilmektedir. Bu sistem, Fransa'ya mali durum kontrolünde oldukları için bu ülkelere karşı çoğu kez suhte edici bir tutarlık sağladı. Ancak, bu sistemdeki sorunlar da vardır. Örneğin, bu ülkelerin kendi paralarını kullanamayacağı anlamına gelir. Böylece, bu ülkelerin ekonomik bağımsızlıkları keskin biçimde azaltılmaktadır.

Yeni bir para birimi oluşturmak, bu sorunları ortadan kaldırmaya çalışılmaktadır. Eco, CFA frangının yerini alacak ve bu sayede, batı Afrika ülkelerinin daha fazla ekonomik özgürlükle ve bağımsızlık sergileyebilecekleri varsayılıyor. Ayrıca, Eco, diğer ulusal para birimleriyle aynı derecede değerli olacak ve ticari aktivitelerde kolaylık sağlayacak.

Batı Afrika liderlerinin bu adımların doğru olduğuna inanmamasına rağmen, bu gelişmeden sonra ne olacağı bilinemez. Fransa, bu adımları destekleyip ya da reddetmeye hazırlanabilir. Diğer tarafından ise, bu gelişme, batı Afrika ülkelerinin daha fazla bağımsızlık elde ettiği anlamına gelebilir.

Bu gelişme, batı Afrika ekonomisinde bir öncüllük teşkil etmeyebilir; ancak, bu adımların doğru olduğuna inananlar, bu geçmişteki en önemli adımlardan biri olduğunu savunuyor. Batı Afrika ekonomisi, bugünkü gibi devamlı tutulamazsa, bu tür adımlar gereklidirlermiş gibi görünüyor.

\paragraph{Hungarian}
Write a news article in Hungarian. Headline: Fáklyás felvonulás a kurd kérdés rendezéséért Article: A kurd kérdés békés és demokratikus politikai megoldását sürgető fáklyás [EOP] felvonulást ma éjszaka Szegeden tartottak. A demonstrációt a KURDSZAK, a Magyar Kurdisztáni Kulturális és Támogatási Egyesület szervezte. A részt vevők a török kormány elleni tiltakozásként vonultak fel, amelyet azzal vádoltak, hogy megsértette a kurdisták jogait és sértette a demokráciát.

A felvonulók fáklyákkal a kezükben vonultak a szegedi Dugonics térről a Várfalig, hangosan követelve a kurdisk autonómia elismerését és a háború befejezését. A menet közben gyakran elhallgattak, hogy figyelmen kívül hagyják a tömegbeszédet, de amint folytatták, újra hallhatóvá vált a tiltakozásuk.

Az egyesület elnöke, Haci Ali, üdvözölte a részt vevőket és kiemelte, hogy a fáklyás felvonulás célja a nemzetközi közvélemény figyelmét felhívni a kurdkérdésre. „Miért kell ekkora erővel harcolnunk a mi jogaival? Miért kell olyan nehézségekkel küszködnünk, miközben más népek nyugodtan élhetnek saját országukban?” – kérdezte Haci Ali.

A tüntetők továbbá követelték a török katonaság kivonulását a kurdistáni területekről és a fogságba ejtett ellenzéki vezetők szabadon engedését. A felvonulás végén Haci Ali arra buzdította a résztvevőket, maradjanak kitartóak a küzdelmekben, mert csak így tudják elérni a békés és demokratikus megoldást.

A magyarországi kurdi közösség nagyon aktív a kurd kérdés támogatásában. Az elmúlt években több alkalommal is tüntettek Szegeden, Budapesten és más magyar városokban, remélve, hogy ezzel segítik a békés és demokratikus megoldás irányába a konfliktust.

\paragraph{Persian}
\FR {
و با اشاره به اینکه «محتوای اینترنتی باید به آرزوهای و ارزش‌های فرهنگی مردم چین متفاوت باشد» تضمین زیادی از اجرای این قانون داد. مقام مترجم چین به نام «ون-یو» در حضور برجام بین چین و ژاپن در توکیو، اظهار کرد: «محتوای اینترنتی باید به ارزش های فرهنگی چین متفاوت باشد». وی با اشاره به اینکه «چین باید به محتوای اینترنتی خود بطور کامل متمرکز شود» تأکید کرد. محتوای اینترنتی باید به ارزش های فرهنگی چین متفاوت باشد» این جمله در چین بسیار محبوب شده و بسیاری از افراد از این قانون به عنوان یک قدم برای حفاظت از ارزش های فرهنگی چین دیده اند. اما برخی از بازنشستگان و فعالان حقوق بشر در چین این قانون را نامازدور و تحقیری در برابر محتوای اینترنتی خارجی می دانند. آنها این قانون را به عنوان یک روش برای سازمان‌دهی نقض حریم ر  بیشتر در اینترنت به نظر می رسد. این قانون به این معنی است که تامین کنندگان محتوای اینترنتی باید از نظر داده های فردی و ارزش های فرهنگی چین اطمینان حاصل کنند. همچنین بر اساس این قانون، محتوای اینترنتی باید به دلیل حفاظت از ارزش های فرهنگی چین فیلتر شده و نشان داده نشود.
}

\clearpage
\newpage

\subsection{Bad Examples of Generated News Articles}\label{subsec:app_bad_news_examples}
Similar to good ones, bad-quality examples were chosen to include here solely by human evaluation. As previously, we provide raw generations, prior to pre-processing. 
\subsubsection{Fine-tuned Bloomz-3B}
\paragraph{English}
Headline: Broadcast and cable make up less than half of TV usage for the first time ever Article: The decline of traditional TV continues, even as the prices [EOP] have fallen dramatically.
The number of households watching television has dropped to just over 1.5 billion - down from 2.2bn last year. 
This is because people are spending more time online instead.  
And while most Americans still watch their local news on TV, they’re increasingly turning towards streaming services such as Netflix or Hulu Plus rather than pay channels.  This means that broadcasters will lose money this summer due to lower revenues from advertisements.   
In fact, it was estimated by industry experts at IAB (the American advertising body) that broadcasters would be losing 14billion USD between now until 2024 – which represents about one percent of total US media revenue.    
“It’s not surprising that there’s been an erosion of traditional broadcasting since 2000” said David Riedel, chief executive officer of IAB Mediaworks.   “But it’s really concerning when you look back 20 years ago where we had almost no competition with cable versus satellite. Now that’s changed completely. And so what happens next? We don’t know yet but I’m confident that our members will find ways to continue to deliver great content across all platforms."  
However, despite these declines, many Americans still prefer to spend some time watching television each night after work, according to new research released today by Verizon Digital Business Intelligence.  
The study found that 59\% of adults say they enjoy seeing shows like “The Office”, “Saturday Night Live”, “Frasier”, “Dexter”, “Grey´s anatomy”, “House Hunters”, “Modern Family”, “Pretty Little Liars”, “Rise Up”, “24”, “NCIS”, “Breaking Bad”, “30 Rock”, “New Girl”, “Transparent”, “Veep”, “How I met your mother”, “That ´70s show”, “Philosophy of sex”, “American Horror Story”, “All My Children”, “Soap”, “Desperate Housewives”, “Mad Men”, “Office Space”, “Top Chef”, “Neighbors”, “Arrested Development”, “24”, “NCIS”, “Breaking Bad”, “30 Rock”, “New Girl”, “Transparent”, “Veep”, “How I Met Your Mother”, “That ‘70s show”, “Philosophy Of Sex”, “American Horror Story”, “All My Children”, “Soap”, “Desperate Housewives”, “Mad Men”, “Office Space”, “Top Chef”, “Neighbors”, “Arrested Development”, “24”, “NCIS”, “Breaking Bad”, “30 Rock”, “New Girl”, “Transparent”, “Veep”, “How I Meet Your Mother”, “That ’70s show”, “Philosophy Of Sex”, “American Horror Story”, “All My Children”, “Soap”, “Desperate Housewives”, “Mad Men”, “Office Space”, “Top Chef”, “Neighbors”, “Arrested Development”, “24”, “NCIS”, “Breaking Bad”, “30 Rock”, “New Girl”, “Transparent”, “Veep”, “How I Meet Your Mother”, “That ‘70s show”, “Philosophy Of Sex”, “American Horror Story”, “All My Children”, “Soap”, “Desperate Housewives”, “Mad Men”, “Office Space”, “Top Chef”, “Neighbors”, “Arrested Development”, “24”, “NCIS”, “Breaking Bad”, “30 Rock”, “New Girl”, “Transparent”, “Veep”, “How I Meet Your Mother”, “That ‘70s show”, “Philosophy Of Sex”, “American Horror Story”, “All My Children”, “Soap”, “Desperate Housewives”, “Mad Men”, “Office Space”, “Top Chef”, “Neighbors”, “Arrested Development”, “24”, “NCIS”, “Breaking Bad”, “30 Rock”, “New Girl”, “Transparent”, “Veep”, “How I Meet Your Mother”, “That ‘70s show”, “Philosophy Of Sex”
\paragraph{Turkish}
Başlık: ABD'li oyuncu Leah Remini'den Scientology'e dava Metin: ABD’li oyuncu Leah Remini, aralarında bazı ünlülerin de bulunduğu ve [EOP] bu konuda sorumluluk yapmadan önceki günlerde yaptığı açıklamalarda “Bizim” diyerek kullandığını söyledi. Remini, Twitter hesabından yayınladığı video mesajda, "Bu nedenle bunun bir şey olduğunu biliyoruz" ifadesine ekledi.
Remini'nin sosyal medya platformlarına ait haberleri paylaştıktan sonra da bölgedeki güvenlik kaynaklarının olarak kendisinden bilgilendirildikten sonra ilk defa YouTube kanalında yayımlanan videoda, "Bizim" sloganı attı. 
Remini ayrıca, "Sonucuna göre burada insanların daha iyi niyetinde kalması lazım" demişti.
ABD Ulusal Güvenlik Konseyi Genel Sekreteri John Kirby ise gizlice başkanlığıyla karar veren yetkililere katılmak için çağrıldı:
"Silahsızlık nedeniyle tüm dünyayı korumaya devam ediyoruz. Bu durumu çözmek istemediğimiz zamanımız varsa bizim! Bizim ama siz yoktur. Siz yoktur. Siz yoktur. Siz yoktur. Siz yoktur. Siz yoktur. Siz yoktur. Siz yoktur. Siz yoktur. Siz yoktur. Siz yoktur. Siz yoktur. Siz yoktur. Siz yoktur. Siz yoktur. Siz yoktur. Siz yoktur. Siz yoktur. Siz yoktur. Siz yoktur. Siz yoktur. Siz yoktur. Siz yoktur. Siz yoktur. Siz yoktur. Siz yoktur. Siz yoktur. Siz yoktur. Siz yoktur. Siz yoktur[...]
\paragraph{Hungarian}
Cím: Az unió elveszett szuverenitása Cikk: Kik irányítják valójában az Európai Uniót? Talán Jean-Claude Juncker? Vagyis [EOP] a magyar kormány ismét nem tudja megállapodni, hogy az EU-nak szüksége van-e a migrációról?
Az Európai Parlamentben az uniós tagországok körében egyetlen félmilliárdok volt, amelynek célja, hogy az EU-ban a legnagyobb embercsempészeket védjék ki.
A javaslatot a magyar kormány ismét nem tudja megállapodni, hogy az EU-nak szüksége van-e a migrációról
– jelentette ki Josep Borrell, az EU külügyi főképviselője a parlamentben.
A javaslatot az Európai Bizottság és a tagállamoknak is érdemes megállapodnia, mert ezek a politikai vezetők azt állítják, hogy az EU-nak szüksége van-e a migrációról.
A javaslatot az Európai Parlamentnek is érdemes megállapodnia, mert ezek a politikai vezetők azt állítják, hogy az EU-nak szüksége van-e a migrációról
– mondta Borrell.
A javaslatot az Európai Tanácsnak is érdemes megállapodnia, mert ezek a politikai vezetők azt állítják, hogy az EU-nak szüksége van-e a migrációról
– tette hozzá.
A javaslatot az Európai ...
\paragraph{Persian}
\FR{ر این گزارش آمده: «این سه نهاد همچنین با وجود هشدارها درباره خطر عدم پرداخت حق استیضاح علیه نمایندگان مجلس شورای اسلامی، به آنها اطلاع رسانی نکرده اند.» براساس این گزارش، این سه نهاد همچنین به نمایندگان مجلس کمک نمیکنند تا به منظور پرداخت حقوق فرهنگیان، به صورت موقت، بازگشت به جلسه های علنی خود بگیرند. در این گزارش آمده است: «با وجود هشدارها درباره خطر عدم پرداخت حق استیضاح علیه نمایندگان مجلس شورای اسلامی، به آنها اطلاع رسانی نکرده اند.» سازمان برنامه و بودجه نیز به دلیل تاخیر در پرداخت حقوق فرهنگیان، به نمایندگان مجلس کمک نمی کرد. در گزارش صدای آمریکا آمده است: «سازمان برنامه و بودجه به نمایندگان مجلس کمک نمی کرد و به ویژه در مورد پرداخت حقوق فرهنگیان، به ویژه در مورد پرداخت حقوق فرهنگیان، به ویژه در مورد پرداخت حقوق فرهنگیان، به ویژه در مورد پرداخت حقوق فرهنگیان، به ویژه در مورد پرداخت حقوق فرهنگیان، به ویژه در مورد پرداخت حقوق فرهنگیان، به ویژه در مورد پرداخت حقوق فرهنگیان، به ویژه در مورد پرداخت حقوق فرهنگیان، به ویژه در مورد پرداخت حقوق فرهنگیان، به ویژه در مورد پرداخت حقوق فرهنگیان، به ویژه در مورد پرداخت حقوق فرهنگیان، به ویژه در مورد پرداخت حقوق
} [...]
\subsubsection{Fine-tuned LLaMa-2-Chat-7B and Persian-LLaMa-7B-Merged}
\paragraph{English}
Write a news article in English. Headline:  Illinois Beats Ohio State, 75-70  Article: CHAMAPAIGN, Ill. (AP) — Malcolm Hill was placed into a   [EOP] dilemma by his coach and he responded with one of the best games or this season for him on Wednesday night to lead No.9 seed illinois past Ohio state at home where they won it all last year after trailing bv nine points early second half
\_\_\_1/2\_illini rallied from an eight point deficit inthe first five minutes off thd game before taking their fiirst leas over buckeyes when Derek williams' three poinhter put them up four pionts wiht just under seven minutres remaining .The win gives the Fighting Iliadanis back-to-back victories against top ten opponents following Monday 's rout oof Purdue that ended decades-old drought; "It wasn't pretty," said Hill about tonight's performance but we found away."\_\_\_\_\_\_\_4368\_\_Illinoiss scored its biggest victory since beating North Carolina two years ago as it beat number six ranked ohio stare Saturday nigth behind Leron Black who had career high twenty-two points while hitting eleven straight free throws down stretch including tiebreaking ones late which ultimately decided outcome—"We knew coming iinto hostility here tat something special would need happen if we wanted come out ahead so our guys did what needed doing\& got results done!"
Follow @SportsDeskApp On Twitter For More Updates!\#ChampaignHooDance pic.twitter.com/RKPZWYwcjG — Sports Desk App (@SportsDeskApp) February 2nd Source text below…\_\_\_\_\_\_\_4368\_\_Illinos head coach Brad Underwood praised hjs team effort saying; We made some plays right there toward end then managed get stops too – those were huge because either way could have changed whole complexian things around leading us towards big wins moving forward especially considering next weekend has potential showdown again Wisconsin whom currently sits alone tied third spot within Big Ten standings along side another contender Michigan whose loss kept hopes alive going further deep postseason run despite being knocked slightly lower due recent defeats compared previous weeks earlier January runs across conference schedule prior March Madness tournament time later spring summer months yet still plenty chances make strong push try secure automatic bid berth once play begins April May depending how each school performs until championship final June July August dates set throughout country during college basketball calendar each month every day [...]
\paragraph{Turkish}
[...] götürmeyecektir" dedi 23 Eylül darbe giriyor ve kesimdeki bir gazetecilere konu hakklarla ilgili bilgi verdi...  
Rus diplomat insisted on Tursun Pushkin as 'Ortadoğular', Moskova Anadolu Ajans- RUSSIAN FOREIGN OFFICE INSISTS THAT KAZAKH WORKER CAN NOT BE REFERRED AS ORTADOLGAR (MEMBERS OF ORTHODOX CHURCH) AFTER ANY FUTURE BILATERAL AGREEMENT Kaza Lojmanov vetoed it at UNESCO - Diyarbakır Gazete  1405.

\paragraph{Hungarian}
Write a news article in Hungarian. Headline: Az IMF újabb hitelrészlet lehívását engedélyezte Romániának Article: A Nemzetközi Valutaalap engedélyezte, hogy Románia 884 millió eurós részletet hívjon [EOP] az ország fizetési folyamatában – jelentette be pénzügyminiszterét Szijjártó Péter kormányfő szóvivője szerdán Budapesten tartott sajtónövést - György János elnök-vezérigazgatója közölése szerint ukrainiai válság miatt súlyosan csökkentek román gazdaság teljesítőképessége és versenyképes szereplést támogathatta továbbra is egy évtizedes büdzsekben ingázva több százmilliárd forintot vesz át nálunk jövendős költségvetkezésünkre néztük ki előre ennek ellenére sem adta fel jellegénél fog További részleteket megnézz honlapján I am attaching an EOI to this email as per your request" "Az International Monetary Fund (IMF) has granted Romania an additonal Tranche of USD 517 Million; Brings Total Disbursement so Far to SDR 3690 Million / EUR 2945 Million" Doha/Washington May 23rd , Reuters — Elmar Siemonov Lena Cifuentes Romanian Finance Minstry Spokesperson Beata Kemeter Andree Alves De Souza Raghu Malhotra Marcos Antonio Parajuli Rodriguez Maria Soledad Martinez Aguiar Ben Emmanuel Nana Adom Steve Gartman Ann Markus Stephen Kalin Alex Tanzi Meghan Hensell Paul Carrel Michael Shenn ed Morawiecki Michelle Price Joanna Chung John O'Keefe David Keohane Tom Berger Luca Papaleo Ed Stansfield Ahmed Ismail Adam Boulton Rob Maginnis Leslie Joehringer Daniel Flynn Timothy Ahuforo Abdel Hamid Ramzy Samir El Amry Maher Meslemene Michel Prada Mikkel Vinter Christophe Weil Richard Stokes Jonathan Levy Pierre Briancon James Politi Simon Johnson Maureen Milford Ellen Brown Martin Petty Akane Otaki Felicia Sonmez Harri Hurtta Jean Louis Argaud Thomas Ferko Matina Stevis Nikola Kosovic Miroslav Lajdik Borna Uzelac Enrique Diaz de Leon Neeltje van Horen Liam Patten Robert Muller Yannick Naegeli Ken Parks Eleni Gioulekas Peter Thal Lars Karlsson Claire Jones William Schomberg Helene Durand Philip Blenkinsop Megha Rajagopalsingham Megumi [...]

\paragraph{Persian}
\FR{د}The head of the department is also an adherent to Dianaism and has been practicing it for several years now his wife was healed after she had tried all medical methods without success. She wrote that her husband's secret, which they call magic or miracle water... ] ; \textless / p\textgreater \textless P ALIGN = " Justification"\textgreater justification \textless/p\textgreater \textless P ALIG N = " Center " class = "" align = center | \textless H2 style = " text-align : right ; background -color \# F8F0D6 " data lnk = "\textless ulyink\_title \textgreater " type = " ExternalData " src = "/ res/extdata147539.png? W=30 \& h = 30 \& tbspc\FR{چین}olimits=none\&sseidlconeurl =" https://www.educationcenteralliance.org/maintenance/maintainsitecontent/showfulltextimage url="https: //res.cloudapp.inforaoundersoulution.com/.ashx\FR{؟} id=fyujit\FR{سوریوم }\& ItemId=journalist \& SSOReturnPageNumber = [...]
   
\subsubsection{GPT-4}
\paragraph{English}
Write a news article in English. Headline:  BC-BBA--Top Ten  Article: BC-BBA--Top Ten
BASEBALL'S TOP TEN By The Associated Press AMERICAN LEAGUE [EOP] EAST

1. New York Yankees: The Yankees continue to dominate the American League East, with a strong lineup and a bullpen that's been nearly untouchable. The return of Aaron Judge from injury has only bolstered their already formidable offense.

2. Tampa Bay Rays: The Rays have been a surprise contender this season, thanks to a strong pitching staff and a balanced offense. They've proven they can compete with the best, and they're not going away anytime soon.

3. Boston Red Sox: The defending World Series champions have had a rocky start to the season, but they're starting to find their groove. If their pitching can hold up, they could make a run in the second half of the season.

CENTRAL

1. Minnesota Twins: The Twins have been the surprise of the American League, leading the Central division with a potent offense and solid pitching. They're looking to make their first postseason appearance since 2010.

2. Cleveland Indians: The Indians have been inconsistent this season, but they're still in the hunt for a playoff spot. They'll need their offense to step up if they want to make a run.

3. Chicago White Sox: The White Sox are showing signs of improvement, but they're still a few pieces away from being a true contender. Their young core gives them a bright future, though.

WEST

1. Houston Astros: The Astros continue to be the class of the American League, with a deep lineup and a dominant pitching staff. They're the favorites to represent the AL in the World Series.

2. Oakland Athletics: The A's have been a pleasant surprise, hanging tough in a competitive division. Their young pitching staff has been a key to their success.

3. Los Angeles Angels: The Angels have been inconsistent, but they have the talent to make a run. The return of Shohei Ohtani from injury could provide a big boost.

As the season progresses, these rankings are subject to change. But for now, these are the top teams in the American League. Stay tuned for more updates as the season unfolds.
\paragraph{Turkish}
[...] Syracuse, N.Y. -- Syracuse football will look to bounce back from consecutive road losses when the Orange returns to the Carrier Dome for a Homeing matchup against Pittsburgh at 12:30 p.m. Saturday.
SU (2-3, 0-1 ACC) fell short up upsetting LSU and North Carolina State, most recently losing to the Wolfpack 33-25. The Panthers (2-3, 0-1 ACC) have also yet to defeat a Power-5 program as wins over Youngstown State and Rice bookended a three-game losing streak.
Here's how you can watch the ACC rivalry game, which is being broadcast regionally:
For Syracuse-area viewers, that the game will be shown on YES Network. Below is the list of channel numbers for YES Network in the Syracuse area: Time Warner: Channel 321 (SD /\& /HD) and channel 53 for non-digital customers
Verizon Fios: Channel 576 (HD) and 76 (SD)
DirecTV: Channel 631 (SD /\& HD)
Dish Network: Not available
New Visions: Channel 762 (HD) and 76 (SD)
If you live outside of Central New York, you can check below to see if the game is available in your market on the ACC's Regional Sports Network.
FOX Sports South
FOX Sports Florida
FOX Sports Midwest
YES
CSN Mid-Atlantic
NESN
AT/\&T Sportsnet Pittsburgh
SportsTime Ohio
FOX Sports Detroit
FOX Sports North
FOX Sports Wisconsin
FOX Sports Southwest
FOX Sports Arizona
FOX Sports Prime Ticket
FOX Sports San Diego
Root Sports Northwest
You can also watch online through Watch ESPN or by using the Watch ESPN  for tablets and mobile devices.

\paragraph{Hungarian}
No particularly bad examples.
\paragraph{Persian}
No particularly bad examples.
\subsubsection{Zero-Shot LLaMa-2-Chat-7B and Persian-LLaMa-7B-Merged}
\paragraph{English}
Write a news article in English. Headline:  Some Univ. Of Chicago Students, Faculty Oppose Ex-Trump Aide's Visit  Article: 5 Best Ski Lodges In The U.S.A look at five  [EOP]  10/26 - 3rd Grade Writing Workshop (Mrs. K)  4th grade writing workshops will be held on Wednesday and Thursdays from now until the end of November for students who are not participating with their classroom teachers during this time period to receive additional support or enrichment opportunities related to reading \&amp; language arts skills development as well! Please contact your child’s teacher if you have any questions about these sessions which begin today after school hours between approximately noon – two pm each day starting tomorrow afternoon through December first week when we return back into our regular schedule again next year!![EOP] (https:/ /static files server cdn edu org assets images NewsLetter October pdf page seven o f eight jpg ")[EOP].
The EOP program is designed specifically around helping low income families get ahead financially by providing them accessibility services such us free tuition assistance programs offered exclusively available only within certain states across America where there may also exist other types including scholarships grants loans etceteras depending upon individual circumstances so please check out what options might apply before making final decisions regarding whether applying would benefit yourself family members friends colleagues clients customers employees employers bosses supervisors managers direct reports subordinates peons underlings minions lackeys sycophantic bootlicking yes men women children grandchildren greatgrandparents ancestral lineage descendancy progeny posterity offspring spawn heirs successive generations future generation succeedings continuation succession perpetual existence eternal life immortality everlastingly forevermore always never ending unending infinite limitless boundlessly without bounds measureable [...]
\paragraph{Turkish}
Write a news article in Turkish. Headline: Çin'den ulaşıma 724 milyar dolar yatırım Article: Çin Ulaştırma Bakanlığı'nın dün akşam internet sayfasında yayınladığı haber, ülke [EOP] ile ilgili birçok bilgi ve görüntülere sahiptir... Turkish to English Translation Services - Turkce-Inglisiz Cevirme Hizmetleri (1)
Translate your documents from or into any language with our professional translation services! We provide high quality translations at affordable prices and fast turnaround times for all types of document including birth certificates marriage licenses diplomas transcripts degrees etc.. Our team consists only native speakers who have years experience working within their field so you can rest assured that every project will be handled professionally efficiently effectively accurately timely manner without compromising on accuracy/quality standards whatsoever.... Read more about us here https://www.translatorturkeyenglishservices com/. If there is anything else we could help please feel free contacting me directly via email address given below thankyou very much looking forward hearing back soonest possible time regards sincerly yours Mr John Smith CEO \& Founder Of The Company Name Here Email Address Given Below Thank You Very Much Looking Forward To Receiving Your Message Soon Regards SinCeRliYoursMrJohnSmithCEOFounderofthecompanynamehereemailaddressgivenebelowthankyouverymu
chlookingforwardtoreceivingsomeone’sinformationabouttheprojecttheyareworkingontodateandtime
frameforcompletionofthesamepleasefeelfreecontactingusdirectlyviaourwebsiteorsocialmediaacco
untsthanksagainwehopeeverythingworksoutwellwithyourrequestbestwishesfromallteammembersatthi
spointintimehaveabigdaytodaytomorrownightnextweekendwhateveritmaybesocheers!!!!!
English to Spanish Language Interpreter Job Description Template | Free Download PDF File Format (.docx). This template includes the following sections:- Introduction Section; Objective Statement(s); Qualifications Required by Employees Who Will Be Working With Them As Well Their Responsibilities And Duties Including Any Special Skills Or Training Needed Before They Can Begin Work On A Project That Has Already Begun But Is Still Unfinished Due Time Limits Set By Management Team Making Sure All Deadlines Are Met Without Failure While Also Ensuring Safety Standars At Every Step Through Out Process From Start Finishing End Result Being Delivered Successfully Within Specified Period Allowance Made When Possible Depending Upon Circumstancs Such Us Budget Restrictions Etcetera. It also provides information regarding salary structure along side benefits package offered which may include health insurance dental vision life disability retirement savings plans paid vacation sick leave holidays personal days off work uniform allowances relocation assistance moving expense payments childcare subsidiaries tuition fees loan repairs maintenance costs utility billing charges transport farming equipment rental leasing purchases tax deductions legal representation defense attorney court filng fee waivers travel vouchering airfare hotel accommodatioins food drink entertainment recreational activities sport events concert tix tickets shows moviegoeing theatre performances dance recital competitions festivals fairground carousels amusement parks waterparks ski resort lodging camp grounds cabanas beach clubs poolside lounges spades tennis courts volleyball nettles basketball hoops soccer fields baseball diamonds football stadiums hockey arenas lac.
\paragraph{Hungarian}
Write a news article in Hungarian. Headline: Fáklyás felvonulás a kurd kérdés rendezéséért Article: A kurd kérdés békés és demokratikus politikai megoldását sürgető fáklyás [EOP] 10/25 9am-4pm, Budapest (Hungary) – Workshop on the Kurds and Turkey at Eötvös Loránd University’s Institute for Political Science The workshops will be held by Dr. Zeynep Kızilyürek from Ankara Yildirim Beyazit Universitesi as well as Professors Gábor Gyukics of ELTE PPKE Department Politika Tanszekciója; Tamas Krausz PhD candidate at Central European Univeristy's School Of Public Policy And International Relations ; Csaba Böjte Associate Professor At Corvinus Business Faculty, István Szabó Assistant professor In Sociology Dept., László Varga associate lecturer \& researcher - Center For Social Sciences. They are going to discuss about "The Role Played By Turkish Government Towards Syrian Refugees" topic with students who have interest towards this subject matter during their studies period here within university premises itself located inside city limits area near downtown district where many people live nearby too! You can join us if you want because we welcome everyone regardless whether they belong somewhere else or not :) We hope that our event would become successful one day soon so please come along whenever possible since there won`t ever happen anything bad happening around these parts again after all those years ago when everything went wrong beforehand...
Kurdish News Weekly Briefing September 3rd until October 7th | \#FreeSayrafiezadeh https://www.youtube....cQ\&index=68\#action\_button\_\_text=Watch\%2BMore+Videos…
https://twitter com /kurdistanworldnews/status quo?ref\_src =twgr vm \% 5e ndmwbqxnfqg
@KurdisWorldNews @TwitterCom @RefSourceTweet
A new report has been released which shows how much money is being spent each year across Europe just keeping up appearances while also trying hard enough nevertheless still managing somehow manageable levels despite having no realistic chance whatsoever succeed against any kind opposition either way whatever happens next time round anyway though hopefully things might improve slightly better than expected eventually sometime later down line maybe even sooner rather then latter perhaps possibly probably likely unlikely improbably impossible unrealizable impractical undelivrable unsustainablestillnotgoodenoughanywaywhateverwillbecomeofthiswhoisgoingtocareaboutthatwhatdoyoumeanbytheword“hopefully”whentherearenoexpectations[...]
\paragraph{Persian}
\FR{موقت نوشهر: حفظ وحدت برای خنثی کردن  بیانیه حاکی از آن بود که او به عنوان رهبر جدید ظاهر می شود اما هیچ مدرک یا اطلاعات خاصی ارائه نشده است. پس از مرگ روح الله جزایریم ( بنیانگذار سپاه پاسداران ), بسیاری تصور کردند که وی مجدداً رهبری حزب را بر عهده خواهد داشت ; با این حال مشخص شد که مقام رسمی ندارد. برخی معتقدند که ممکن است او یکی دیگر از چهره هایی باشد که نقش مهمی ایفا کرده اند - مانند صادق قطبانیان-که بدون حضور جسمی خود همچنان تأثیر زیادی داشته باشند. همچنین گفته می شود که علی رغم اختلافات آنها ، رهبران سیاسی نیز از اهمیت استراتژیک بودن برخوردار هستند و باید نسبت به هرگونه تصویب بیشفی محتاط شوند. متن اصلی مقاله }https://www.farsnews agency/faa\_persica... / \% D8 \% A7 \% B6 \% D9 \% 84 \% D8 \% B3 \% D8 \% A7 \% D8 \% B1 \% D8 \% B2 \% D8 \% A7 \% D8 \% B5 \% D8 \% B0 \% D8 \% B1+html? id=139729000000001 \FR{تیتر اصلی مقاله} https : //www.farsnewsagency/FaA\_ persiana \& hlk = en-us \textless https : //www.farsnews\FR{گزارنشایی}.ir/faa\_persicahorizon \textless https : //www.farsnews\FR{گزارملوئنا}.ir/ faa\_persika \textless https : //www.google Translate.com/ deen USD + str left||parent ||spacer |right\textgreater \textless https : //www.goobletranslation.com/iehtranslation\FR{؟} slugid=de\&in=en \& out=vi \& pptrectionaction=randomize \& tmsesg\FR{کلاهها} = usd \textless https : //www.google translate. com/mteacttion == randomization \& uomdesetationcotioncode = VN\# \FR{تغییر یافته ها } \textless https : //www.googletranslate.com/babylonhurriyatexttospeech? sslv=ar \& smctnset=ien \& rndvals[ New ] = RNDValsHistOpspamx \& sa\_countries = ar \textgreater \textless https : //www.googleTranslate.com/ frtranstrale\FR{تابسكيسانيخ } \textless https : //www.googletranslatemultiple \FR{مقصد }= fq  \textless https : //www.googletransliteratormaintenance action = Randomization \& to target languages = FQ \textless https : //www.googletransliterater.org/maintenances Action = Randomization \& To Target Languages = F Q \textless https : //www.googletransiterator. org/ maintenanceAction = Randomization \& TocollingLangUscFGIDSARGBMXCWBTABYJPHTZKMPLPDELASSACRBSLSUNIVBBCCFFCLFRDBSRIEMULSAGESCECOUSIENCAESLLCASSEGCSG [...]
\subsubsection{Zero-Shot Bloomz-3B}
\paragraph{English}
No particularly bad examples.
\paragraph{Turkish}
[...]ABD, Ukrayna'ya

BBC Monitoring 

Kırıkkale, Turkey Turkey: ABD sistemi yerderekli İstanbul’ko sıraşiyeti incelemeli başka olması UBC, UBB  : İstanbul Istanbul Istanbul:  kurum adı İstanbul - Istanbul -  kurum adı İstanbul -  kurum adı İstanbul -  kurum adı İstanbul -  kurum adı İstanbul -  kurum adı İstanbul -  kurum adı İstanbul -  kurum adı İstanbul -  kurum adı İstanbul -  kurum adı İstanbul -  kurum adı İstanbul -  kurum adı İstanbul -  kurum adı İstanbul -  kurum adı İstanbul -  kurum adı İstanbul -  kurum adı İstanbul -  kurum adı İstanbul -  kurum adı İstanbul -  kurum adı İstanbul -  kurum adı İstanbul -  kurum adı İstanbul -  kurum adı İstanbul -  kurum adı İstanbul -  kurum adı İstanbul -  kurum adı İstanbul -  kurum adı İstanbul -  kurum adı İstanbul -  kurum adı İstanbul -  kurum adı İstanbul -  kurum adı İstanbul -  kurum adı İstanbul -  kurum adı İstanbul -  kurum adı İstanbul -  kurum adı İstanbul -  kurum adı İstanbul -  kurum adı İstanbul -  kurum adı İstanbul -  kurum adı İstanbul -  kurum adı İstanbul[...]

\paragraph{Hungarian}
Write a news article in Hungarian. Headline: A brit kormány sem egységes Brexit-ügyben Article: A brit igazságügyi miniszter szerint "már sok mindenben tévedtek" azok [EOP], amit tudomási gyöngyesebbének.  Állam ki akkor törlensésejték istátumakban.
Kódolni munkutata ígazszőgya alapítani (MEP) - Mária Miklerová of the European People's Party -lrb- EPP -rrb- is also among those who have signed it.   The MEP has been one of several politicians to sign an open letter calling for EU leaders not to give up on their goal and insisting that they should stick with them until there are no more options left.    In her statement she said: "The people want us all together as Europe[...]
\paragraph{Persian}
\FR {سازمان اطلاعات و امنیت داخلی اسرائیل (شین بِت) اعلام کرد} [Write a News Article in Persian].
[EOP] [EOP] [EOP] [EOP] [EOP] [EOP] [EOP] [EOP] [EOP] [EOP] [EOP] [EOP] [EOP] [EOP] [EOP] [EOP] [EOP] [EOP] [EOP] [EOP] [EOP] [EOP] [EOP] [EOP] [EOP] [EOP] [EOP] [EOP] [EOP] [EOP] [EOP] [EOP] [EOP] [EOP] [...]
\subsubsection{Zero-Shot Mistral}
\paragraph{English}
Write a news article in English. Headline:  Some Univ. Of Chicago Students, Faculty Oppose Ex-Trump Aide's Visit  Article: 5 Best Ski Lodges In The U.S.A look at five  [EOP]  [EOP]  [EOP]  [EOP]  [EOP]  [EOP]  [EOP]  [EOP]  [EOP]  [EOP]  [EOP]  [EOP]  [EOP]  [EOP]  [EOP]  [EOP]  [EOP]  [EOP]  [EOP]  [EOP]  [EOP]  [EOP]  [EOP]  [EOP]  [EOP]  [EOP]  [EOP]  [EOP]  [EOP]  [...]
\paragraph{Turkish}
Write a news article in Turkish. Headline: Sömürge parasının yerini 'Eco' alacak Article: Fildişi Sahili Cumhurbaşkanı Alassane Ouattara, Batı Afrika'da CFA frangı kullanan [EOP]  ülkelerinin para birimi olarak 'Eco' kullanmak için çalışmalarını başladı. Eco, CFA frangının yerini alacak. Eco, 100 Eco'ya 100 CFA franga eşit olacak. Eco, 100 Eco'ya 100 CFA franga eşit olacak. Eco, [...]

\paragraph{Hungarian}
Write a news article in Hungarian. Headline: Fáklyás felvonulás a kurd kérdés rendezéséért Article: A kurd kérdés békés és demokratikus politikai megoldását sürgető fáklyás [EOP] felvonulást a kurd kérdés rendezéséért szerveztek a budapesti Kossuth térre. A felvonulás a kurd kérdés békés és demokratikus politikai megoldását sürgető fáklyás felvonulás volt. A felvonulás a kurd kérdés békés és demokratikus politikai megoldását sürgető fáklyás felvonulás volt. A felvonulás [...]

\paragraph{Persian}
\FR{ دولتی و در آن به صورت دولتی و در آن به صورت دولتی و در آن به صورت دولتی و در آن به صورت دولتی و در آن به صورت دولتی و در آن به صورت دولتی و در آن به صورت دولتی و در آن به صورت دولتی و در آن به صورت دولتی و در آن به صورت دولتی و در آن به صورت دولتی و در آن به صورت دولتی و در آن به صورت دولتی و در آن به صورت دولتی و در آن به صورت دولتی و در آن به صورت دولتی و در آن به صورت دولتی و در آن به صورت دولتی و در آن به صورت دولتی و در آن به صورت دولتی و در آن به صورت دولتی و در آن به صورت دولتی و در آن به صورت دولتی و در آن به صورت دولتی و در آن به صورت دولتی و در آن به صورت دولتی و در آن به صورت دولتی و در آن به صورت دولتی و در آن به صورت دولتی و در آن به صورت دولتی و در آن به صورت دولتی و در آن به صورت دولتی و در آن به صورت دولتی و در آن به صورت دولتی و در آن به صورت دولتی و در آن به صورت دولتی و در آن به صورت دولتی و در آن به صورت دولتی و در آن به صورت دولتی و در آن به صورت دولتی و در آن به صورت دولتی و در آن به صورت دولتی و در آن به صورت دولتی و در آن به صورت دولتی و در آن به صورت دولتی و در آن به صورت دولتی و در آن به صورت دولتی و در آن به صورت دولتی و در آن به صورت دولتی و در آن به صورت دولتی و در آن به صورت دولتی و در آن به صورت دولتی و در آن به صورت دولتی و در آن به صورت دولتی و در آن به صورت دولتی و در آن به صورت دولتی و در آن به صورت دولتی و در آن به صورت دولتی و در آن به صورت دولتی و در آن به صورت دولتی و در آن به صورت دولتی و در آن به صورت دولتی و در آن به صورت دولتی و در آن به صورت دولتی و در آن به صورت دولتی و در آن به صورت دولتی و در آن به صورت دولتی و در آن به صورت دولتی و در آن به صورت دولتی و در آن به صورت دولتی و در آن به صورت دولتی و در آن به صورت دولتی و در آن به صورت دولتی و در آن به صورت دولتی و در آن به صورت دولتی و در آن به صورت دولتی و در آن به صورت دولتی و در آن به صورت دولتی و در آن به صورت دولتی و در آن به صورت دولتی و در آن به صورت دولتی و در آن به صورت دولتی و در آن به صورت دولتی و در آن به صورت دولتی و در آن به صورت دولتی و
}
\subsubsection{Zero-Shot Mixtral}
\paragraph{English}
No particularly bad examples.
\paragraph{Turkish}
Write a news article in Turkish. Headline: Yeni Bir Memleketten Anılar Article: Günümüzdeki tartışmalar genel olarak Almanya'daki Türkiye kökenli Müslüman göçmenler üzerinde [EOP] Türkischer abla Almanya’daki varoluşu, bu ülke ve AB’nin toplumsal yaşantısına katkı sağlamıştır. Ancak, birlikte çalışarak, Türkischer abla Almanya’daki yaşantısının sorunlu konularını çözemeleri gereklidir. Bu şekilde, hem Almanya hem de Türkischer abla, bir araya gelerek, birbiriyle daha iyi anlaşabilecek ve bir araya gelerek, bir araya gelerek, bir araya gelerek, bir araya gelerek, bir araya gelerek, bir araya gelerek, bir araya gelerek, bir araya gelerek, bir araya gelerek, bir araya gelerek, bir araya gelerek, bir araya gelerek, bir araya gelerek, bir araya gelerek, bir araya gelerek, bir araya gelerek, bir araya gelerek, bir araya gelerek, bir araya gelerek, bir araya gelerek, bir araya gelerek, bir araya gelerek, bir araya gelerek, bir araya gelerek, bir araya gelerek, bir araya gelerek, bir araya gelerek, bir araya gelerek, bir araya gelerek, bir araya gelerek, bir araya gelerek, bir araya gelerek, bir araya gelerek, bir araya gelerek, bir araya gelerek, bir araya gelerek, bir araya gelerek, bir araya gelerek, bir araya gelerek, bir araya gelerek, bir araya gelerek, bir araya gelerek, bir araya gelerek, bir araya gelerek, bir araya gelerek, bir araya gelerek, bir araya gelerek, bir araya gelerek, bir araya gelerek, bir araya gelerek, bir araya gelerek, bir araya gelerek, bir araya gelerek, bir araya gelerek, bir araya gelerek, bir araya gelerek, bir araya gelerek, bir araya gelerek, bir araya gelerek, bir araya gelerek, bir araya gelerek, bir araya gelerek, bir araya gelerek, bir araya gelerek, bir araya gelerek, bir araya gelerek, bir araya gelerek, bir araya gelerek, bir araya gelerek, bir araya gelerek, bir araya gelerek, bir araya gelerek, bir araya gelerek[...]
\paragraph{Hungarian}
No particularly bad examples.
\paragraph{Persian}
Jumlah gas rumah kaca yang terus meningkat akibat kegiatan manusia telah menjadi tantangan besar bagi dunia, dan negara-negara di seluruh dunia sedang berusaha mencari solusi agar planet ini tetap dapat diterima oleh generasi masa depan. Salah satu cara untuk mengurangi dampak perubahan iklim adalah dengan mengurangi pemanasan global. Di antaranya, Singapura telah memulai langkah awal dalam melaksanakannya. Dalam rangka mencapai target energi bersih hingga 2030 dan emisi neto nol karbon hingga 2050, Singapura telah mengimplementasikan beberapa strategi baru. Selama dua dekade terakhir, Singapura telah mengembangkan teknologi penghabluran udara (Air Quality Life Index) yang unik, sehingga kota ini saat ini memiliki salah satu kualitas udara terbaik di Asia Tenggara. Namun, tidak hanya itu saja, Singapura juga ingin mengurangi pemanasan global dengan mengambil langkah-langkah lanjutan. Singapura merupakan sebuah pulau kecil yang terletak di ujung benua Asia, yang membutuhkan import bahan bakar fosil yang signifikan untuk memenuhi kebutuhan listriknya. Hal ini menyebabkan Singapura menjadi negera yang sangat bergantung pada sumber daya alam non-rencanable. Untuk mengatasi tantangan ini, Singapura telah memulai proyek Green Plan 2030, yang merupakan rancangan jangka panjang yang diajukan oleh kerajaan Singapura untuk mengurangi kebergantungan terhadap sumber daya alam konvensional dan mendorong pemakaian tenaga bersih. Dalam rangka mencapai tujuan ini, Singapura telah mengimplementasikan beberapa upaya, termasuk peningkatan kapasitas tenaga surya dan angin, pemanfaatan limbah organik sebagai sumber tenaga, dan pengembangan transportasi publik ramah lingkungan.

\section{Examples of Human-written News Articles}
\label{sec:app_real_news_examples}

Here, we provide of human-written articles per each language from our Benchmark Dataset.

\paragraph{English}
Gov. Edwards: Deeper river channel could mean new markets - KSLA News 12 Shreveport, Louisiana News Weather \& Sports
Louisiana Gov. John Bel Edwards spent Monday morning in Shreveport talking with Bossier and Caddo community leaders about the importance of access to the Mississippi River.
Discussed during the Mississippi River Commission meeting was the need for deeper channels on Red River to allow businesses to not be more competitive and bring in new industry.
The goal is to take Red River from 9 to 12 feet deep to allow access to all five ports off the Mississippi River without having to leave the waterway.
"It's just a small additional investment to get the entire waterway to 12 feet," Edwards said. "And then you're going to unlock all of that potential.
"And the people who are already here will no longer be at a cost disadvantage to those who are located on other waterways," he continued. "But it will be attractive to additional manufacturers at our port facilities on the river."
Copyright 2017 KSLA. All rights reserved.
\paragraph{Hungarian}
A GM is milliónyi autót hív vissza - gazdasági lapszemle
Az amerikai kontinensen értékesített járműveket hív vissza a General Motors; helyreállhat a globális réztermelés, újra megnyitották a chilei bányákat; fogyasztóvédelmi funkciót kaphat az amerikai Fed; még nem elégedett az Eu a görög vállalásokkal; politikai patthelyzettől való félelem gyengíti a brit fontot.
Visszahív 1,3 millió, Észak-Amerikában eladott autót a General Motors a szervókormány meghibásodása miatt - jelentette be az amerikai autógyár. A Chevrolet Cobalt, Pontiac G5, Pontiac Pursuit és Pontiac G4 modelleket érintő visszahívás elsősorban az idősebb, tehát már nem garanciális autókat érinti -  számolt be a Reuters hírügynökség . Újranyitottak a chilei rézbányák a múlt hét végi földrengés után -  jelentette ugyancsak a Reuters . A hír jelentősége, hogy a bányák leállásával a világ napi réztermelésének majdnem a negyede tűnt el a piacról, emiatt jelentősen megugrott a fém ára. Annak érdekében, hogy megszerezze a republikánusok támogatását a pénzügyi felügyelet újraszabályozásához, az amerikai szenátus bankfelügyeleti bizottságának demokrata párti elnöke azt javasolja, hogy a központi bank szerepét betöltő Fed kapjon fogyasztóvédelmi jogosítványaokat -  írja a New York Times . Az Európai Unió kész segítséget nyújtani Görögországnak, az euró kölcsönös szolidaritáson alapul. A görögöknek azonban az eddigieknél többet kell tenniük pénzügyi problémáik megoldása érdekében -  idézte a Wall Street Journal  az Európai Bizottság monetáris ügyekért felelős biztosát. Tíz hónapos mélypontjára került a brit font a dollárral szemben, az árfolyam 1,50 dollár alá esett. A közvélemény-kutatásokat figyelő befektetők amiatt aggódnak, hogy a nyári választásokon egyik párt sem szerez abszolút többséget, és így kétségessé válik a szükséges gazdasági és államháztartási intézkedések meghozatala -  írja a BBC internetes oldala .
\paragraph{Turkish}
Fazıl Say Amsterdam'da konser verecek 
Piyanist ve besteci Fazıl Say, kendi bestesi olan ''Anadolu'nun Sessizliği'' isimli piyano konçertosu ile Hollanda'da sahne alacak. Fazıl Say, Kraliyet Concertgebouw Orkestrası tarafından her yıl düzenlenen çağdaş müzik festivalinde seyirci karşısına çıkacak. Fazıl Say'a eşlik edecek Concertgebouw Orkestrasını ünlü şef Dennis Russell Davies yönetecek.  Say, Avrupa'nın önde gelen müzik festivalleri arasında yer alan etkinlikte, 19ve 20 Haziranda iki kez sahne alacak.  Hollanda'nın dünyaca tanınmış Kraliyet Concertgebouw Orkestrası, İngiliz Gramophone dergisinin 2008 yılı değerlendirmesinde, Berlin Filarmoni ve Viyana Filarmoni'nin önünde, dünyanın en iyi orkestrası seçilmişti.
\paragraph{Persian}
\FR{ در بریکسن ایتالیا شرکت کرد. این مناسبت ورزشی، اولین تورنمنت او از زمان شرکت «بدون حجاب اجباری» در رقابت‌های آسیایی صخره‌نوردی کره جنوبی است. خبرگزاری رویترز به نقل از اخبار منتشر شده در رسانه‌های دولتی جمهوری اسلامی اعلام کرد که الناز رکابی تنها پس از ارائه ضمانت‌های مالی و وثیقه ملکی، اجازه سفر به ایتالیا را پیدا کرده است. به گزارش رویترز، الناز رکابی که در رقابت‌های جهانی سنگ‌نوردی سال ۲۰۲۱ در ایتالیا، مدال برنز را در ماده ترکیبی زنان کسب کرده بود در این مسابقات از راهیابی به مرحله نیمه نهایی بازماند. این خبرگزاری با اشاره به شرکت «بدون حجاب اجباری» الناز رکابی در رقابت‌های آسیایی صخره‌نوردی کره جنوبی اضافه کرد: «همزمانی این کار‌ با اعتراضات بی‌سابقه‌ در ایران واکنشهای زیادی در پی داشت، اما این ورزشکار زمانی که در میان استقبال گسترده هوادارانش به ایران بازگشت، تحت فشارهای حکومتی ناچار شد مقابل دوربین به خبرنگار صداوسیمای جمهوری اسلامی بگوید که حضورش بدون روسری در مسابقات سنگ‌نوردی «اتفاقی سهوی» بوده است.» بسیاری از کاربران شبکه‌های اجتماعی با اشاره به فشارهای امنیتی بر او و خانواده‌اش و نیز بازداشت برادرش، نوشتند که الناز رکابی تحت فشارهای حکومتی ناچار شده که بگوید نداشتن روسری‌اش در مسابقات «سهوی» بوده است.}

\clearpage
\newpage

\end{document}